\documentclass{article}
\usepackage{arXiv_v1,times}

\usepackage{amsmath,amsfonts,bm}

\def\eqref#1{equation~\ref{#1}}

\def\1{\bm{1}}

\DeclareMathAlphabet{\mathsfit}{\encodingdefault}{\sfdefault}{m}{sl}
\SetMathAlphabet{\mathsfit}{bold}{\encodingdefault}{\sfdefault}{bx}{n}

\usepackage{booktabs}
\usepackage{graphicx}
\usepackage{array}
\usepackage{xcolor}
\usepackage{colortbl}
\usepackage{wrapfig}
\usepackage{placeins}
\usepackage{fontawesome5}

\definecolor{rankfirst}{HTML}{A0D8B5}
\definecolor{ranksecond}{HTML}{E0F2A2}
\definecolor{rankthird}{HTML}{FFF4B8}

\newcommand{\rkone}[1]{%
  \cellcolor{rankfirst}\textcolor{black}{#1}}
\newcommand{\rktwo}[1]{%
  \cellcolor{ranksecond}\textcolor{black}{#1}}
\newcommand{\rkthree}[1]{%
  \cellcolor{rankthird}\textcolor{black}{#1}}

\DeclareRobustCommand{\rkonecap}[1]{%
  \colorbox{rankfirst}{\textcolor{black}{\strut #1}}%
}

\DeclareRobustCommand{\rktwocap}[1]{%
  \colorbox{ranksecond}{\textcolor{black}{\strut #1}}%
}

\DeclareRobustCommand{\rkthreecap}[1]{%
  \colorbox{rankthird}{\textcolor{black}{\strut #1}}%
}

\usepackage{caption}
\usepackage{hyperref}
\hypersetup{hidelinks}
\usepackage{url}
\usepackage{amsmath}
\usepackage{cleveref}

\title{Touch-JEPA: Tactile Representation Learning from Hand-Object Video with Limited Tactile Supervision}

\author{
\textbf{Minseong Kweon}$^{1,2}$,
\textbf{Qiao Feng}$^{2}$,
\textbf{Yiming Huang}$^{2}$,
\textbf{Lingjie Liu}$^{2}$\\
$^{1}$University of Minnesota,
$^{2}$University of Pennsylvania
}

\iclrfinalcopy

\begin{document}

\maketitle

\vspace{-2em}
\begin{center}
    \small
    \textbf{\textcolor{purple!80}{Project Page: }}%
    \href{https://touch-jepa.github.io}{\textcolor{black}{\nolinkurl{https://touch-jepa.github.io}}}
\end{center}

\begin{abstract}
Understanding contact and tactile magnitude in hand-object interaction is a key step toward physically intelligent systems, yet learning such representations is limited by the cost and restricted coverage of tactile measurements. We present Touch-JEPA, a framework for learning tactile representations from hand-object videos with limited tactile supervision. 
We first introduce a self-supervised pretraining objective that learns from video alone without tactile supervision. Using video-derived hand geometry, Touch-JEPA predicts masked visual features at corresponding hand-surface locations from spatial and temporal context, encouraging the representation to capture latent contact dynamics. We then extend this formulation to joint representation learning, where masked prediction is applied to all videos while tactile supervision is available only for a labeled subset, grounding the shared representation in measured contact and magnitude. With self-supervised pretraining alone, simple supervised readouts on frozen Touch-JEPA features outperform supervised baselines in contact discrimination and tactile magnitude estimation across three human-hand benchmarks, and transfer to robotic hands without robot tactile training data. Joint learning with limited tactile supervision further improves contact and magnitude estimation on the human-hand benchmarks. Together, these results show that hand-object video can provide scalable self-supervision for tactile representation learning, while limited tactile measurements provide complementary physical grounding.

\end{abstract}

{\centering
\includegraphics[width=\textwidth]{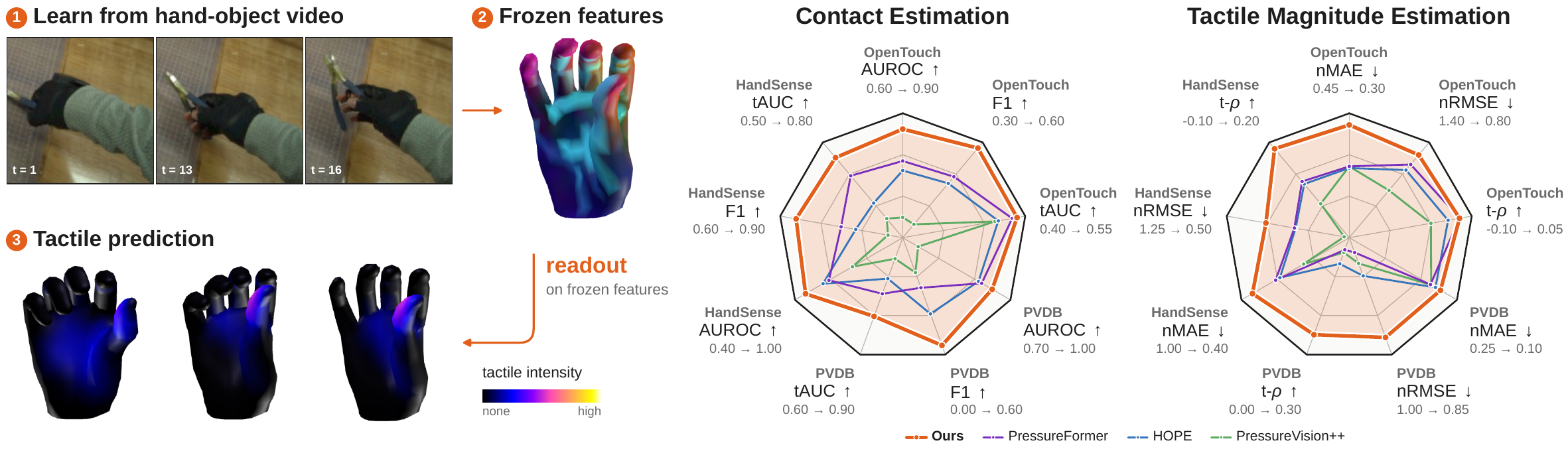}
\captionof{figure}{
Touch-JEPA learns tactile representations from video.
Readouts on frozen features outperform the supervised baselines
in contact and tactile magnitude estimation.
Axis numbers give the center-to-ring range, and outer is better.
}
\label{fig:teaser}
}

\section{Introduction}
\label{sec:intro}

Physical understanding of hand-object interaction underpins contact-rich manipulation~\citep{xue2025reactive}, the transfer of dexterous skill from human video to robots~\citep{qin2022dexmv,shaw2023videodex}, and clinical hand-function assessment~\citep{zhao2024automated}. Yet visual observations reveal hand-object geometry and motion while contact state and tactile magnitude remain only indirectly observable~\citep{grady2022pressurevision}. Learning representations that capture such tactile information from visual and temporal cues could therefore provide a richer understanding of hand-object interaction beyond geometry and motion~\citep{zhao2025egopressure}.

Prior work has largely approached tactile understanding from video through supervised prediction. Early approaches exploited load-related visual cues, including soft-tissue deformation, fingertip color changes, and cast shadows, to predict planar pressure maps~\citep{grady2022pressurevision,grady2024pressurevision++}. More recent methods extend this setting to pressure on 3D hand meshes~\citep{zhao2025egopressure,jeon2026hope}, forces on hand and object meshes given object geometry~\citep{ilic2026egophi}, and dense hand-surface contact across diverse interactions~\citep{jung2026learning}. Alongside these methods, paired hand-object video and sensor datasets have enabled supervised learning and evaluation for pressure prediction, contact estimation, retrieval, and classification~\citep{song2025opentouch,zhang2026glove2hand,zhou2026touchanything,zeng2026egotactile}.  These advances show the potential of learning tactile information from visual observations, but depend on paired tactile measurements for supervision.

However, paired tactile data are difficult to acquire at scale~\citep{jiang2024capturing}. Collecting such supervision often requires specialized hardware and careful calibration. Moreover, tactile gloves alter bare-hand appearance~\citep{zhang2026glove2hand}, planar sensors restrict contact geometry, and instrumented objects limit object diversity~\citep{brahmbhatt2019contactdb}. Simulation and derived contact labels alleviate some of these sensing constraints but often depend on task-specific pipelines~\citep{wang2022tacto}. Consequently, existing tactile datasets provide strong physical grounding but only fragmented coverage of hand regions, objects, and interactions. 

In contrast, large-scale video datasets capture diverse hand-object interactions across broad visual and behavioral distributions, but lack explicit tactile annotation~\citep{damen2022rescaling,grauman2022ego4d}. Our key insight is that tactile supervision and large-scale unlabeled video provide complementary learning signals: tactile measurements ground the representation in physically meaningful quantities, while diverse interaction videos provide broader spatial, anatomical, and temporal context. Self-supervised masked prediction~\citep{assran2025v} can therefore leverage unlabeled videos to learn interaction regularities beyond the limited tactile-labeled subset, while explicit tactile supervision preserves grounding in contact and magnitude. Together, these signals enable tactile representations to benefit from both physically grounded supervision and the scale and diversity of natural interaction videos.

We introduce Touch-JEPA, a framework for tactile representation learning from hand-object interaction videos with limited tactile supervision. In the self-supervised setting, Touch-JEPA learns from
video without tactile annotations through masked prediction.
Video-derived hand geometry defines corresponding
hand-surface locations, allowing visual features to be
predicted from spatial and temporal interaction context
across heterogeneous videos. A Geometry Encoder integrates evidence across hand regions and frames under interaction and anatomical masking. We then extend the framework to joint representation learning, where the JEPA objective is applied across all videos while tactile supervision is available only for the labeled subset. The two objectives share the encoder, allowing tactile measurements to ground the learned representation in contact and magnitude while self-supervised prediction continues to exploit the scale and diversity of the broader video distribution. We further control interference between the two learning signals during joint training to preserve their complementary information. Our contributions are summarized below:

\begin{itemize}

\item We introduce Touch-JEPA, a unified tactile representation learning framework that treats abundant hand-object video and limited tactile measurements as complementary sources of supervision: video provides scalable interaction coverage, while tactile measurements provide direct physical grounding of the shared representation.

\item We propose a geometry-guided self-supervised pretraining objective that learns tactile-relevant hand-surface representations from video alone. Touch-JEPA predicts masked latent features at anatomically corresponding surface locations, using video-derived hand geometry together with interaction and anatomical masking to integrate spatial and temporal evidence without requiring tactile annotations.

\item We develop a joint representation learning strategy that incorporates tactile supervision from a limited labeled subset while retaining self-supervised learning across all videos. Tactile supervision shapes the shared encoder and its EMA teacher, allowing physically grounded information to propagate through representation-level self-distillation without tactile pseudo-labels, while gradient control preserves the complementary learning signals.
\end{itemize}

\section{Related Work}
\label{sec:relatedwork}

\noindent\textbf{Tactile Representation and Estimation.}
Instrumented gloves enable dense hand-object contact sensing~\citep{sundaram2019learning,song2025opentouch}, while prior work learns tactile
representations through autoencoding, cross-modal self-supervision, and
cross-sensor alignment~\citep{murali2018learning,zambelli2021learning,yang2022touch,kerr2022self,
yang2024binding,feng2025anytouch,wi2025tactalign}.
Recent methods further use masked visuo-tactile reconstruction or joint-embedding
objectives for tactile observations~\citep{zhu2026touch,higuera2024sparsh}.
Tactile-JEPA~\citep{kovtun2026tactile} applies joint-embedding prediction to the
signals of distributed tactile sensors, whereas
Touch-JEPA learns touch from video and never takes a tactile signal as input.
A complementary line of work estimates tactile quantities from vision, including
dense contact~\citep{jung2026learning}, planar pressure~\citep{grady2022pressurevision,grady2024pressurevision++}, hand-surface pressure~\citep{zhao2025egopressure,jeon2026hope}, and joint hand-object forces~\citep{ilic2026egophi}.
Paired video-tactile datasets support supervised prediction~\citep{zhou2026touchanything,zeng2026egotactile}, and concurrent work scales such
supervision with large glove recordings and RGB-only inference~\citep{zhang2026egotac,zhou2026touchsight}.
These methods, including vertex-token video models~\citep{jeon2026hope}, are trained on
tactile or contact labels.
In contrast, Touch-JEPA learns tactile representations from video without tactile
targets during self-supervised pretraining and evaluates whether contact and tactile magnitude
can be decoded from the resulting frozen features.

\noindent\textbf{Self-Supervised and Hybrid Representation Learning.}
Self-supervised visual models learn transferable representations through
self-distillation and masked prediction~\citep{caron2021emerging,zhou2021ibot,oquab2023dinov2,simeoni2025dinov3}.
Joint-embedding predictive architectures learn by predicting unobserved
representations in images~\citep{assran2023self} and videos~\citep{bardes2024revisiting,assran2025v,mur2026v}, and such video encoders
already capture fine-grained physical-contact cues~\citep{alrasheed2026latent}.
Related work combines representation prediction with geometric, generative,
or action objectives.
VGGT-$\Omega$~\citep{wang2026vggtomega} learns reconstruction from
geometric annotations and subsequently extends training to unlabeled
videos through EMA self-distillation.
On paired text-to-code tasks, LLM-JEPA~\citep{huang2026llm} jointly
optimizes JEPA latent prediction and token-level supervision,
combining learned representation targets with explicit output sequences.
JEPT~\citep{luo2025learning} combines self-supervised visual transition
prediction on labeled and unlabeled videos with supervised inverse
dynamics learning on action-labeled demonstrations.
Touch-JEPA adapts latent prediction to anatomically corresponding
hand-surface locations across time and combines masked prediction
on all videos with tactile supervision on the labeled subset.
\section{Touch-JEPA}
\label{sec:method}

\begin{figure}[t]
\centering
\includegraphics[width=\textwidth]{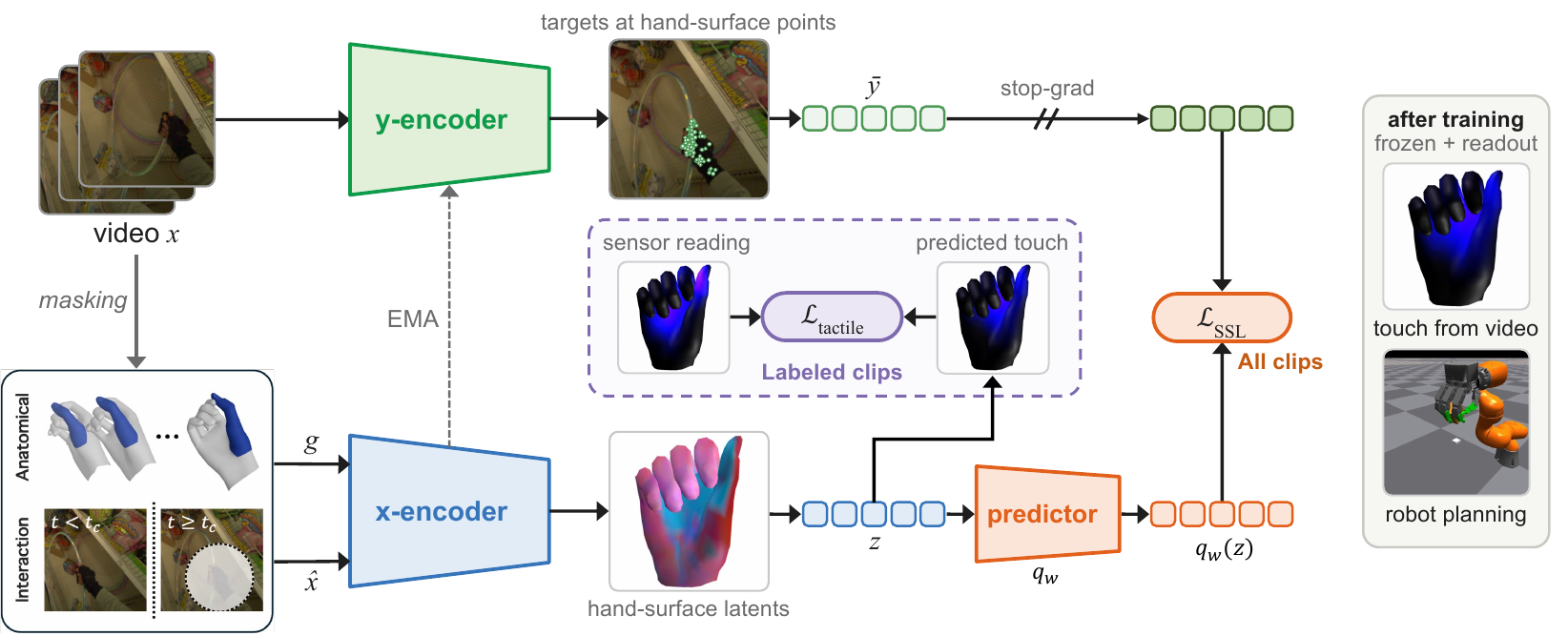}
\caption{Touch-JEPA overview.
Given a hand-object interaction video, the $y$-encoder processes unmasked RGB to produce target visual features at geometry-aligned hand-surface locations. The $x$-encoder combines masked RGB with estimated hand
geometry to predict these targets. When tactile annotations are available, joint training additionally supervises contact and magnitude predictions decoded from masked $x$-encoder states before the predictor, providing complementary physical grounding for the shared representation.}
\vspace{-5pt}
\label{fig:overview}
\end{figure}

Touch-JEPA learns hand-surface representations by predicting
masked visual features at anatomically corresponding locations
(Figure~\ref{fig:overview}).
The student $x$-encoder predicts targets from the teacher
$y$-encoder, updated by an exponential moving average (EMA).
Visual prediction uses all videos, while tactile supervision anchors the shared representation to measured contact and magnitude.

\paragraph{Problem formulation.}
Given a $T$-frame RGB clip $x$, a frozen HaWoR
estimator~\citep{zhang2025hawor} recovers
MANO~\citep{MANO:SIGGRAPHASIA:2017} geometry $g$ using
dataset-provided camera intrinsics.
Each frame contains $V=778$ vertices and $J=21$ joints
with consistent anatomical indices.
For vertex $v$ at time $t$, $u_{t,v}$ denotes its image
projection, $\iota_{t,v}$ its in-image indicator, and
$s_{t,v}$ its mesh-based visibility indicator.
Mesh visibility requires image inclusion, camera-facing
orientation, and no self-occlusion.
Occlusion by objects or the other hand is not modeled.

\subsection{Geometry-Guided Masked Visual Prediction}
\label{sec:prediction}

The $x$-encoder produces vertex and joint states
$z=f_\theta(\hat x,g,a)$ from masked RGB $\hat x$,
unmasked geometry $g$, and image-read permissions $a$.
The $y$-encoder observes unmasked RGB.
Its targets are computed directly from the visual field
without surface-state feedback.
Geometry specifies where to sample, while video determines
the target content, requiring prediction of visual features
at anatomically corresponding hand-surface locations.

Let $\bar h_t$ denote the $y$-encoder's video-feature field
aligned to frame $t$.
For backbone width $D$ and surface width $d=256$,
we define the projected target field and vertex targets as
\begin{equation}
\bar Y_t=\texttt{LayerNorm}(\bar h_tR),
\qquad
\bar y_{t,v}=\texttt{Sample}(\bar Y_t,u_{t,v}),
\label{eq:target}
\end{equation}
where $R\in\mathbb R^{D\times d}$ has fixed orthonormal
columns and $\texttt{Sample}$ denotes bilinear sampling.
Joint targets are sampled analogously at projected
joint locations.

\paragraph{Interaction masking.}
We select a split $t_c$ between context and target frames
based on valid hand observations.
For image-patch center $c_p$ and masking radius $r$,
the interaction mask is
\begin{equation}
m_{t,p}
=
\mathbf{1}[t\geq t_c]
\max_v
\left\{
s_{t,v}\,
\mathbf{1}[\|c_p-u_{t,v}\|\leq r]
\right\}.
\label{eq:interaction_mask}
\end{equation}
Pixels in masked patches are zeroed after RGB normalization.
The $x$-encoder must recover withheld evidence in target frames
using context frames, remaining scene regions, and
target-frame geometry.

\paragraph{Anatomical masking.}
We additionally select a set of finger or palm regions
$\mathcal H$ and mask neighborhoods around their projected
vertices throughout the clip.
Their vertex and joint tokens lose direct image-read access
during initial sampling and per-layer lifting.
Local image-window attention is also blocked for their
vertex tokens.
Geometry remains available, together with spatial and
temporal communication.
Interaction masking preserves visual context before $t_c$.
Anatomical masking additionally removes direct visual
evidence for selected regions across all frames,
encouraging inference from other hand regions.
Sampling rules are provided in Appendix~\ref{sec:supp-masking}.

\subsection{Geometry Encoder}
\label{sec:geometry_encoder}
\label{sec:arch}

Vertex states $H_t\in\mathbb R^{V\times d}$ and joint states
$S_t\in\mathbb R^{J\times d}$ pass through four layers
interleaved with the final four V-JEPA~2.1 ViT-B/16
blocks~\citep{mur2026v}.
Backbone adaptation is restricted to rank-$8$
LoRA~\citep{hu2021lora} in these blocks.
The visual path without surface feedback and the
interleaved path share backbone weights.
Local attention preserves anatomical structure, while
global and temporal exchange integrates evidence across
regions and frames.

\paragraph{Visual lifting.}
The initial field $F_t$ sums projected, temporally aligned
features of the masked clip $\hat x$ from the
$x$-encoder's visual path without surface feedback.
For vertex attributes $g_{t,v}$ extracted from $g$,
we initialize
\begin{equation}
\begin{aligned}
H^0_{t,v}
={}&\gamma(g_{t,v})+\texttt{PE}(v,t)
+a_{t,v}\phi_{\mathrm{img}}
  \bigl(\texttt{Sample}(F_t,u_{t,v})\bigr)
 +(1-a_{t,v})e_{\emptyset},
\end{aligned}
\label{eq:point}
\end{equation}
where $\gamma$ embeds geometry and visibility,
$\texttt{PE}$ encodes template position, identity,
anatomical region, and time,
$\phi_{\mathrm{img}}$ is an affine projection, and
$e_{\emptyset}$ represents missing image evidence.
Permission $a_{t,v}$ equals $\iota_{t,v}$ outside
anatomically masked regions and zero otherwise.
Joints are initialized analogously.
Each layer adds permitted samples from its ViT features.
Vertices also attend to local image windows containing
initial and current-layer features.

\paragraph{Anatomical attention.}
Vertices are partitioned into $20$ parts $\mathcal P$,
comprising $15$ phalangeal parts and five palm zones
assigned by geodesic proximity to the fingers.
Within-part attention first updates the vertex states.
Pooled part tokens then exchange information and broadcast
updates to their constituent vertices through
\begin{equation}
\begin{aligned}
h_{t,P}
&=
\phi_{\downarrow}\!\left(
|P|^{-1}\sum_{w\in P}
\texttt{LayerNorm}(H_{t,w})
\right),\\
H_{t,v}
&\leftarrow H_{t,v}
+\phi_{\uparrow}\!\left(
\texttt{Attn}
(h_{t,P(v)},\{h_{t,P}\}_{P\in\mathcal P})
\right),
\end{aligned}
\label{eq:inter_part}
\end{equation}
where $P(v)$ denotes the part containing vertex $v$,
$\phi_{\downarrow}$ projects to $64$ dimensions,
$\texttt{Attn}$ denotes multi-head attention,
and $\phi_{\uparrow}$ is a zero-initialized projection
back to width $d$.
A token-wise feed-forward network completes the block.

\paragraph{Global and temporal exchange.}
A global attention core combines $195$ downsampled surface
tokens with $21$ joint tokens at width $128$.
Full self-attention exchanges information across these tokens,
with residual updates returned directly to joints and through
fixed mesh upsampling to vertices.
A shared causal block then integrates each vertex and joint
track across frames, while the video encoder remains noncausal.
The first three layers feed surface states back to video
patches through gated bilinear splatting.
Final $\texttt{LayerNorm}$ operations produce $z$.
Architecture and initialization details appear in
Appendix~\ref{sec:supp-global-temporal}.

\subsection{Self-Supervised Training}
\label{sec:training}

A token-wise residual predictor
$q_\omega(z)=z+\texttt{MLP}_\omega(z)$
maps $x$-encoder states to $y$-encoder features without
mixing tokens or frames.
The MLP's final layer is zero-initialized, making
$q_\omega$ initially the identity.
For vertex or joint token $i$, the dense prediction loss is
\begin{equation}
\mathcal L_{\mathrm{dense}}
=
\Big\langle
\|q_\omega(z_{t,i})
-\texttt{sg}(\bar y_{t,i})\|_1/d
\Big\rangle_w,
\label{eq:jepa}
\end{equation}
where $\texttt{sg}$ denotes stop-gradient and
$\langle\cdot\rangle_w$ denotes a weighted mean computed
within each data lane and then averaged across lanes.
Each lane corresponds to a data-source group.
Weights prioritize mesh-visible, anatomically masked
vertices and visually supported target-frame tokens over
context tokens.
Visual support requires mesh visibility for vertices
and valid image projections for joints.
Eligible target vertices receive additional weight based
on changes in $y$-encoder features from the last context frame.
Exact weighting rules are provided in Appendix~\ref{sec:supp-dense}.

A position-conditioned decoder $D_\xi$ predicts patch
targets $\bar F_{t,p}\in\mathbb R^{d_p}$ from vertex and
joint states.
These targets are the normalized counterpart of $F_t$
computed by the $y$-encoder without surface feedback.
For patch-feature width $d_p$, the reconstruction loss is
\begin{equation}
\mathcal L_{\mathrm{patch}}
=
\Big\langle
\|D_\xi(z)_{t,p}
-\texttt{sg}(\bar F_{t,p})\|_1/d_p
\Big\rangle_{m_{t,p}=1}.
\label{eq:patch}
\end{equation}
The mean is computed over interaction-masked patches
within each lane and then averaged across lanes.
This loss encourages the representation to retain
surrounding interaction evidence beyond individual
vertex and joint samples.
The target-field construction and decoder are detailed
in Appendices~\ref{sec:supp-geometry} and~\ref{sec:supp-dense}.
The self-supervised objective is
\begin{equation}
\mathcal L_{\mathrm{SSL}}
=
\mathcal L_{\mathrm{dense}}
+0.5\mathcal L_{\mathrm{patch}}
+0.05\mathcal L_{\mathrm{wSIG}}.
\label{eq:representation_objective}
\end{equation}
Inspired by SIGReg~\citep{balestriero2025lejepa},
WeakSIGReg encourages the covariance of randomly projected vertex
features toward identity, discouraging dimensional collapse.

\subsection{Tactile Supervision}
\label{sec:tactile_sup}

Visual prediction learns from all videos, while tactile
labels anchor the shared representation to measured contact
and sensor-native magnitude.
Fixed dataset-specific correspondences map annotations
to sensor-supported vertices, with supervision determined
by annotation validity independently of image visibility.
A two-layer head maps masked $x$-encoder states before the
predictor to $(o,\eta)=h_\psi(z)$, with contact logit $o$
and raw magnitude $\eta$.
The contact probability and contact-conditional magnitude
are $\hat c=\sigma(o)$ and
$\hat m=\texttt{softplus}(\eta)$, respectively,
yielding the gated prediction $\hat y=\hat c\,\hat m$,
where $\sigma$ denotes sigmoid.

\begin{wrapfigure}[23]{r}{0.4\textwidth}
\vspace{-\intextsep}
\centering
\captionsetup{
    font=footnotesize,
    justification=raggedright,
    singlelinecheck=false,
    skip=3pt
}

\definecolor{compSSL}{HTML}{F4A259}
\definecolor{compLabels}{HTML}{3775BA}
\definecolor{compJoint}{HTML}{E2601B}

{\scriptsize
\quad
\textcolor{compLabels}{\rule[0.4ex]{0.8em}{1pt}}
\,Supervised
\quad
\textcolor{compJoint}{\rule[0.4ex]{0.8em}{1pt}}
\,Ours (Joint)
\par}

\vspace{3pt}
\includegraphics[width=\linewidth]
    {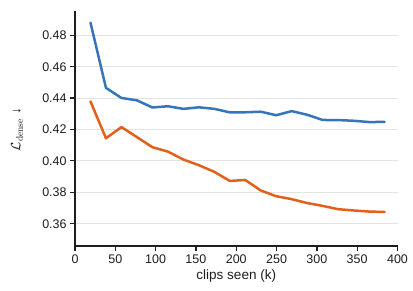}
\par
{\footnotesize (a) Visual prediction\par}

\vspace{4pt}
\includegraphics[width=\linewidth]
    {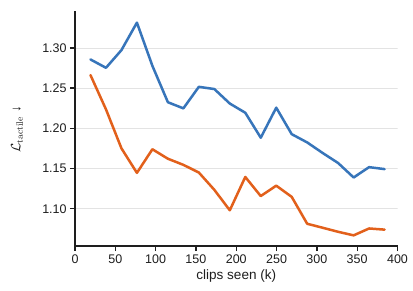}
\par
{\footnotesize (b) Tactile prediction\par}

\caption{
Visual and tactile prediction.
}
\label{fig:complementarity}
\end{wrapfigure}

For ground-truth magnitude $y$ normalized by a fixed
training-subset standard deviation, we define
\begin{equation}
\begin{aligned}[t]
&\mathcal L_{\mathrm{tactile}}
=\mathcal L_{\mathrm{contact}}\\
&\quad+\lambda_m
\Big\langle
|\log(1+\hat m)-\log(1+y)|
\Big\rangle_{\mathcal C}\\
&\quad+\mathcal L_{\mathrm{head}},
\end{aligned}
\label{eq:tactile_loss}
\end{equation}
where $\mathcal L_{\mathrm{contact}}$ is class-weighted
binary cross-entropy, $\mathcal C$ contains valid
ground-truth contact entries, and $\lambda_m$ weights
magnitude supervision per dataset.
Contact and log-magnitude losses update the encoder
and head.
The L1 loss $\mathcal L_{\mathrm{head}}$ fits gated
predictions to $y$ over all valid entries using
$\texttt{sg}(z)$, updating only the head.
Losses are normalized within labeled clips and averaged
across clips.

The joint objective is
\begin{equation}
\mathcal L
=
\mathcal L_{\mathrm{SSL}}
+\lambda_{\mathrm{tac}}(k)\mathcal L_{\mathrm{tactile}},
\label{eq:joint_objective}
\end{equation}
where $k$ denotes the training step and
$\lambda_{\mathrm{tac}}$ is zero for SSL only or warmed
up to one for joint training.
At $z$, we remove the tactile gradient component opposing
the SSL gradient~\citep{yu2020gradient} and cap its remaining
norm relative to an SSL-gradient reference
(Appendix~\ref{sec:supp-objectives}).
Tactile supervision updates the Geometry Encoder and
LoRA-adapted backbone through the masked $x$-encoder states,
without tactile pseudo-labels.
Visual targets are read from the $y$-encoder's image stream
without surface feedback.
For both variants, tactile estimation removes input masks
and fits a new readout on the frozen $y$-encoder using
training-split tactile labels.
The SSL predictor and any training-time tactile head
are discarded after training.

We compare joint training with Supervised using the
same encoder architecture and matched clip budgets.
Supervised omits visual prediction gradients to the
encoder, and its dense prediction loss plateaus.
Joint training continues to reduce this loss against
its own $y$-encoder targets
(Figure~\ref{fig:complementarity}a)
while maintaining lower held-out tactile loss than
Supervised throughout training
(Figure~\ref{fig:complementarity}b).
These trends support the benefit of visual prediction
for tactile learning while the predictive task itself keeps improving during training.

\begin{figure}[t]
  \centering
  \setlength{\tabcolsep}{1pt}
  \renewcommand{\arraystretch}{0.5}
  \newcommand{\rlabel}[1]{\rotatebox[origin=c]{90}{\tiny #1}}

  \begin{tabular}{cccc}
    & {\scriptsize PVDB} & {\scriptsize HandSense} & {\scriptsize OpenTouch} \\[2pt]
    \rlabel{RGB} & \raisebox{-0.5\height}{\includegraphics[width=0.318\linewidth]{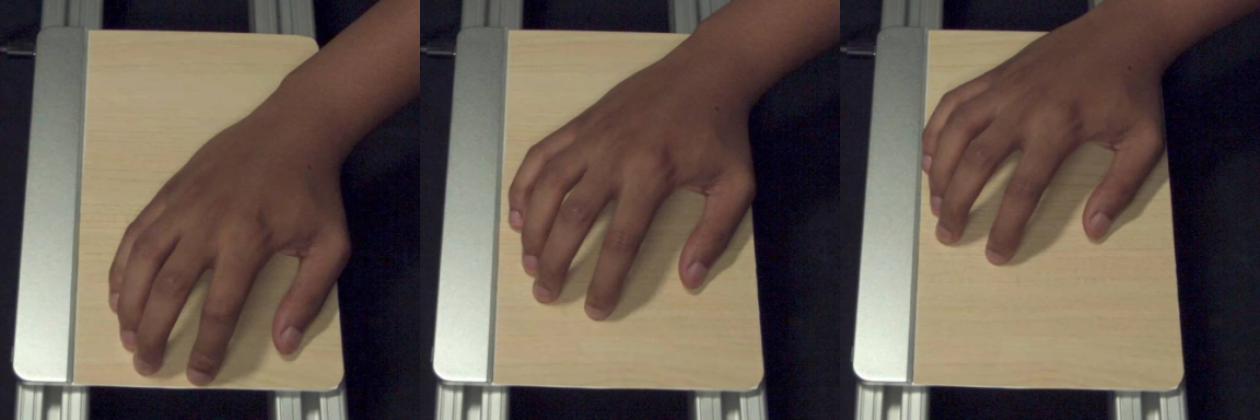}} & \raisebox{-0.5\height}{\includegraphics[width=0.318\linewidth]{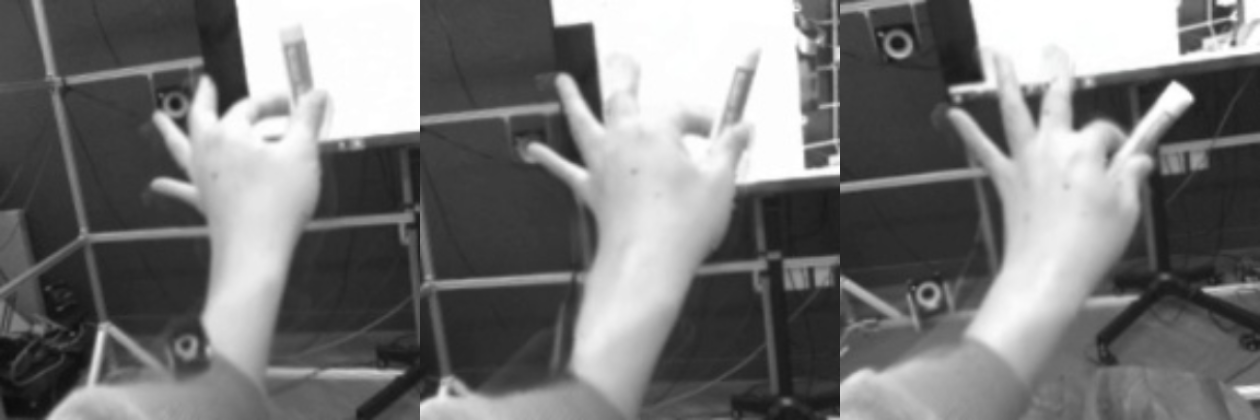}} & \raisebox{-0.5\height}{\includegraphics[width=0.318\linewidth]{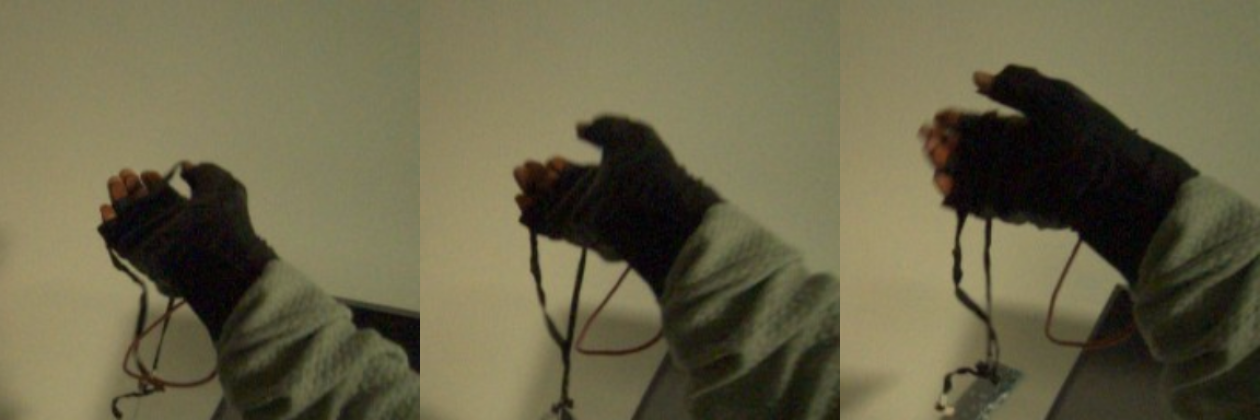}} \\
    \rlabel{GT} & \raisebox{-0.5\height}{\includegraphics[width=0.318\linewidth]{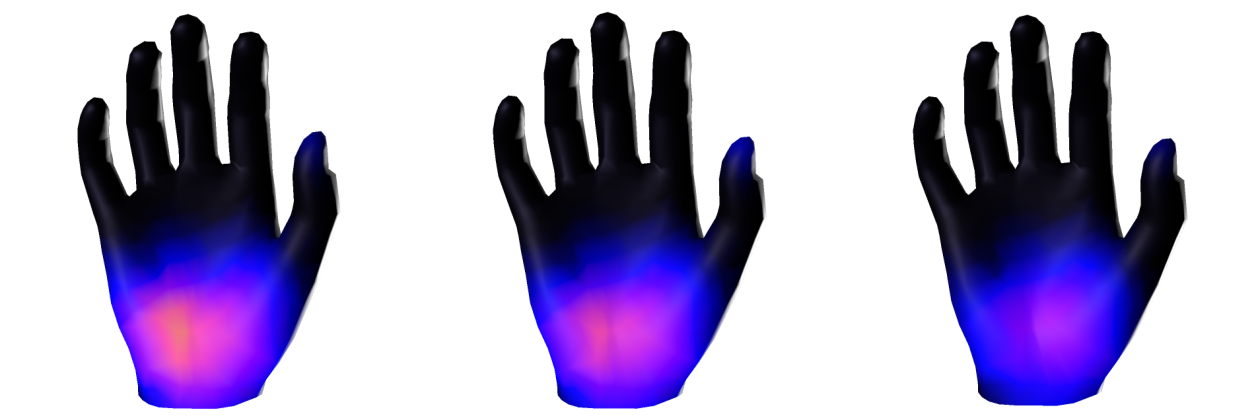}} & \raisebox{-0.5\height}{\includegraphics[width=0.318\linewidth]{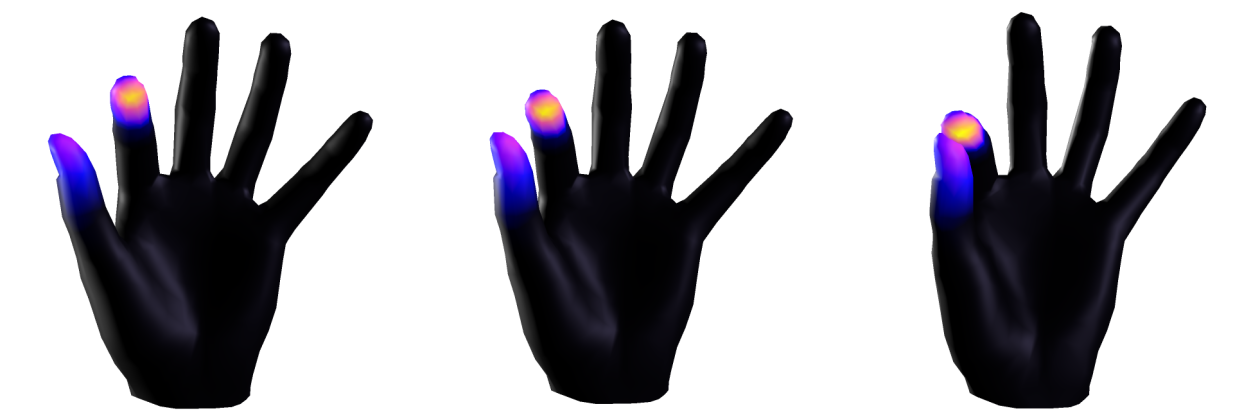}} & \raisebox{-0.5\height}{\includegraphics[width=0.318\linewidth]{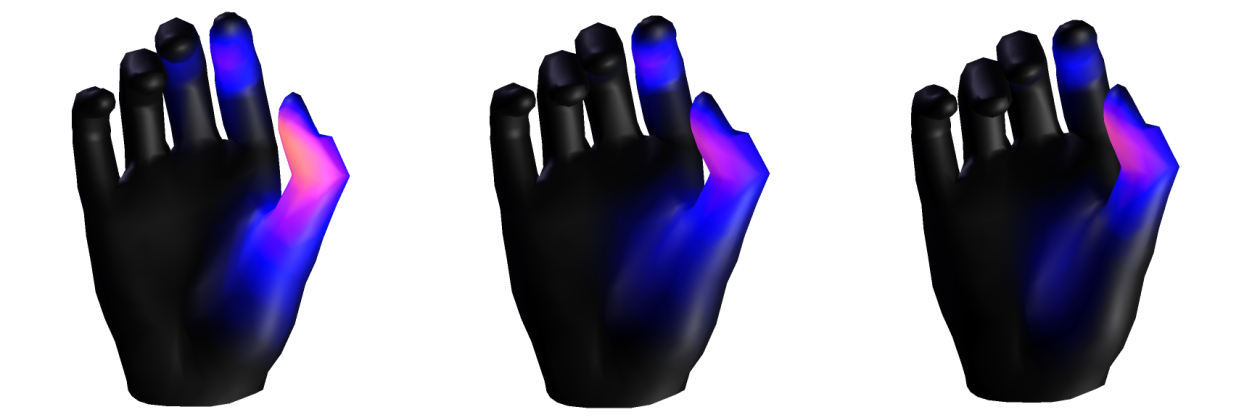}} \\
    \rlabel{\textbf{Ours (Joint)}} & \raisebox{-0.5\height}{\includegraphics[width=0.318\linewidth]{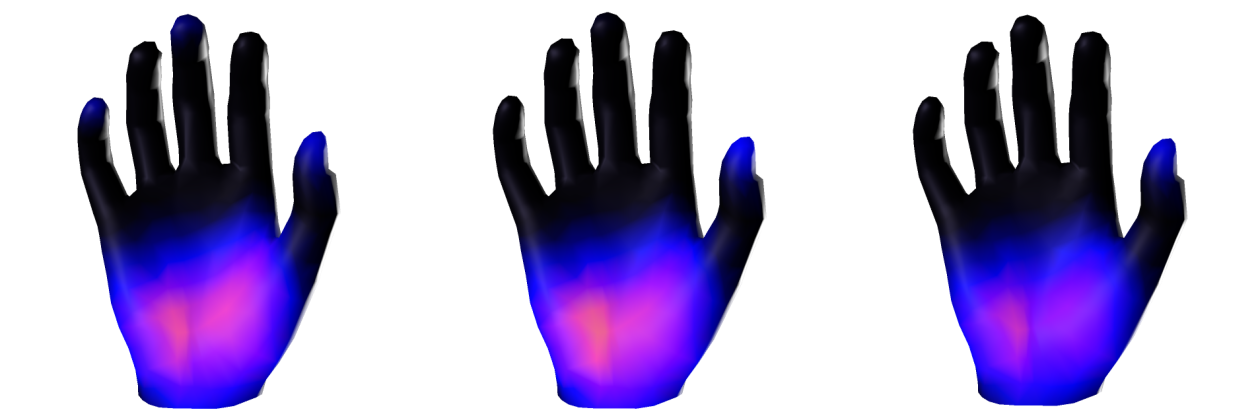}} & \raisebox{-0.5\height}{\includegraphics[width=0.318\linewidth]{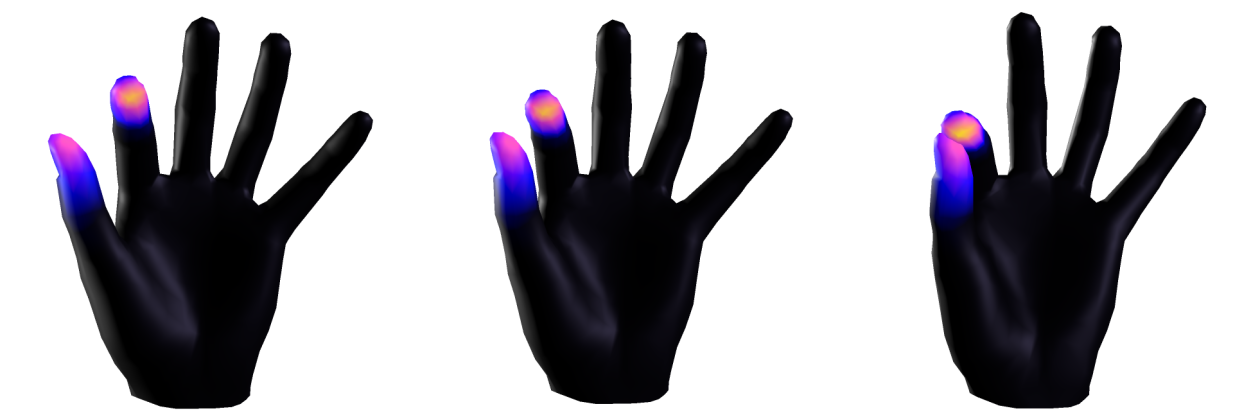}} & \raisebox{-0.5\height}{\includegraphics[width=0.318\linewidth]{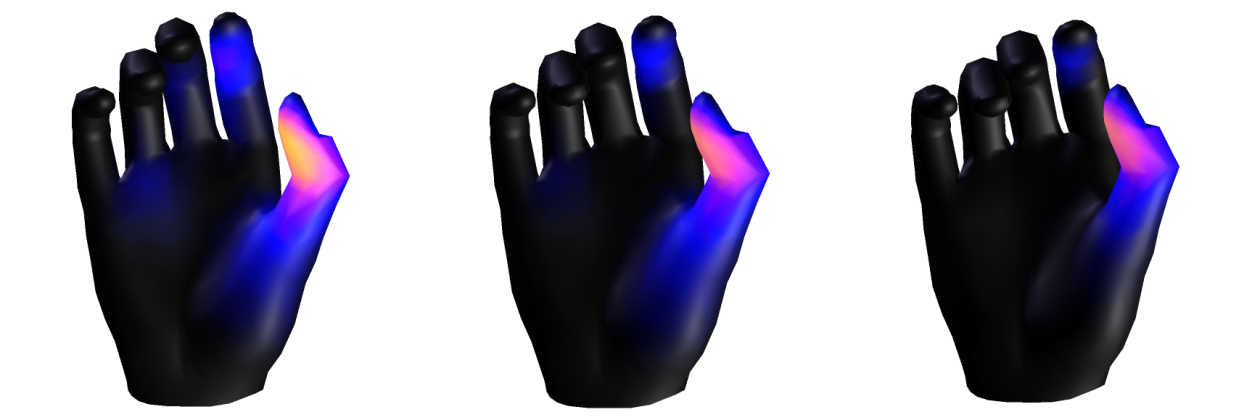}} \\
    \rlabel{\textbf{Ours (SSL)}} & \raisebox{-0.5\height}{\includegraphics[width=0.318\linewidth]{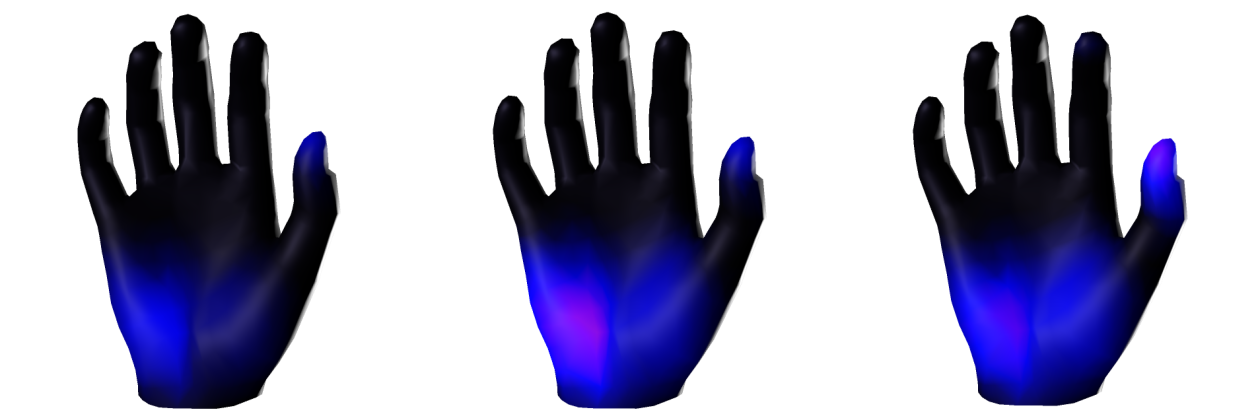}} & \raisebox{-0.5\height}{\includegraphics[width=0.318\linewidth]{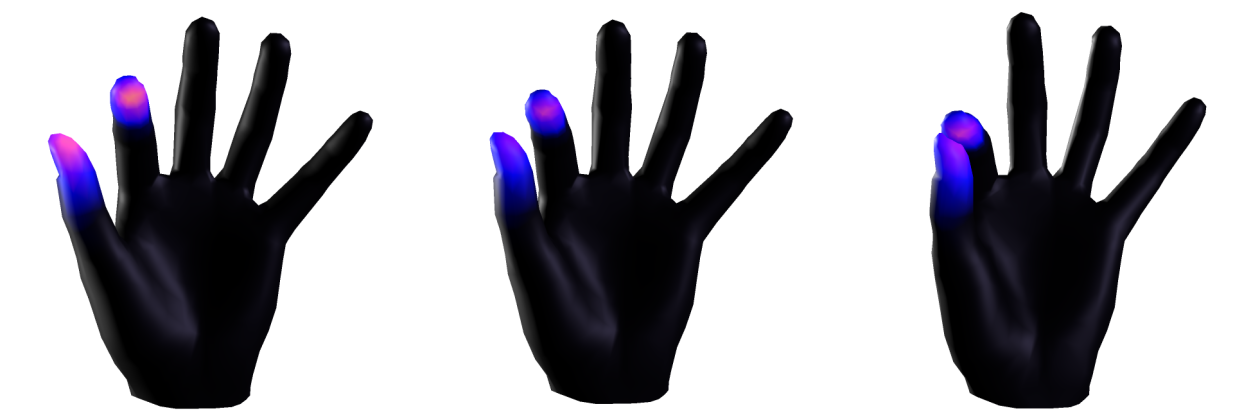}} & \raisebox{-0.5\height}{\includegraphics[width=0.318\linewidth]{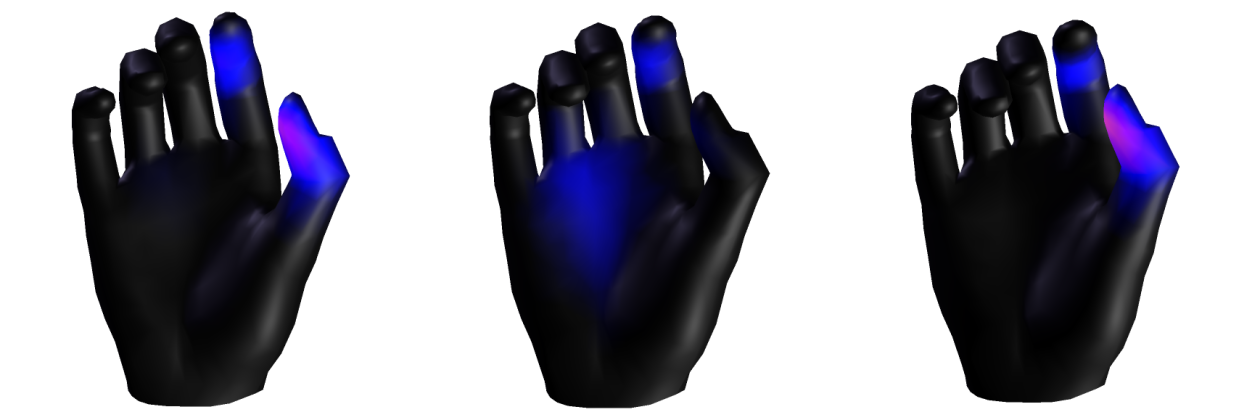}} \\
    \rlabel{HOPE} & \raisebox{-0.5\height}{\includegraphics[width=0.318\linewidth]{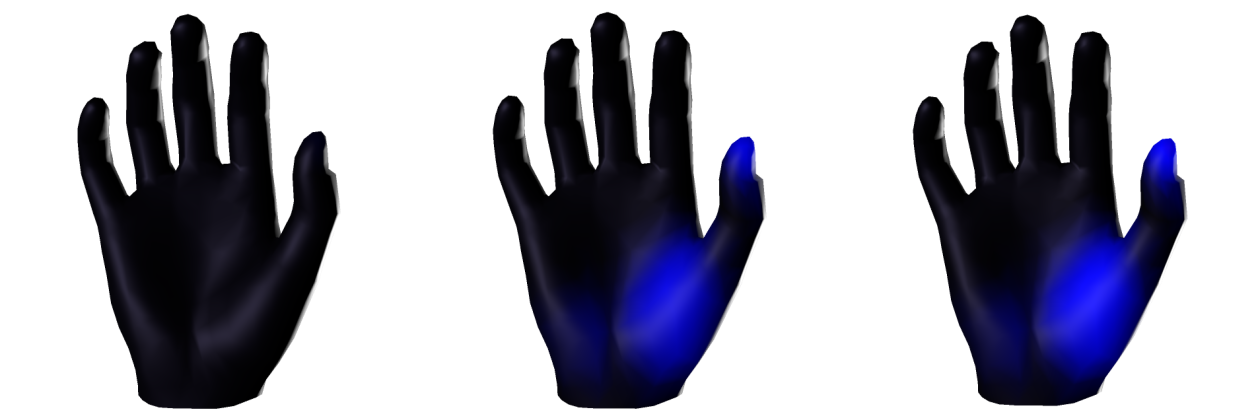}} & \raisebox{-0.5\height}{\includegraphics[width=0.318\linewidth]{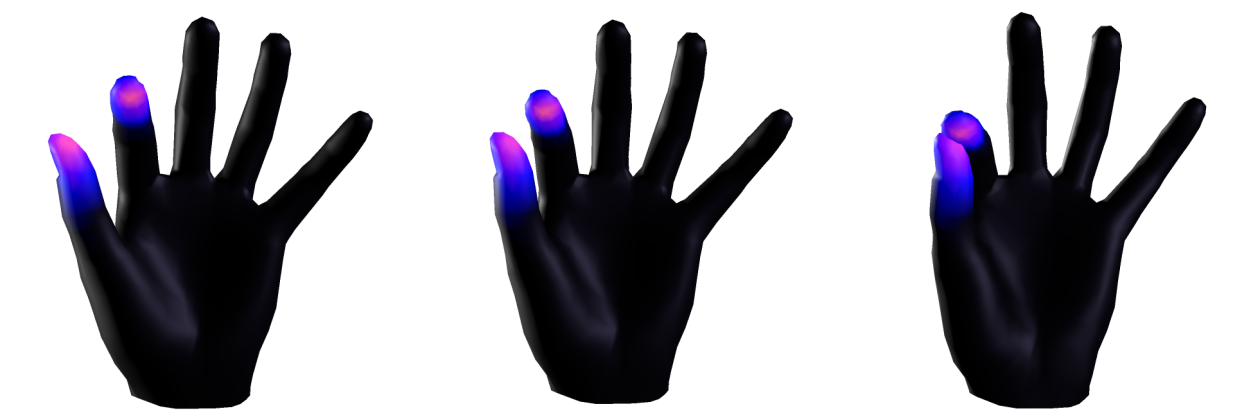}} & \raisebox{-0.5\height}{\includegraphics[width=0.318\linewidth]{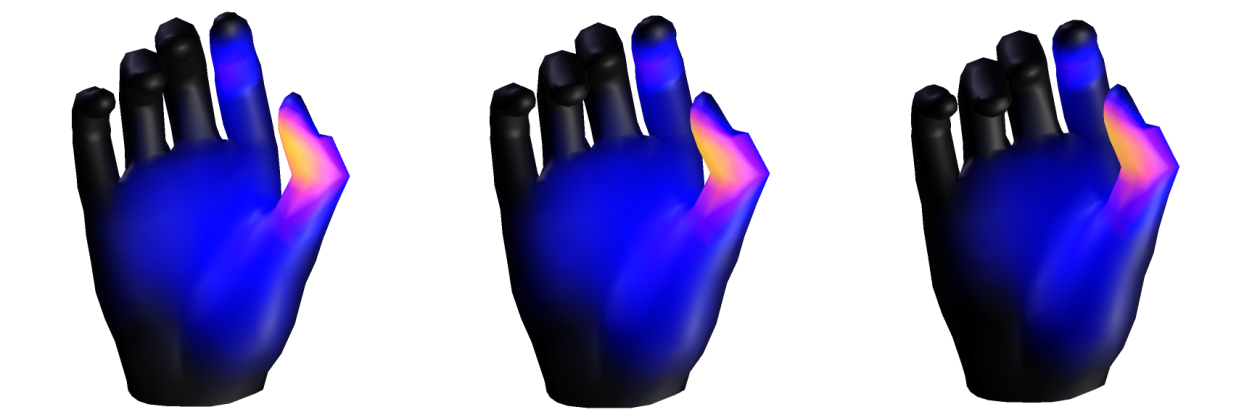}} \\
    \rlabel{PressureFormer} & \raisebox{-0.5\height}{\includegraphics[width=0.318\linewidth]{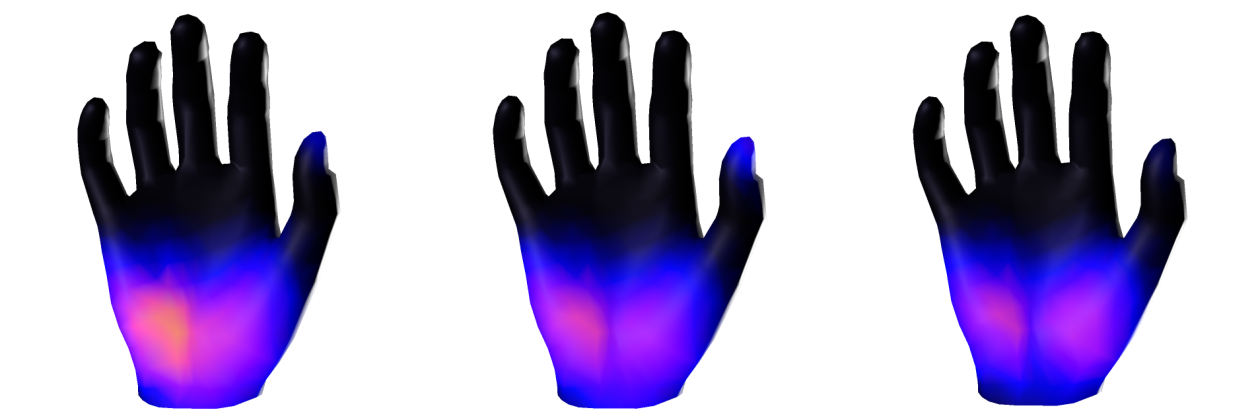}} & \raisebox{-0.5\height}{\includegraphics[width=0.318\linewidth]{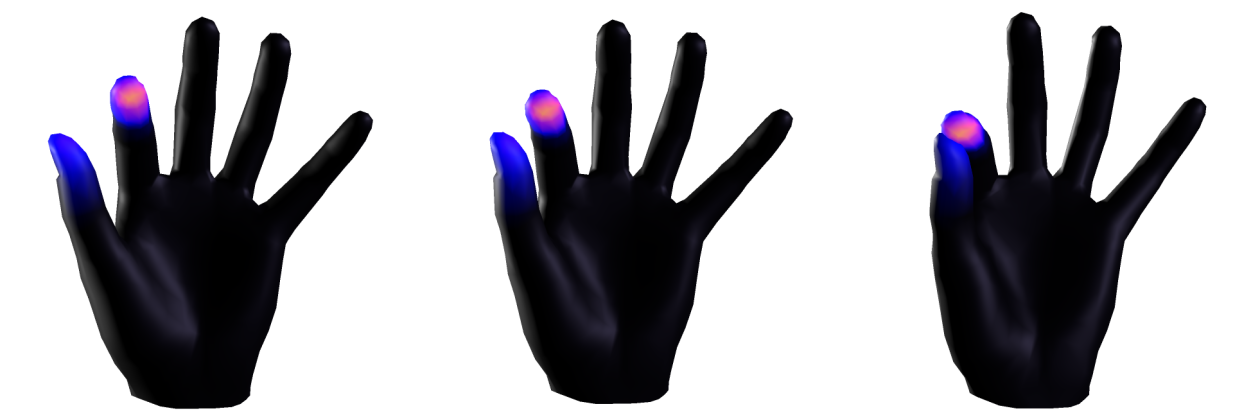}} & \raisebox{-0.5\height}{\includegraphics[width=0.318\linewidth]{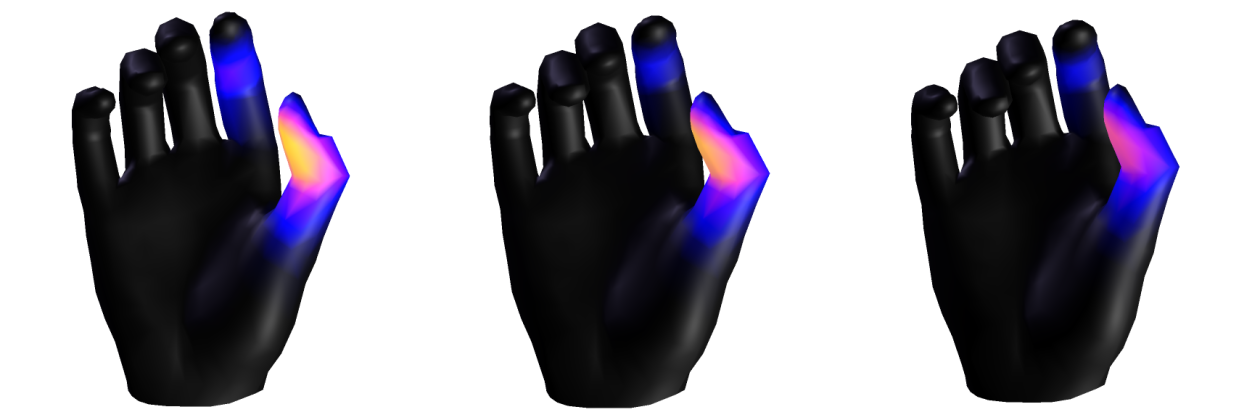}} \\
  \end{tabular}
  \caption{Qualitative comparison of tactile magnitude estimation. Three interaction sequences unfold from left to right, with brighter colors indicating greater tactile magnitude.}
  \vspace{-10pt}
  \label{fig:main-vis}
\end{figure}

\subsection{Touch-JEPA-AC}
\label{sec:touchjepa-ac}

We assess the utility of the learned representations for
simulated robot manipulation through action-conditioned
planning~\citep{assran2025v}.
Touch-JEPA-AC freezes the $y$-encoder and trains a
block-causal predictor to forecast pooled video features,
pooled hand-surface representations, and robot state
from observed history and issued actions.
A MANO proxy fitted to robot fingertips through forward
kinematics supplies the hand geometry.
Robot contact and force annotations are used only for evaluation.

At decision $\tau$, candidate action $\mathbf a$ produces
predicted scene and surface features
$\hat B_{\tau+1}(\mathbf a)$ and $\hat U_{\tau+1}(\mathbf a)$.
We compare them with reference subgoal features using
\begin{equation}
\begin{aligned}
E_\tau(\mathbf a)
={}&
\texttt{MAE}
\bigl(\hat B_{\tau+1}(\mathbf a),B_\tau^\star\bigr)
+
w_\tau\texttt{MAE}
\bigl(\hat U_{\tau+1}(\mathbf a),U_\tau^\star\bigr),
\end{aligned}
\label{eq:ac_energy}
\end{equation}
where $\texttt{MAE}$ averages absolute differences,
starred features encode the current subgoal, and $w_\tau$
is a task-dependent surface weight.
The cross-entropy method selects palm and grip commands
with a one-step horizon and replans after execution.
Scripted closing, test lifts, release, and motion constraints
complement planning.
Details are provided in Appendix~\ref{sec:supp-robot}.

\section{Experiments}

\subsection{Experimental Setting}
\label{sex:exp_setup}

\paragraph{Implementation details.}
We use 16-frame clips at $384\times384$ resolution.
The Geometry Encoder has four layers of width $256$
with four attention heads.
All three training settings process $384{,}000$ clip
presentations over $12{,}000$ updates with batch size $32$.
Our model trains on two NVIDIA L40 GPUs.
We use AdamW with learning rates of $10^{-4}$ for LoRA
and $3\times10^{-4}$ for new modules, with $4\%$ warmup
followed by cosine decay.
EMA momentum increases from $0.996$ to $0.9995$.
Further training and readout details appear in
Appendices~\ref{sec:supp-training}
and~\ref{sec:supp-readout}.

\paragraph{Datasets.}
Encoder training uses three tactile sensing datasets,
HandSense~\citep{zhang2026glove2hand},
PressureVision DB~\citep[PVDB,][]{grady2022pressurevision}, and
OpenTouch~\citep{song2025opentouch},
alongside standard hand-object interaction datasets,
HRDexDB-Human~\citep{lim2026hrdexdb},
DexYCB~\citep{chao2021dexycb},
HO3D~\citep{hampali2021ho},
HO-Cap~\citep{wang2026ho}, and
SHOWMe~\citep{swamy2023showme}.
All three training settings use the same video sources.
Joint training and the Supervised control also use the
tactile measurements. Contact and magnitude estimation are
evaluated on held-out test clips from OpenTouch, PVDB
(meshes fused across views), and HandSense.

\paragraph{Evaluation metrics.}
Within each comparison, methods are scored on identical valid entries.
For contact, we report pooled AUROC and F1.
Temporal AUROC (tAUC) is computed separately for each sensor
within a clip and averaged across eligible sequences.
For tactile magnitude, nMAE and nRMSE measure errors over these
entries, normalized by a fixed training-subset standard deviation
of each dataset's target magnitude.
Temporal Spearman correlation (t-$\rho$) is computed separately
for each sensor within a clip using only ground-truth contact frames,
then averaged across eligible sequences.
It measures within-contact temporal rank agreement in sensor-native
magnitude, not calibrated force recovery.
Differences are tested with paired bootstrap intervals over test clips.
Additional details for metrics are provided in Appendix~\ref{sec:supp-readout}.

\begin{table*}[t]
\centering
\caption{
Contact estimation.
Joint training achieves the best performance across all datasets,
while SSL alone outperforms supervised baselines on most metrics.
The \rkonecap{first}, \rktwocap{second}, and \rkthreecap{third}
ranks are assigned among learned methods,
and equal displayed values share a rank.
}
\vspace{-5pt}
\label{tab:contact_comparison}
\begingroup
\setlength{\tabcolsep}{4pt}
\renewcommand{\arraystretch}{1.15}
\small
\resizebox{0.85\textwidth}{!}{%
\begin{tabular}{@{}l*{9}{c}@{}}
\toprule
Method
& \multicolumn{3}{c}{OpenTouch}
& \multicolumn{3}{c}{PVDB}
& \multicolumn{3}{c}{HandSense} \\
\cmidrule(lr){2-4}
\cmidrule(lr){5-7}
\cmidrule(lr){8-10}
& AUROC $\uparrow$ & F1 $\uparrow$ & tAUC $\uparrow$
& AUROC $\uparrow$ & F1 $\uparrow$ & tAUC $\uparrow$
& AUROC $\uparrow$ & F1 $\uparrow$ & tAUC $\uparrow$ \\
\midrule
PressureVision
& 0.676 & 0.364 & 0.506
& 0.890 & 0.368 & \rkthree{0.748}
& 0.734 & 0.672 & 0.588 \\
PressureVision++
& 0.649 & 0.342 & 0.513
& 0.743 & 0.179 & 0.654
& 0.680 & 0.636 & 0.560 \\
PressureFormer
& \rkthree{0.785} & \rkthree{0.492} & \rktwo{0.534}
& \rkthree{0.920} & 0.257 & 0.744
& 0.811 & \rkthree{0.751} & \rkthree{0.695} \\
HOPE
& 0.762 & 0.471 & 0.517
& 0.911 & \rkthree{0.391} & 0.705
& \rkthree{0.844} & 0.715 & 0.609 \\
\midrule
\textbf{Ours (SSL)}
& \rktwo{0.836} & \rktwo{0.549} & \rkthree{0.527}
& \rktwo{0.925} & \rktwo{0.475} & \rktwo{0.783}
& \rktwo{0.914} & \rktwo{0.826} & \rktwo{0.699} \\
\textbf{Ours (Joint)}
& \rkone{0.862} & \rkone{0.582} & \rkone{0.540}
& \rkone{0.949} & \rkone{0.553} & \rkone{0.802}
& \rkone{0.941} & \rkone{0.861} & \rkone{0.752} \\
\midrule
\textit{Static prior}
& 0.832 & 0.542 & 0.500
& 0.819 & 0.335 & 0.500
& 0.559 & 0.618 & 0.500 \\
\bottomrule
\end{tabular}%
}
\endgroup

\vspace{-8pt}
\end{table*}

\begin{table*}[t]
\centering
\caption{
Tactile magnitude estimation.
Joint training achieves the best displayed results,
while SSL alone surpasses the retrained supervised baselines
on most metrics.
Targets are nominal pressure in PVDB and uncalibrated sensor
magnitudes in OpenTouch and HandSense.
Errors are computed over the common valid entries.
The t-$\rho$ metric measures temporal rank agreement within
ground-truth contact.
N/A denotes undefined correlation for the static prior.
}
\label{tab:force_comparison}
\begingroup
\setlength{\tabcolsep}{4pt}
\renewcommand{\arraystretch}{1.15}
\small
\resizebox{0.85\textwidth}{!}{%
\begin{tabular}{@{}l*{9}{c}@{}}
\toprule
Method
& \multicolumn{3}{c}{OpenTouch}
& \multicolumn{3}{c}{PVDB}
& \multicolumn{3}{c}{HandSense} \\
\cmidrule(lr){2-4}
\cmidrule(lr){5-7}
\cmidrule(lr){8-10}
& nMAE $\downarrow$ & nRMSE $\downarrow$ & $\mathrm{t}\text{-}\rho$ $\uparrow$
& nMAE $\downarrow$ & nRMSE $\downarrow$ & $\mathrm{t}\text{-}\rho$ $\uparrow$
& nMAE $\downarrow$ & nRMSE $\downarrow$ & $\mathrm{t}\text{-}\rho$ $\uparrow$ \\
\midrule
PressureVision
& \rkthree{0.363} & 1.099 & $-$0.001
& \rkthree{0.130} & \rkthree{0.919} & \rkthree{0.105}
& 0.738 & 1.210 & 0.026 \\
PressureVision++
& 0.364 & 1.102 & 0.000
& 0.136 & 0.967 & 0.038
& 0.746 & 1.219 & 0.007 \\
PressureFormer
& 0.364 & \rkthree{0.939} & \rkone{0.035}
& 0.137 & 0.981 & 0.031
& \rkthree{0.590} & \rkthree{0.914} & \rkthree{0.077} \\
HOPE
& 0.366 & 0.974 & 0.021
& \rkthree{0.130} & 0.951 & 0.067
& 0.614 & 0.922 & 0.069 \\
\midrule
\textbf{Ours (SSL)}
& \rktwo{0.326} & \rktwo{0.914} & \rkthree{0.030}
& \rktwo{0.129} & \rktwo{0.893} & \rktwo{0.239}
& \rktwo{0.495} & \rktwo{0.781} & \rktwo{0.152} \\
\textbf{Ours (Joint)}
& \rkone{0.314} & \rkone{0.879} & \rkone{0.035}
& \rkone{0.123} & \rkone{0.872} & \rkone{0.249}
& \rkone{0.461} & \rkone{0.740} & \rkone{0.180} \\
\midrule
\textit{Static prior}
& 0.398 & 0.898 & N/A
& 0.221 & 0.955 & N/A
& 0.755 & 1.047 & N/A \\
\bottomrule
\end{tabular}%
}
\endgroup

\vspace{-12pt}
\end{table*}

\subsection{Experimental Results}
\label{sex:exp_contact}

\begin{figure}[t]
\centering
\begin{minipage}[t]{0.53\textwidth}
\vspace{0pt}
\centering
\captionof{table}{
Effectiveness of individual components.
SSL variants after half of the $12{,}000$ updates,
averaged over the three test sets on all valid entries.
}
\label{tab:component_ablation}

\begingroup
\setlength{\tabcolsep}{4pt}
\renewcommand{\arraystretch}{1.5}
\small
\resizebox{\linewidth}{!}{%
\begin{tabular}{@{}lccc|ccc@{}}
\toprule
& \multicolumn{3}{c|}{Contact}
& \multicolumn{3}{c}{Magnitude} \\
\cmidrule(lr){2-4}
\cmidrule(lr){5-7}
Variant
& AUROC $\uparrow$
& F1 $\uparrow$
& tAUC $\uparrow$
& nMAE $\downarrow$
& nRMSE $\downarrow$
& $\mathrm{t}\text{-}\rho$ $\uparrow$ \\
\midrule
\textbf{Full model (SSL)}
& \rkone{0.892}
& \rkone{0.613}
& \rkone{0.674}
& \rkone{0.312}
& \rkone{0.853}
& \rkone{0.130} \\
\midrule
w/o anatomical masking
& \rktwo{0.883}
& \rktwo{0.605}
& \rkthree{0.663}
& \rktwo{0.323}
& \rktwo{0.870}
& \rktwo{0.121} \\
w/o interaction masking
& \rkthree{0.871}
& \rkthree{0.599}
& 0.652
& \rktwo{0.323}
& \rkthree{0.883}
& 0.094 \\
w/o geometry encoder
& 0.815
& 0.527
& 0.651
& 0.345
& 0.937
& \rkthree{0.111} \\
w/o video (geometry only)
& 0.811
& 0.430
& 0.611
& 0.350
& 0.946
& 0.069 \\
V-JEPA~2.1 (frozen)
& 0.832
& 0.558
& \rktwo{0.664}
& 0.337
& 0.918
& 0.102 \\
\bottomrule
\end{tabular}%
}
\endgroup

\end{minipage}\hfill%
\begin{minipage}[t]{0.45\textwidth}
\vspace{0pt}
\centering
\includegraphics[width=\linewidth]{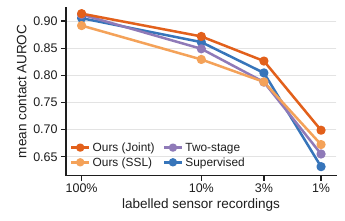}
\captionsetup{font=footnotesize,skip=4pt}
\captionof{figure}{Scaling annotated sensor recordings. Mean contact AUROC over test sets.}
\label{fig:label_efficiency}
\end{minipage}
\vspace{-6pt}
\end{figure}

\begin{figure}[t]
\centering
\begin{minipage}[t]{0.49\textwidth}
\vspace{0pt}
\centering
\captionof{table}{
Zero-shot transfer to Allegro. AP is average precision
and $\rho$ per-episode Spearman.
}
\label{tab:r2r_allegro}

\begingroup
\setlength{\tabcolsep}{6.8pt}
\renewcommand{\arraystretch}{1.30}
\small
\resizebox{\linewidth}{!}{%
\begin{tabular}{@{}lcccc@{}}
\toprule
Method
& AUROC $\uparrow$
& AP $\uparrow$
& F1 $\uparrow$
& $\rho$ $\uparrow$ \\
\midrule
PressureVision
& 0.506 & 0.443 & 0.474 & $-0.032$ \\
PressureVision++
& 0.435 & 0.395 & 0.429 & $-0.039$ \\
PressureFormer
& 0.537 & 0.485 & 0.502 & $-0.017$ \\
HOPE
& \rkthree{0.582}
& \rkthree{0.510}
& \rkthree{0.535}
& \rktwo{0.159} \\
\midrule
\textbf{Ours (SSL)}
& \rkone{0.676}
& \rkone{0.608}
& \rkone{0.608}
& \rkone{0.183} \\
\textbf{Ours (Joint)}
& \rktwo{0.628}
& \rktwo{0.529}
& \rktwo{0.569}
& \rkthree{0.106} \\
\bottomrule
\end{tabular}%

}
\endgroup
\end{minipage}\hfill%
\begin{minipage}[t]{0.49\textwidth}
\vspace{0pt}
\centering
\begingroup
\setlength{\tabcolsep}{1.5pt}
\renewcommand{\arraystretch}{0}
\edef\robotviswidth{\the\dimexpr\linewidth-10pt-2\tabcolsep\relax}
\begin{tabular}{@{}>{\centering\arraybackslash}m{10pt}>{\centering\arraybackslash}m{\robotviswidth}@{}}
\rotatebox{90}{\tiny GT} & \includegraphics[width=\linewidth]{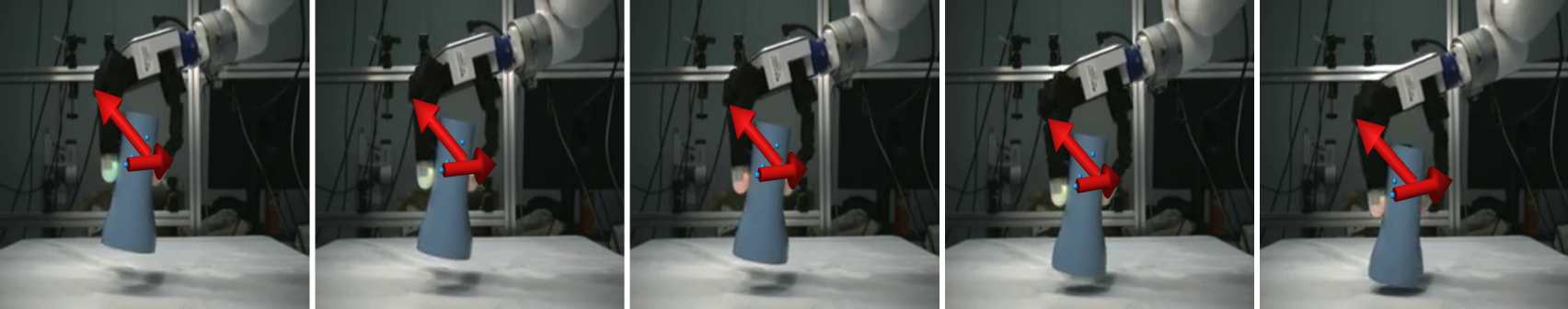} \\[1pt]
\rotatebox{90}{\tiny \textbf{Ours (SSL)}} & \includegraphics[width=\linewidth]{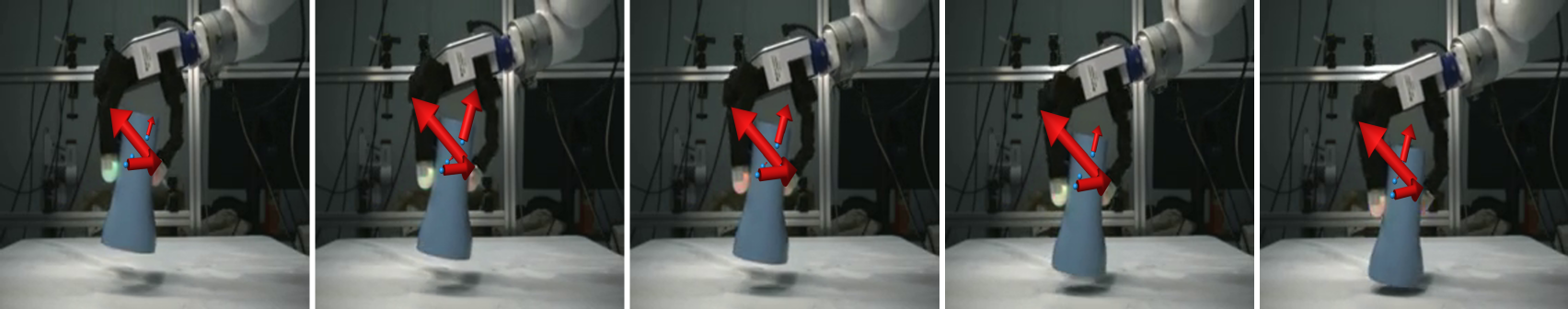} \\[1pt]
\rotatebox{90}{\tiny HOPE} & \includegraphics[width=\linewidth]{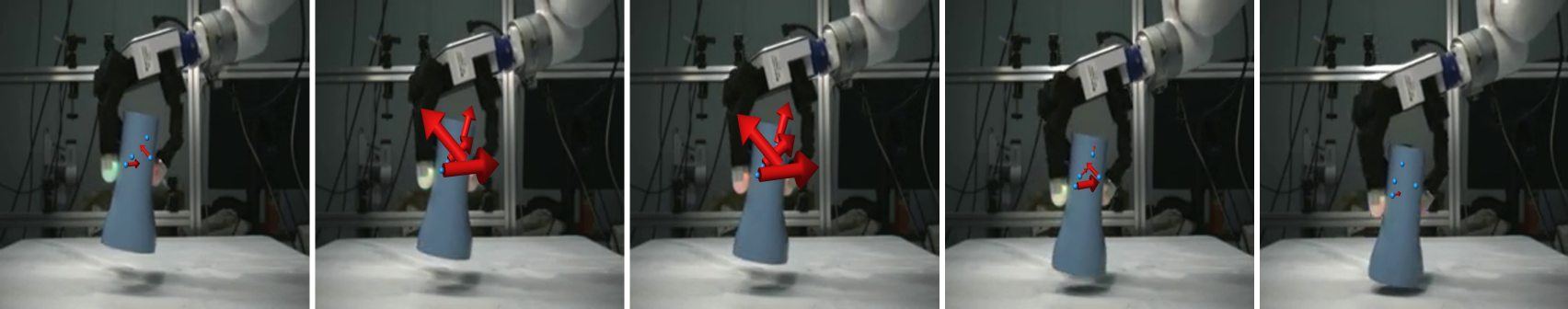} \\
\end{tabular}
\endgroup
\captionsetup{skip=4pt}
\captionof{figure}{
Comparison on the Allegro hand.
}
\label{fig:robot_qualitative}
\end{minipage}
\end{figure}

\noindent\textbf{Contact Estimation.}
Without tactile annotations during encoder pretraining,
Ours (SSL) outperforms all four retrained supervised
baselines in pooled AUROC and F1 across all three
datasets using supervised readouts on frozen encoders
(Table~\ref{tab:contact_comparison}).
Hand-surface masked visual prediction therefore
learns contact-relevant features.
Temporal performance varies across datasets.
On PVDB, SSL achieves a tAUC of $0.783$
against the strongest baseline's $0.748$, and it is on par
with PressureFormer~\citep{zhao2025egopressure} on HandSense.
Among compared methods, joint training is best in AUROC,
F1, and tAUC on every dataset, including a tAUC increase over SSL from
$0.699$ to $0.752$ on HandSense.
On OpenTouch, tAUC stays near chance for every method,
and the video-free static prior is $0.500$ by construction,
so pooled discrimination alone does not measure
contact timing.

\noindent\textbf{Tactile Magnitude Estimation.}
Readouts on frozen Ours (SSL) features achieve lower
nRMSE than all four retrained supervised baselines across the
three datasets and lower nMAE on OpenTouch and HandSense, on par
with the best baseline on PVDB (Table~\ref{tab:force_comparison}
and Figure~\ref{fig:main-vis}).
On PVDB, within-contact temporal correlation reaches $0.239$,
compared with $0.105$ for the strongest supervised baseline,
and increases to $0.249$ with joint training.
Joint training achieves the best results across all metrics
in this comparison.
The largest gains from joint training occur on HandSense,
where nMAE decreases from $0.495$ to $0.461$ relative to SSL
and t-$\rho$ rises from $0.152$ to $0.180$.
Visual prediction alone thus improves on the baselines in sensor-native
magnitude estimation, and tactile supervision improves it further.
Temporal correlations nevertheless remain low on
OpenTouch and HandSense, so fine-grained tracking is still challenging.

\noindent\textbf{Ablation Studies.}
\label{sec:ablation}
Each variant is retrained without the named component.
The two controls apply the same readout to per-vertex hand
kinematics and to frozen V-JEPA~2.1 features, respectively.
Table~\ref{tab:component_ablation} shows that the full SSL
model achieves the best results across all six
dataset-averaged metrics.
Compared with frozen V-JEPA~2.1, AUROC improves by $0.060$.
Removing the Geometry Encoder lowers AUROC from
$0.892$ to $0.815$, below frozen V-JEPA~2.1. Using geometry alone reduces it
to $0.811$, so visual evidence adds to hand geometry.
Both masking strategies improve contact and magnitude
estimation.
Combining hand geometry with masked visual prediction
therefore yields tactile-relevant representations
beyond frozen video features.

\noindent\textbf{Label Efficiency.}
With labels on $1$ to $10\%$ of the sensor recordings, a joint model and
a Supervised model would otherwise see each labeled clip
a different number of times.
We therefore oversample the labeled clips identically in both arms,
so that each labeled clip is drawn equally often, and fit the
readouts with the same labeled fraction. At $100\%$, every sensor clip is
labeled and the exposure is equal by construction.
Joint training helps most when labels
are scarce (Figure~\ref{fig:label_efficiency}).
With labels on $1\%$ of the recordings, it raises contact AUROC
over Supervised by $0.170$ on PVDB and by $0.046$
on OpenTouch, and the gain is small once every clip is labeled.
On HandSense, the two do not differ reliably at $1$ to $10\%$,
and joint training is ahead at $100\%$.
For Ours (SSL), the frozen encoder is read out with the same labels,
and the two-stage control further trains Ours (SSL) on the labeled
clips alone.
Averaged over the test sets, joint training stays above Ours (SSL),
the two-stage control, and Supervised at every fraction up to $10\%$.

\subsection{Robotic Transfer and Manipulation}
\label{sec:robot_evaluation}

\noindent\textbf{Transfer to Robotic Hands.}
We evaluate human-trained predictors without adaptation
on $89$ real Allegro episodes ($10{,}562$ frames), with hand geometry
from measured joint angles.
Ours (SSL) leads all four metrics and outperforms
the joint model (Table~\ref{tab:r2r_allegro}), with examples in Figure~\ref{fig:robot_qualitative}.
Its AUROC and magnitude correlation are $0.676$ and $0.183$,
compared with $0.628$ and $0.106$ for joint training.
The joint model still beats the supervised baselines
in Allegro contact estimation.
On the five-fingertip Inspire hand, Ours (SSL) also leads all four
metrics (Appendix~\ref{sec:supp-allegro}).
We therefore use the SSL encoder for planning, so that, like
the V-JEPA~2.1 baseline, it is label-free.

\begin{figure}[t]
\centering
\begingroup

\begin{minipage}[t]{0.637\linewidth}
\vspace{0pt}
\centering
\setlength{\tabcolsep}{1pt}
\renewcommand{\arraystretch}{0}

\edef\rolloutwidth{%
    \the\dimexpr(\linewidth-12pt-10\tabcolsep)/5\relax
}

\begin{tabular}{@{}
    >{\centering\arraybackslash}m{12pt}
    *{5}{>{\centering\arraybackslash}m{\rolloutwidth}}
@{}}
& {\scriptsize Start}
& {\scriptsize Pre-grasp}
& {\scriptsize Grasp}
& {\scriptsize Lift / Carry}
& {\scriptsize End} \\[3pt]

\rotatebox{90}{\scriptsize\textbf{Grasp}}
& \includegraphics[width=\linewidth]{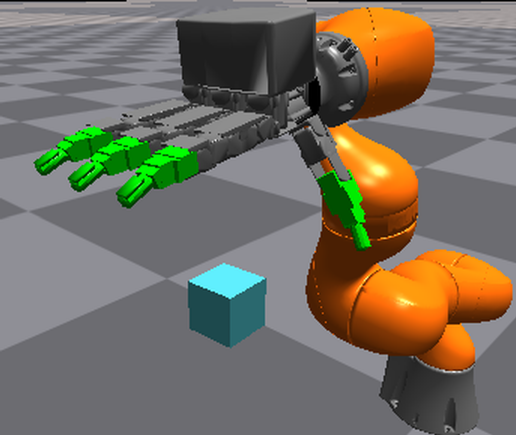}
& \includegraphics[width=\linewidth]{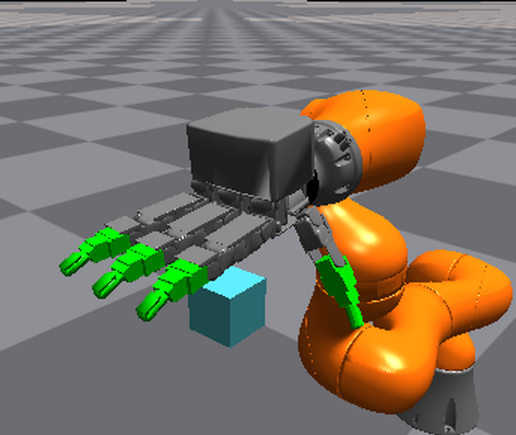}
& \includegraphics[width=\linewidth]{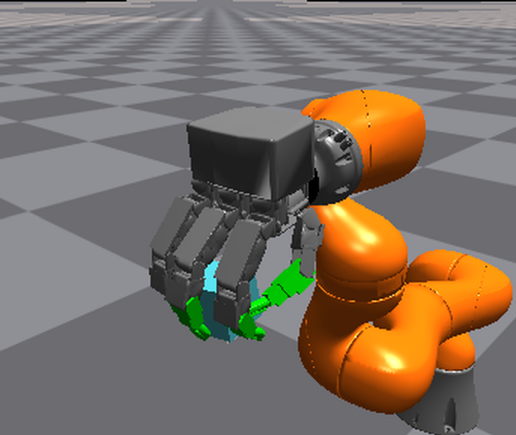}
& \includegraphics[width=\linewidth]{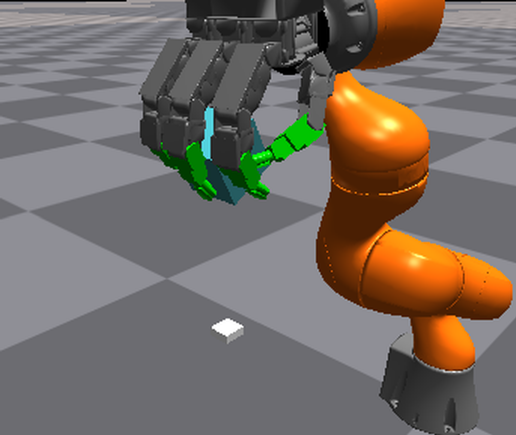}
& \includegraphics[width=\linewidth]{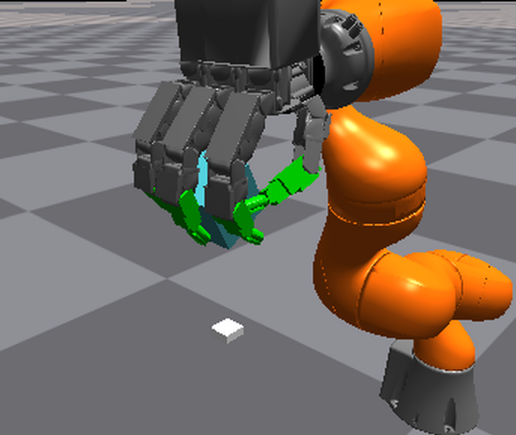} \\[1pt]

\rotatebox{90}{\scriptsize\textbf{Place}}
& \includegraphics[width=\linewidth]{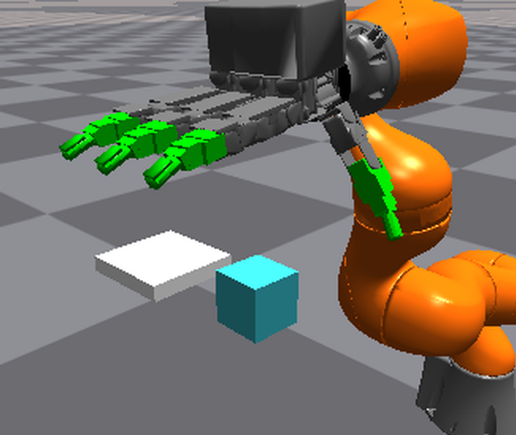}
& \includegraphics[width=\linewidth]{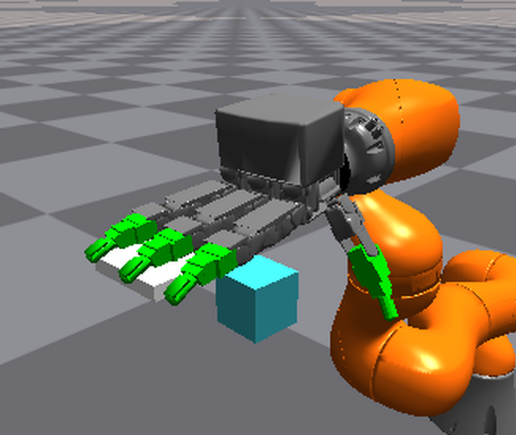}
& \includegraphics[width=\linewidth]{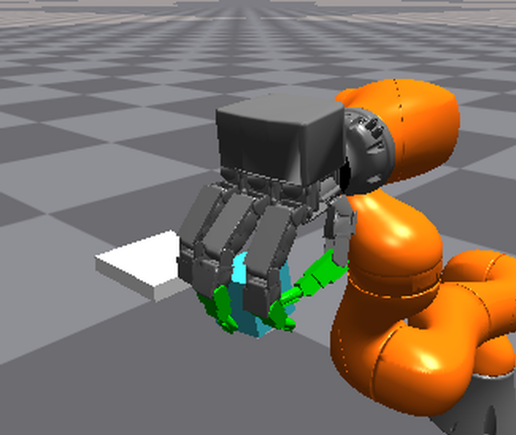}
& \includegraphics[width=\linewidth]{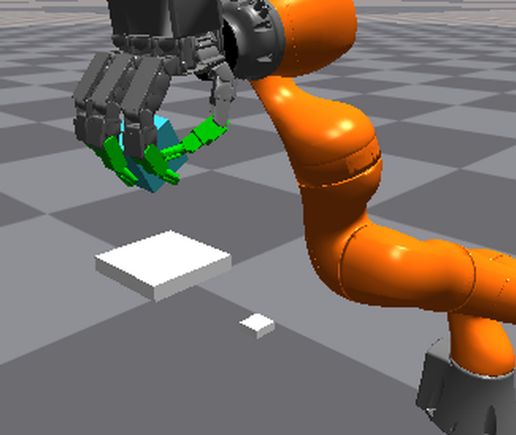}
& \includegraphics[width=\linewidth]{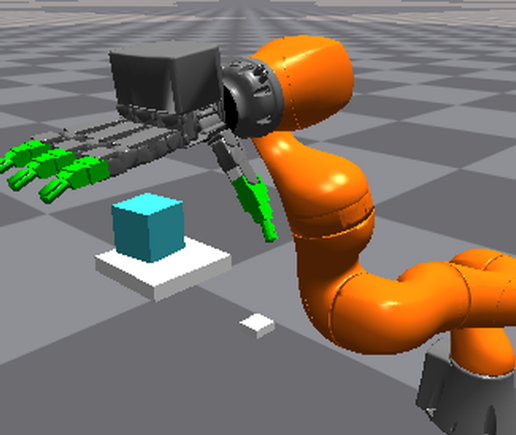} \\
\end{tabular}

\captionsetup{font=footnotesize,skip=4pt}
\captionof{figure}{
Robot manipulation rollouts.
Grasp-lift and place with time progressing from left to right.
}
\label{fig:robot_rollouts}
\end{minipage}\hfill%
\begin{minipage}[t]{0.343\linewidth}
\vspace{0pt}
\centering
\includegraphics[width=\linewidth]
{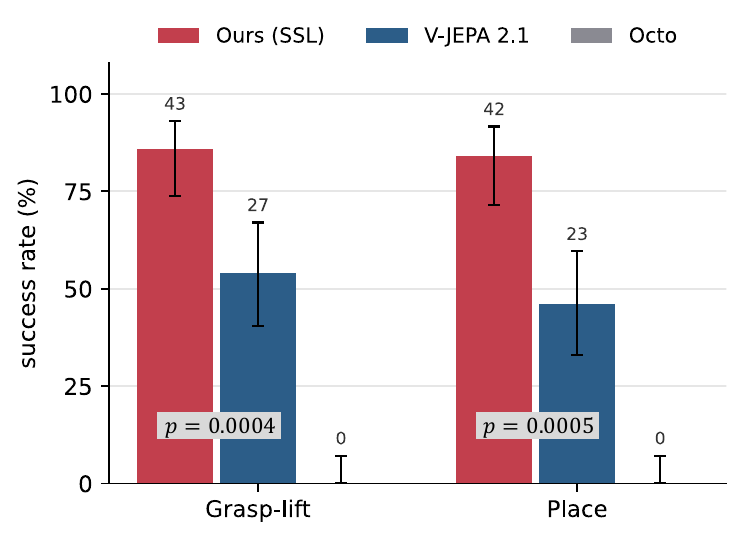}

\captionsetup{font=footnotesize,skip=4pt}
\captionof{figure}{
Task success rates.
}
\label{fig:robot_success}
\end{minipage}

\endgroup
\vspace{-1.5em}

\end{figure}

\noindent\textbf{Action-Conditioned Planning for Robotic Manipulation.}
\label{sec:planning}
Using the frozen SSL encoder, Touch-JEPA-AC plans simulated
grasp-lift and place actions toward reference subgoals
with an action-conditioned predictor trained on robot
trajectories (Appendix~\ref{sec:ac_details}).
Across $50$ matched scenes per task, success rates are
$86\%$ for grasp-lift and $84\%$ for place, compared with
$54\%$ and $46\%$ for V-JEPA~2.1.
Exact two-sided McNemar tests give unadjusted
$p=0.0004$ and $p=0.0005$, respectively.
Octo~\citep{team2024octo} succeeds in neither task (Figure~\ref{fig:robot_success}).
The representations thus also serve simulated manipulation,
as the sample rollouts in Figure~\ref{fig:robot_rollouts} illustrate.

\section{Conclusion}
\label{sec:conclusion}

We presented Touch-JEPA, a unified framework that combines self-supervised learning of tactile-relevant representations from hand-object video with tactile supervision on a labeled subset. Geometry-guided masked visual prediction enables contact and sensor-native magnitude estimation through supervised readouts on frozen features. Joint training further improves human-hand estimation, with particular benefits under limited tactile supervision. The self-supervised representations also transfer to robotic hands and support simulated manipulation, demonstrating their utility beyond tactile estimation.

\noindent\textbf{Limitations and Future Work.}
Magnitude results are sensor-specific and do not establish cross-sensor force calibration. Within-contact temporal correlations remain low on some datasets, where limited tactile variation within \mbox{16-frame} clips constrains the evaluation of fine-grained temporal tracking. Future work includes pretraining on more diverse and longer videos, improving robustness to geometry errors, and evaluating manipulation across broader tasks and on physical robots. Additionally, the joint-embedding framework could be extended to incorporate language or sensor inputs during representation learning.

\providecommand{\bibfont}{}\renewcommand{\bibfont}{\small}
\bibliography{references}
\bibliographystyle{arxiv}

\newpage
\appendix
\raggedbottom

\section*{Supplementary Material}

\begingroup
\setlength{\tabcolsep}{4pt}
\renewcommand{\arraystretch}{1.12}
\section{Data and Training Protocol}
\label{sec:supp-training}

\paragraph{Video corpus and tactile coverage.}
Table~\ref{tab:supp-data} summarizes the $110{,}147$ training
clips used for encoder learning. Each clip contains $T=16$
frames. Tactile measurements are available for $30{,}329$
clips, or $27.5\%$ of the corpus, from three sensor datasets.
The remaining $79{,}818$ clips provide visual self-supervision.
Clip counts refer to temporal windows. OpenTouch windows
overlap, and HRDexDB-Human windows are counted separately
for each camera view. These counts therefore describe
the training roster rather than unique frames.

\begin{table}[!htbp]
\centering
\caption{Encoder-training corpus.
Counts are $16$-frame training clips.
DexYCB, HO3D, HO-Cap, and SHOWMe form the hand-object
interaction (HOI) group and share one sampling lane.
Tactile measurements supervise encoder updates only
in joint and Supervised training.}
\label{tab:supp-data}
\begingroup
\small
\setlength{\tabcolsep}{5pt}
\renewcommand{\arraystretch}{1.08}
\begin{tabular}{@{}lrl@{}}
\toprule
Dataset & \# Clips & Available tactile measurements \\
\midrule
HandSense & $3{,}721$ & Five fingertip channels in glove counts \\
PressureVision DB & $3{,}970$ & Surface pressure in nominal kPa \\
OpenTouch & $22{,}638$ & $169$ taxels in sensor counts \\
\midrule
HRDexDB-Human & $40{,}536$ & None \\
DexYCB & $17{,}200$ & None used \\
HO3D & $4{,}058$ & None used \\
HO-Cap & $15{,}346$ & None used \\
SHOWMe & $2{,}678$ & None used \\
\midrule
Total & $110{,}147$ & $30{,}329$ clips with tactile measurements \\
\bottomrule
\end{tabular}
\endgroup
\end{table}

\paragraph{Partitions and sensor support.}
PressureVision DB uses its participant-based training and
test folds, with $24$ and $6$ participants, respectively.
HandSense and OpenTouch use recording-disjoint partitions.
HandSense contains $161$ training and $18$ test recordings,
while OpenTouch contains $970$ and $123$.
The corresponding test sets contain $447$, $1{,}224$, and
$2{,}519$ clips for HandSense, PressureVision DB, and
OpenTouch. HandSense uses the released synthesized
bare-hand videos with binary contact annotations and glove
readings.
PressureVision DB provides pressure on palmar vertices
within the sensing pad and facing the pad.
OpenTouch retains the $169$ supported taxels and uses
$3072$ minus the released sensor value as magnitude.
Sensor validity determines supervised entries independently
of image visibility. HRDexDB contributes only its human videos to encoder training.

\paragraph{Matched training schedules.}
SSL, Supervised, and joint training share the encoder
initialization, sampling configuration, and seed $912$.
Each processes $384{,}000$ clip presentations through
$12{,}000$ updates with global batch size $32$.
The first $1{,}200$ updates sample only the three sensor
datasets. Each worker receives $3$ HandSense, $3$
PressureVision DB, and $10$ OpenTouch clips per update.
Subsequently, these counts become $1$, $1$, and $3$,
with $4$ HRDexDB-Human and $7$ HOI clips.
In this mixed phase, about $70\%$ of the sampled clips
therefore come from videos without tactile measurements.
Within each lane, sampling uses shuffled permutations
with disjoint worker shards.
An evaluation interval spans $600$ updates or $19{,}200$
clip presentations.

\paragraph{Optimization and supervision.}
Inputs have resolution $384\times384$.
Training selects the observer view or a hand-centered crop
with equal probability. Crop expansion is $1.5$ times a
uniform multiplier in $[0.8,1.2]$, with independent center
shifts of up to $10\%$ of the crop side along each axis.
AdamW uses $\beta=(0.9,0.95)$ and weight decay $0.04$,
with learning rates $10^{-4}$ for LoRA and $3\times10^{-4}$
for new modules. A $480$-update warmup starts at one sixth
of these rates, followed by cosine decay to $10^{-6}$.
We clip the gradient norm at $1.0$ and increase EMA momentum
from $0.996$ to $0.9995$.
Training uses bfloat16 and gradient checkpointing.
SSL receives no tactile gradient in the encoder.
Detached online probes use a separate optimizer and serve
only as diagnostics. Joint training adds contact and
magnitude supervision from the three sensor datasets,
with $\lambda_{\mathrm{tac}}$ increasing from $0.1$ to $1$
over the first $1{,}200$ updates.
Supervised applies the same supervised losses with unit
weight throughout training and no SSL-relative gradient cap.
Neither supervised setting uses geometric contact labels
from DexYCB or tactile pseudo-labels.

\newcommand{\glab}[1]{\rotatebox[origin=c]{90}{\tiny\shortstack{#1}}}

\begin{figure}[!t]
\centering
\begin{minipage}[t]{0.498\linewidth}
\vspace{0pt}
\centering
{\footnotesize (a) PressureVision DB, palm-and-finger press\par}
\vspace{3pt}
{\setlength{\tabcolsep}{0.5pt}\renewcommand{\arraystretch}{0.4}%
\begin{tabular}{@{}cc@{}}
\glab{RGB} & \raisebox{-0.5\height}{\includegraphics[width=0.955\linewidth]{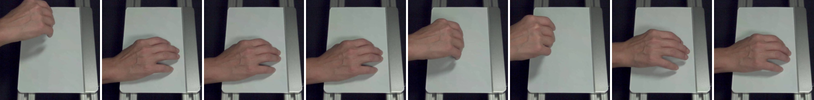}} \\
\glab{GT} & \raisebox{-0.5\height}{\includegraphics[width=0.955\linewidth]{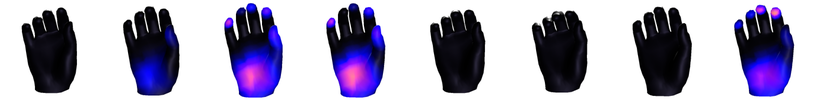}} \\
\glab{\textbf{Ours}\\\textbf{(Joint)}} & \raisebox{-0.5\height}{\includegraphics[width=0.955\linewidth]{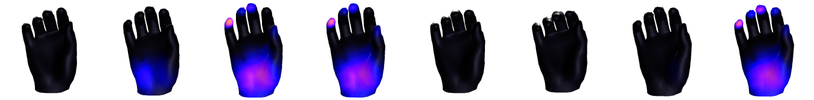}} \\
\glab{\textbf{Ours}\\\textbf{(SSL)}} & \raisebox{-0.5\height}{\includegraphics[width=0.955\linewidth]{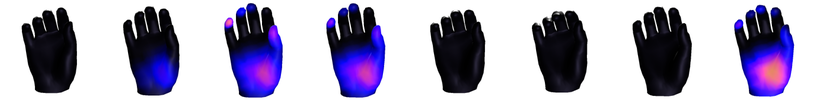}} \\
\glab{HOPE} & \raisebox{-0.5\height}{\includegraphics[width=0.955\linewidth]{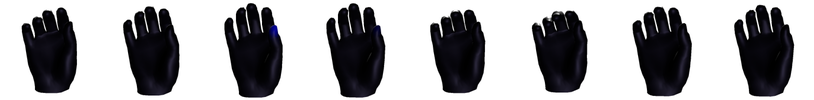}} \\
\glab{Pressure\\Former} & \raisebox{-0.5\height}{\includegraphics[width=0.955\linewidth]{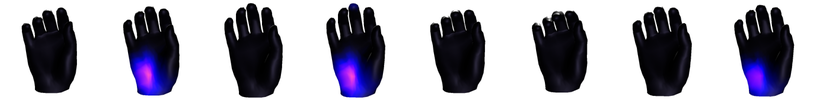}} \\
\glab{Pressure\\Vision++} & \raisebox{-0.5\height}{\includegraphics[width=0.955\linewidth]{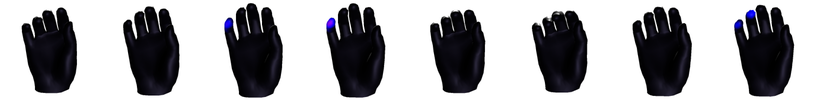}} \\
\glab{Pressure\\Vision} & \raisebox{-0.5\height}{\includegraphics[width=0.955\linewidth]{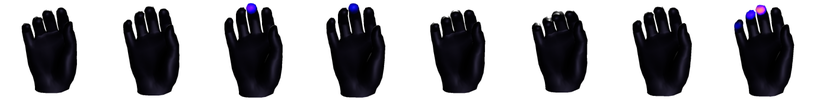}} \\
\end{tabular}}
\end{minipage}\hfill%
\begin{minipage}[t]{0.498\linewidth}
\vspace{0pt}
\centering
{\footnotesize (b) HandSense, piano key press\par}
\vspace{3pt}
{\setlength{\tabcolsep}{0.5pt}\renewcommand{\arraystretch}{0.4}%
\begin{tabular}{@{}cc@{}}
\phantom{\glab{RGB}} & \raisebox{-0.5\height}{\includegraphics[width=0.955\linewidth]{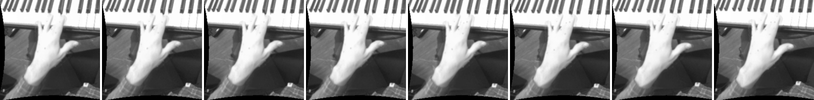}} \\
\phantom{\glab{GT}} & \raisebox{-0.5\height}{\includegraphics[width=0.955\linewidth]{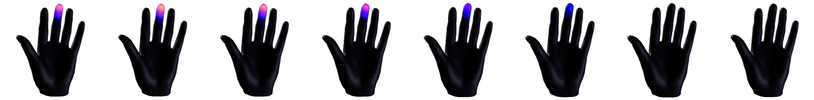}} \\
\phantom{\glab{\textbf{Ours}\\\textbf{(Joint)}}} & \raisebox{-0.5\height}{\includegraphics[width=0.955\linewidth]{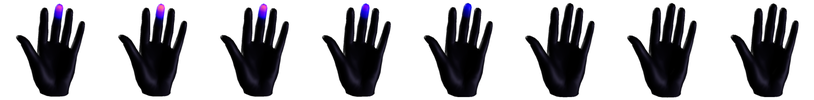}} \\
\phantom{\glab{\textbf{Ours}\\\textbf{(SSL)}}} & \raisebox{-0.5\height}{\includegraphics[width=0.955\linewidth]{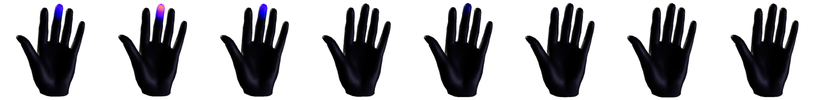}} \\
\phantom{\glab{HOPE}} & \raisebox{-0.5\height}{\includegraphics[width=0.955\linewidth]{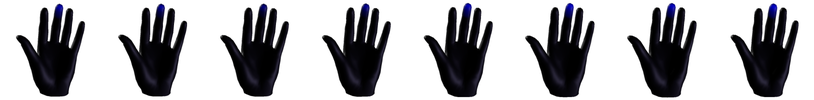}} \\
\phantom{\glab{Pressure\\Former}} & \raisebox{-0.5\height}{\includegraphics[width=0.955\linewidth]{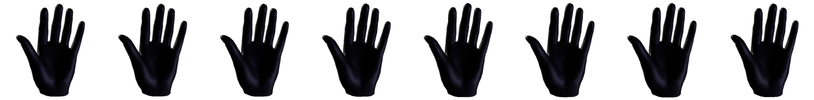}} \\
\phantom{\glab{Pressure\\Vision++}} & \raisebox{-0.5\height}{\includegraphics[width=0.955\linewidth]{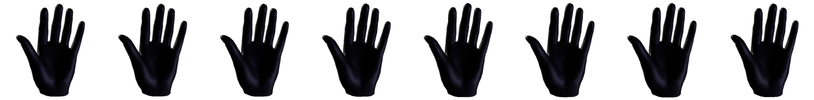}} \\
\phantom{\glab{Pressure\\Vision}} & \raisebox{-0.5\height}{\includegraphics[width=0.955\linewidth]{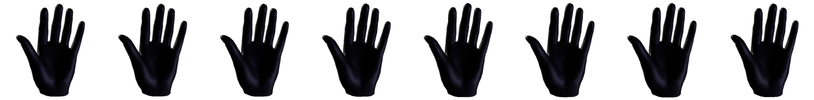}} \\
\end{tabular}}
\end{minipage}
\caption{
Additional comparisons on PressureVision DB and HandSense.
Eight frames per sequence progress from left to right. Brighter colors indicate greater tactile magnitude, and gray marks
entries without a prediction.
}
\label{fig:additional-pvdb-handsense}

\vspace{10pt}
\begin{minipage}[t]{0.498\linewidth}
\vspace{0pt}
\centering
{\footnotesize (a) OpenTouch, fabrication lab\par}
\vspace{3pt}
{\setlength{\tabcolsep}{0.5pt}\renewcommand{\arraystretch}{0.4}%
\begin{tabular}{@{}cc@{}}
\glab{RGB} & \raisebox{-0.5\height}{\includegraphics[width=0.955\linewidth]{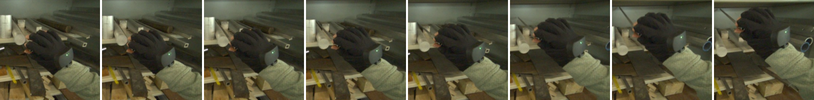}} \\
\glab{GT} & \raisebox{-0.5\height}{\includegraphics[width=0.955\linewidth]{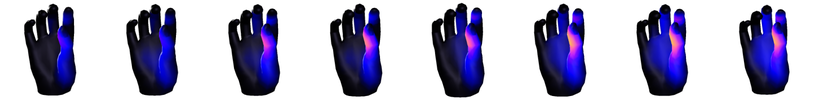}} \\
\glab{\textbf{Ours}\\\textbf{(Joint)}} & \raisebox{-0.5\height}{\includegraphics[width=0.955\linewidth]{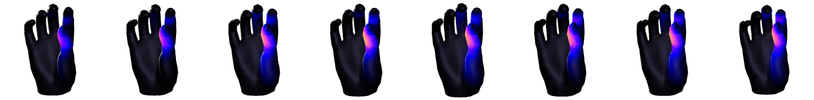}} \\
\glab{\textbf{Ours}\\\textbf{(SSL)}} & \raisebox{-0.5\height}{\includegraphics[width=0.955\linewidth]{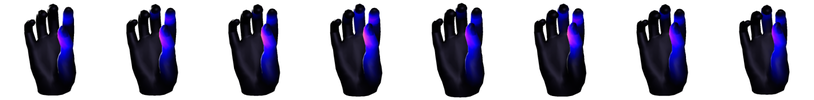}} \\
\glab{HOPE} & \raisebox{-0.5\height}{\includegraphics[width=0.955\linewidth]{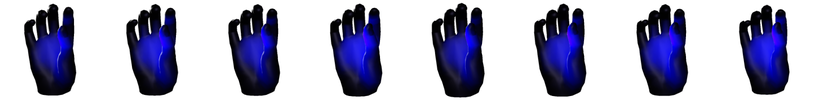}} \\
\glab{Pressure\\Former} & \raisebox{-0.5\height}{\includegraphics[width=0.955\linewidth]{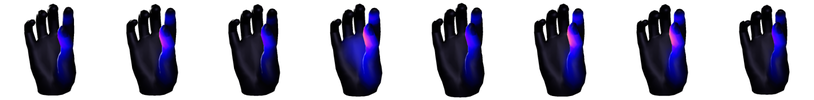}} \\
\glab{Pressure\\Vision++} & \raisebox{-0.5\height}{\includegraphics[width=0.955\linewidth]{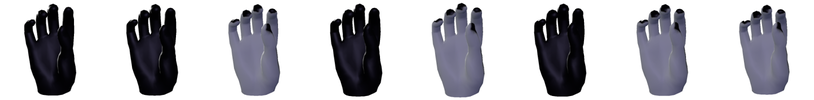}} \\
\glab{Pressure\\Vision} & \raisebox{-0.5\height}{\includegraphics[width=0.955\linewidth]{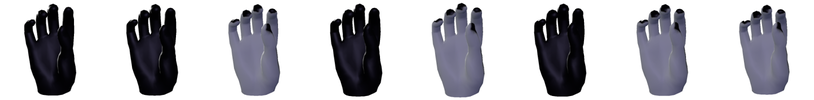}} \\
\end{tabular}}
\end{minipage}\hfill%
\begin{minipage}[t]{0.498\linewidth}
\vspace{0pt}
\centering
{\footnotesize (b) OpenTouch, office\par}
\vspace{3pt}
{\setlength{\tabcolsep}{0.5pt}\renewcommand{\arraystretch}{0.4}%
\begin{tabular}{@{}cc@{}}
\phantom{\glab{RGB}} & \raisebox{-0.5\height}{\includegraphics[width=0.955\linewidth]{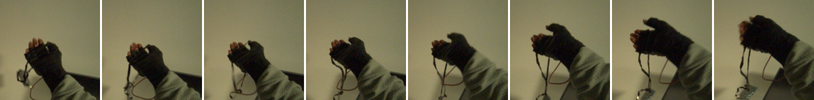}} \\
\phantom{\glab{GT}} & \raisebox{-0.5\height}{\includegraphics[width=0.955\linewidth]{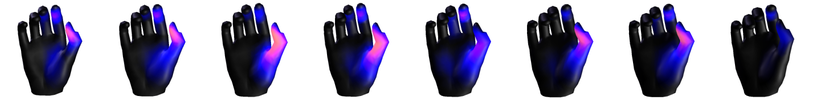}} \\
\phantom{\glab{\textbf{Ours}\\\textbf{(Joint)}}} & \raisebox{-0.5\height}{\includegraphics[width=0.955\linewidth]{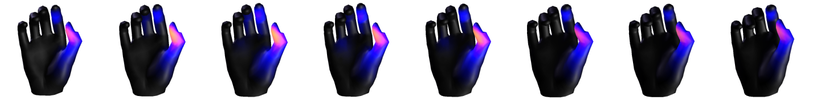}} \\
\phantom{\glab{\textbf{Ours}\\\textbf{(SSL)}}} & \raisebox{-0.5\height}{\includegraphics[width=0.955\linewidth]{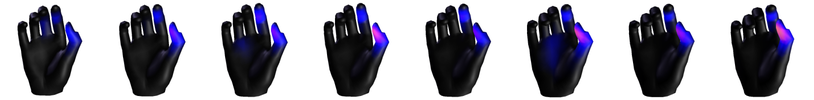}} \\
\phantom{\glab{HOPE}} & \raisebox{-0.5\height}{\includegraphics[width=0.955\linewidth]{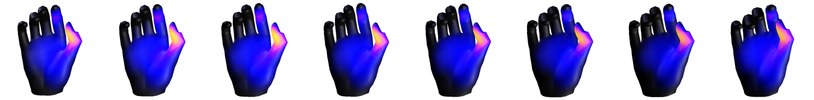}} \\
\phantom{\glab{Pressure\\Former}} & \raisebox{-0.5\height}{\includegraphics[width=0.955\linewidth]{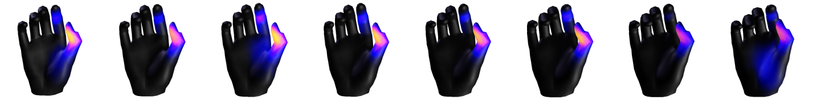}} \\
\phantom{\glab{Pressure\\Vision++}} & \raisebox{-0.5\height}{\includegraphics[width=0.955\linewidth]{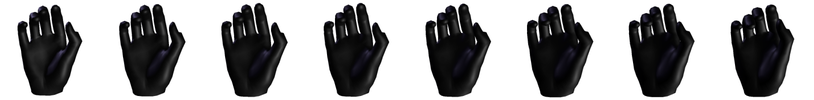}} \\
\phantom{\glab{Pressure\\Vision}} & \raisebox{-0.5\height}{\includegraphics[width=0.955\linewidth]{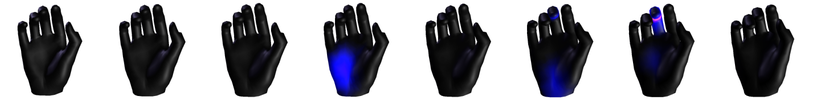}} \\
\end{tabular}}
\end{minipage}
\caption{
Additional comparisons on OpenTouch.
Display conventions follow Figure~\ref{fig:additional-pvdb-handsense}.
}
\label{fig:additional-opentouch}
\end{figure}

\begin{figure}[!t]
\centering
\begin{minipage}[t]{0.498\linewidth}
\vspace{0pt}
\centering
{\footnotesize (a) PressureVision DB, palm press\par}
\vspace{3pt}
{\setlength{\tabcolsep}{0.5pt}\renewcommand{\arraystretch}{0.4}%
\begin{tabular}{@{}cc@{}}
\glab{RGB} & \raisebox{-0.5\height}{\includegraphics[width=0.955\linewidth]{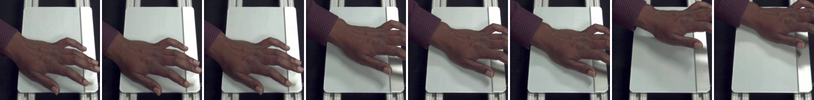}} \\
\glab{GT} & \raisebox{-0.5\height}{\includegraphics[width=0.955\linewidth]{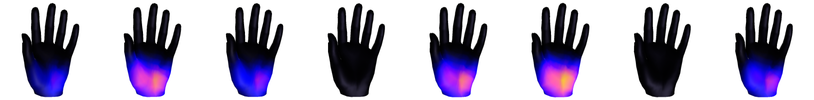}} \\
\glab{\textbf{Ours}\\\textbf{(Joint)}} & \raisebox{-0.5\height}{\includegraphics[width=0.955\linewidth]{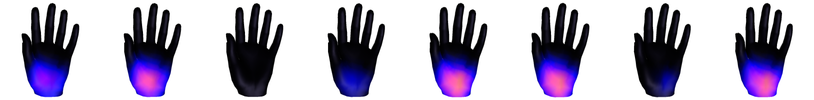}} \\
\glab{\textbf{Ours}\\\textbf{(SSL)}} & \raisebox{-0.5\height}{\includegraphics[width=0.955\linewidth]{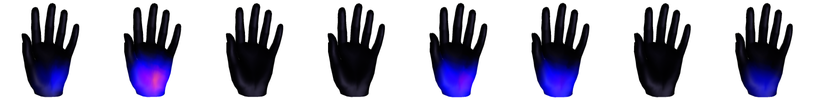}} \\
\glab{HOPE} & \raisebox{-0.5\height}{\includegraphics[width=0.955\linewidth]{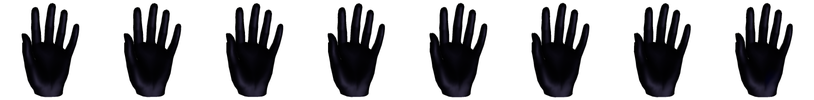}} \\
\glab{Pressure\\Former} & \raisebox{-0.5\height}{\includegraphics[width=0.955\linewidth]{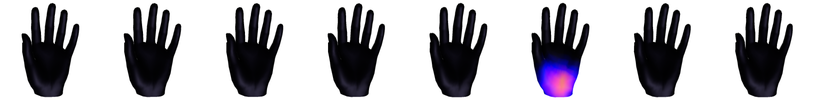}} \\
\glab{Pressure\\Vision++} & \raisebox{-0.5\height}{\includegraphics[width=0.955\linewidth]{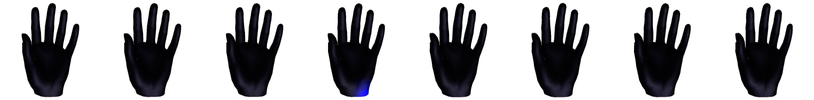}} \\
\glab{Pressure\\Vision} & \raisebox{-0.5\height}{\includegraphics[width=0.955\linewidth]{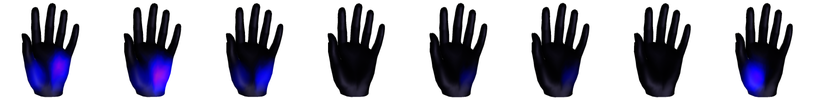}} \\
\end{tabular}}
\end{minipage}\hfill%
\begin{minipage}[t]{0.498\linewidth}
\vspace{0pt}
\centering
{\footnotesize (b) PressureVision DB, cupped-palm press\par}
\vspace{3pt}
{\setlength{\tabcolsep}{0.5pt}\renewcommand{\arraystretch}{0.4}%
\begin{tabular}{@{}cc@{}}
\phantom{\glab{RGB}} & \raisebox{-0.5\height}{\includegraphics[width=0.955\linewidth]{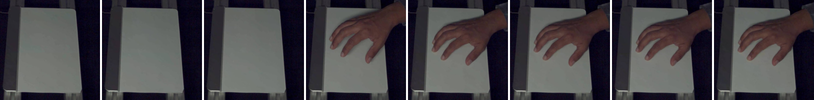}} \\
\phantom{\glab{GT}} & \raisebox{-0.5\height}{\includegraphics[width=0.955\linewidth]{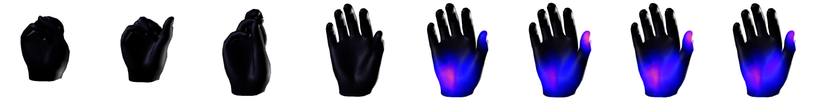}} \\
\phantom{\glab{\textbf{Ours}\\\textbf{(Joint)}}} & \raisebox{-0.5\height}{\includegraphics[width=0.955\linewidth]{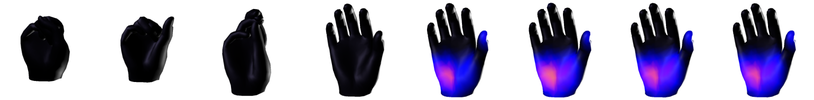}} \\
\phantom{\glab{\textbf{Ours}\\\textbf{(SSL)}}} & \raisebox{-0.5\height}{\includegraphics[width=0.955\linewidth]{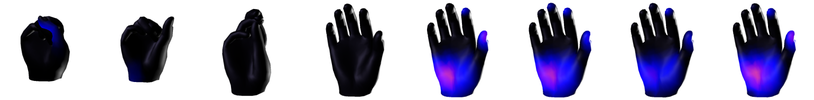}} \\
\phantom{\glab{HOPE}} & \raisebox{-0.5\height}{\includegraphics[width=0.955\linewidth]{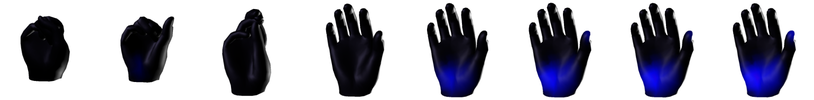}} \\
\phantom{\glab{Pressure\\Former}} & \raisebox{-0.5\height}{\includegraphics[width=0.955\linewidth]{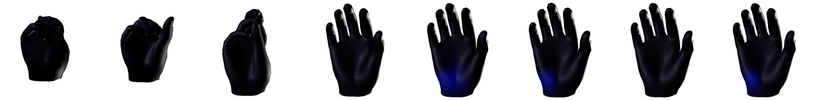}} \\
\phantom{\glab{Pressure\\Vision++}} & \raisebox{-0.5\height}{\includegraphics[width=0.955\linewidth]{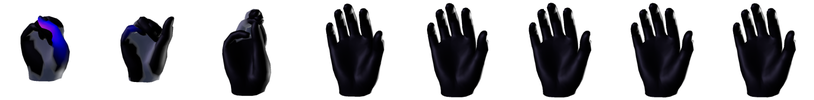}} \\
\phantom{\glab{Pressure\\Vision}} & \raisebox{-0.5\height}{\includegraphics[width=0.955\linewidth]{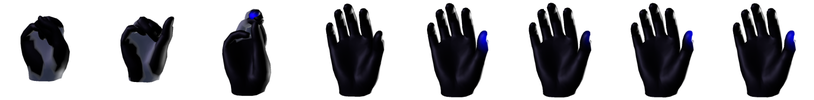}} \\
\end{tabular}}
\end{minipage}
\caption{
More comparisons on PressureVision DB.
Display conventions follow Figure~\ref{fig:additional-pvdb-handsense}.
}
\label{fig:more-pvdb}

\vspace{10pt}
\begin{minipage}[t]{0.498\linewidth}
\vspace{0pt}
\centering
{\footnotesize (a) HandSense, opening a bottle\par}
\vspace{3pt}
{\setlength{\tabcolsep}{0.5pt}\renewcommand{\arraystretch}{0.4}%
\begin{tabular}{@{}cc@{}}
\glab{RGB} & \raisebox{-0.5\height}{\includegraphics[width=0.955\linewidth]{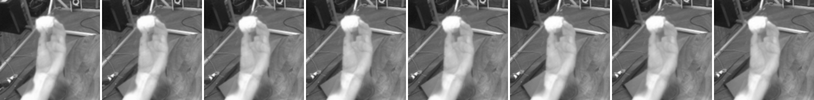}} \\
\glab{GT} & \raisebox{-0.5\height}{\includegraphics[width=0.955\linewidth]{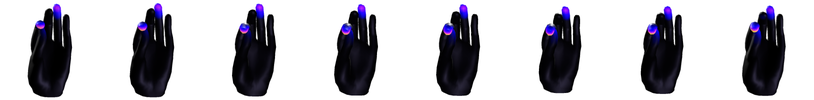}} \\
\glab{\textbf{Ours}\\\textbf{(Joint)}} & \raisebox{-0.5\height}{\includegraphics[width=0.955\linewidth]{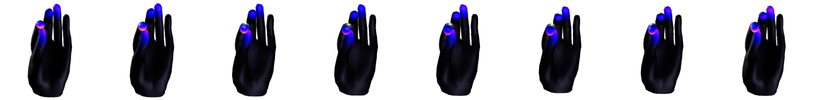}} \\
\glab{\textbf{Ours}\\\textbf{(SSL)}} & \raisebox{-0.5\height}{\includegraphics[width=0.955\linewidth]{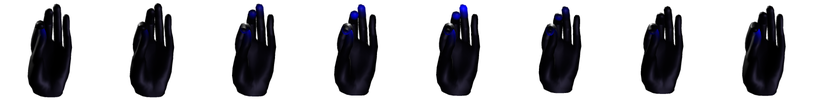}} \\
\glab{HOPE} & \raisebox{-0.5\height}{\includegraphics[width=0.955\linewidth]{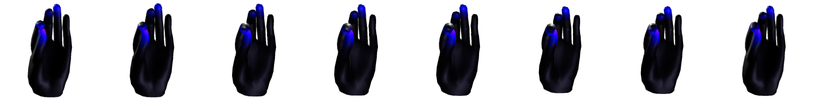}} \\
\glab{Pressure\\Former} & \raisebox{-0.5\height}{\includegraphics[width=0.955\linewidth]{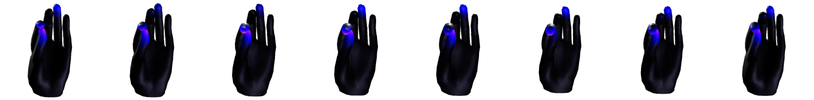}} \\
\glab{Pressure\\Vision++} & \raisebox{-0.5\height}{\includegraphics[width=0.955\linewidth]{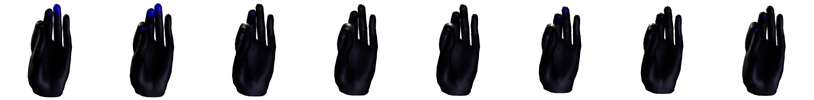}} \\
\glab{Pressure\\Vision} & \raisebox{-0.5\height}{\includegraphics[width=0.955\linewidth]{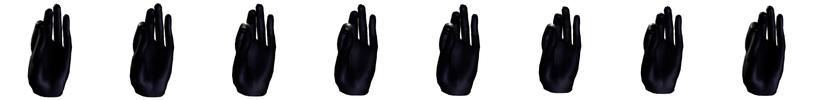}} \\
\end{tabular}}
\end{minipage}\hfill%
\begin{minipage}[t]{0.498\linewidth}
\vspace{0pt}
\centering
{\footnotesize (b) HandSense, opening a bottle (second session)\par}
\vspace{3pt}
{\setlength{\tabcolsep}{0.5pt}\renewcommand{\arraystretch}{0.4}%
\begin{tabular}{@{}cc@{}}
\phantom{\glab{RGB}} & \raisebox{-0.5\height}{\includegraphics[width=0.955\linewidth]{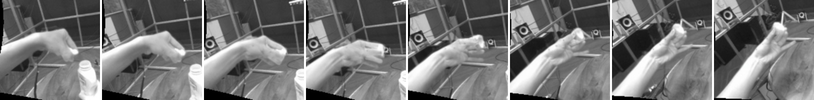}} \\
\phantom{\glab{GT}} & \raisebox{-0.5\height}{\includegraphics[width=0.955\linewidth]{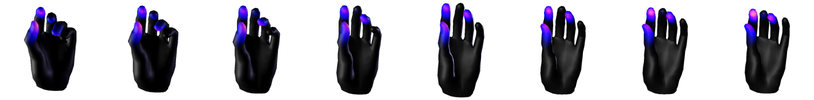}} \\
\phantom{\glab{\textbf{Ours}\\\textbf{(Joint)}}} & \raisebox{-0.5\height}{\includegraphics[width=0.955\linewidth]{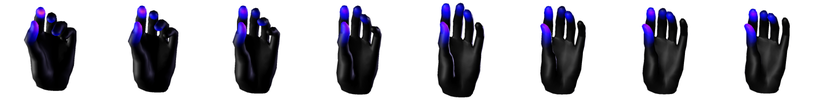}} \\
\phantom{\glab{\textbf{Ours}\\\textbf{(SSL)}}} & \raisebox{-0.5\height}{\includegraphics[width=0.955\linewidth]{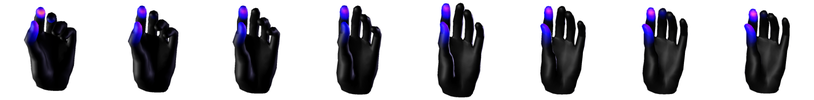}} \\
\phantom{\glab{HOPE}} & \raisebox{-0.5\height}{\includegraphics[width=0.955\linewidth]{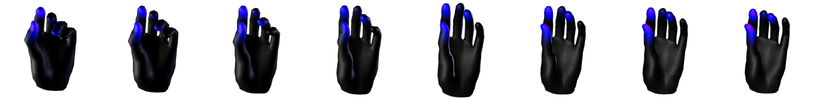}} \\
\phantom{\glab{Pressure\\Former}} & \raisebox{-0.5\height}{\includegraphics[width=0.955\linewidth]{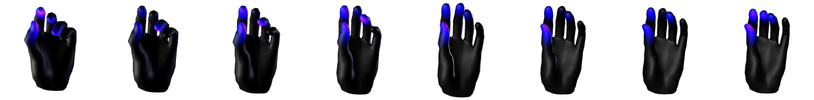}} \\
\phantom{\glab{Pressure\\Vision++}} & \raisebox{-0.5\height}{\includegraphics[width=0.955\linewidth]{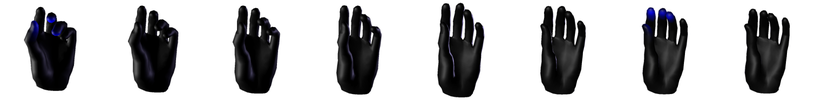}} \\
\phantom{\glab{Pressure\\Vision}} & \raisebox{-0.5\height}{\includegraphics[width=0.955\linewidth]{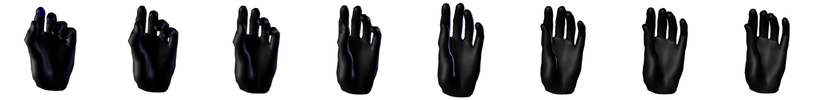}} \\
\end{tabular}}
\end{minipage}
\caption{
More comparisons on HandSense.
Display conventions follow Figure~\ref{fig:additional-pvdb-handsense}.
}
\label{fig:more-handsense}
\end{figure}

\begin{figure}[!t]
\centering
\begin{minipage}[t]{0.498\linewidth}
\vspace{0pt}
\centering
{\footnotesize (a) OpenTouch, grocery store\par}
\vspace{3pt}
{\setlength{\tabcolsep}{0.5pt}\renewcommand{\arraystretch}{0.4}%
\begin{tabular}{@{}cc@{}}
\glab{RGB} & \raisebox{-0.5\height}{\includegraphics[width=0.955\linewidth]{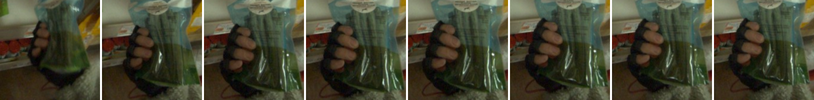}} \\
\glab{GT} & \raisebox{-0.5\height}{\includegraphics[width=0.955\linewidth]{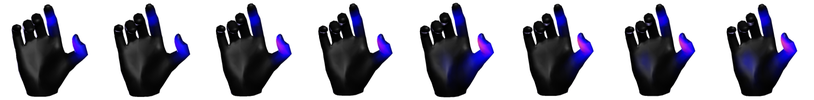}} \\
\glab{\textbf{Ours}\\\textbf{(Joint)}} & \raisebox{-0.5\height}{\includegraphics[width=0.955\linewidth]{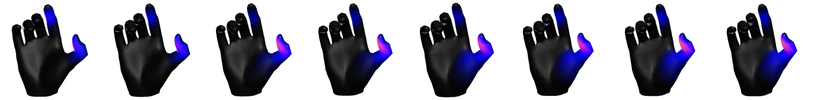}} \\
\glab{\textbf{Ours}\\\textbf{(SSL)}} & \raisebox{-0.5\height}{\includegraphics[width=0.955\linewidth]{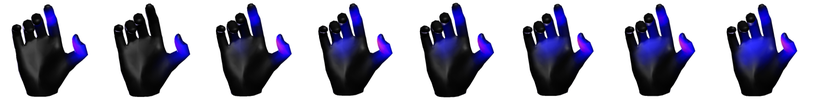}} \\
\glab{HOPE} & \raisebox{-0.5\height}{\includegraphics[width=0.955\linewidth]{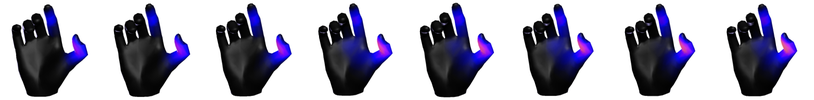}} \\
\glab{Pressure\\Former} & \raisebox{-0.5\height}{\includegraphics[width=0.955\linewidth]{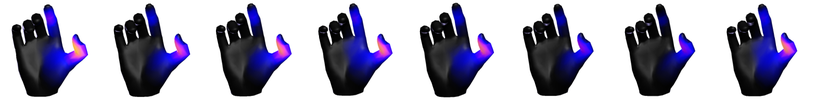}} \\
\glab{Pressure\\Vision++} & \raisebox{-0.5\height}{\includegraphics[width=0.955\linewidth]{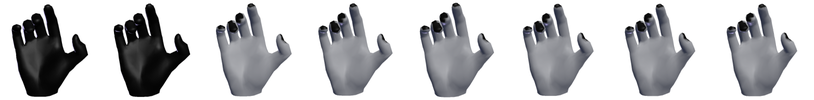}} \\
\glab{Pressure\\Vision} & \raisebox{-0.5\height}{\includegraphics[width=0.955\linewidth]{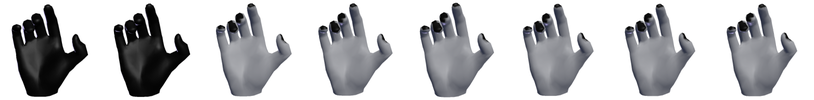}} \\
\end{tabular}}
\end{minipage}\hfill%
\begin{minipage}[t]{0.498\linewidth}
\vspace{0pt}
\centering
{\footnotesize (b) OpenTouch, kitchen\par}
\vspace{3pt}
{\setlength{\tabcolsep}{0.5pt}\renewcommand{\arraystretch}{0.4}%
\begin{tabular}{@{}cc@{}}
\phantom{\glab{RGB}} & \raisebox{-0.5\height}{\includegraphics[width=0.955\linewidth]{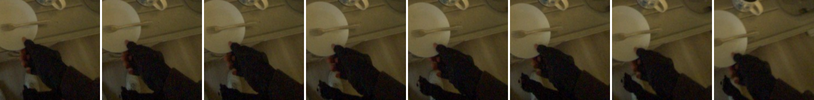}} \\
\phantom{\glab{GT}} & \raisebox{-0.5\height}{\includegraphics[width=0.955\linewidth]{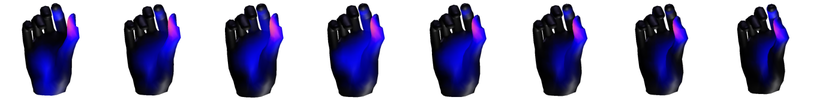}} \\
\phantom{\glab{\textbf{Ours}\\\textbf{(Joint)}}} & \raisebox{-0.5\height}{\includegraphics[width=0.955\linewidth]{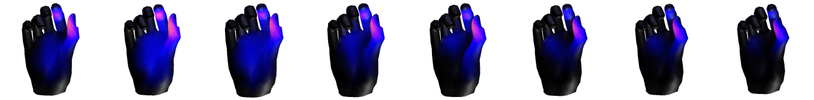}} \\
\phantom{\glab{\textbf{Ours}\\\textbf{(SSL)}}} & \raisebox{-0.5\height}{\includegraphics[width=0.955\linewidth]{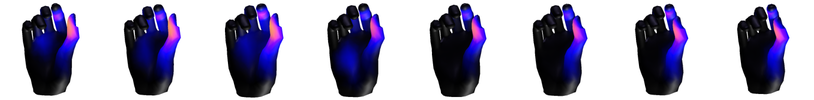}} \\
\phantom{\glab{HOPE}} & \raisebox{-0.5\height}{\includegraphics[width=0.955\linewidth]{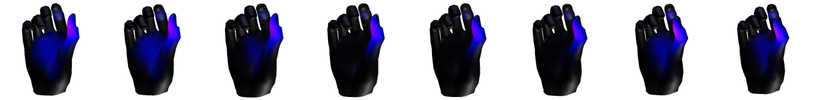}} \\
\phantom{\glab{Pressure\\Former}} & \raisebox{-0.5\height}{\includegraphics[width=0.955\linewidth]{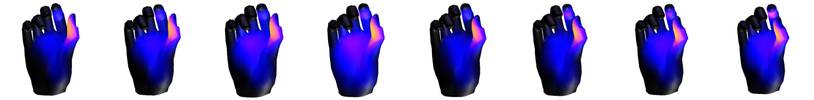}} \\
\phantom{\glab{Pressure\\Vision++}} & \raisebox{-0.5\height}{\includegraphics[width=0.955\linewidth]{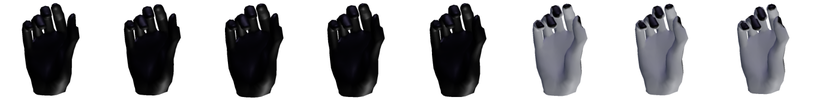}} \\
\phantom{\glab{Pressure\\Vision}} & \raisebox{-0.5\height}{\includegraphics[width=0.955\linewidth]{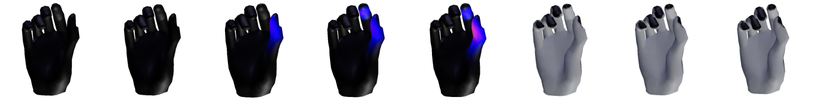}} \\
\end{tabular}}
\end{minipage}
\caption{
More comparisons on OpenTouch.
Display conventions follow Figure~\ref{fig:additional-pvdb-handsense}.
}
\label{fig:more-opentouch}
\end{figure}

\section{Additional Visualizations}
\label{sec:supp-visualizations}

Figures~\ref{fig:additional-pvdb-handsense} to~\ref{fig:more-opentouch} compare tactile magnitude
predictions with ground truth across three datasets, and
Figure~\ref{fig:3d-renderings} shows the examples of Figure~\ref{fig:main-vis} from the video camera.
Figure~\ref{fig:gallery} shows the examples of Figures~\ref{fig:additional-pvdb-handsense} to~\ref{fig:more-opentouch} from the camera and from the palm side.

\section{Architecture and Masking}
\label{sec:supp-architecture}

\subsection{Geometry and Visual Feature Paths}
\label{sec:supp-geometry}

Hand geometry provides correspondence between image observations and a common anatomical surface.
Vertex and joint coordinates are centered at the wrist and divided by the root mean squared vertex distance to the wrist in each frame.
Vertex attributes additionally include surface normals, mesh visibility, and image inclusion.
The anatomical positional encoding sums a vertex embedding, a region embedding, six-frequency Fourier features of the template coordinates, and a linear embedding of normalized frame index.
Joints use joint identity and frame index.
An additional palm frame, defined by the wrist, index metacarpophalangeal joint, and ring metacarpophalangeal joint, supplies canonical coordinates through zero-initialized affine projections.
The same zero-initialized side input also carries the frame's rotation and the wrist position divided by hand size, so the global hand pose remains available ($18$ inputs per vertex and $15$ per joint).
The original camera-oriented coordinates remain available.
Geometry estimation is frozen throughout training.

The backbone has width $D=768$, patch size $16$, and temporal tubelets of two frames.
We number its twelve blocks from $0$ to $11$.
The initial visual field $F_t$ sums features from blocks $5$, $8$, and $11$, each projected to $d_p=128$ dimensions and aligned to the input frames by a learned temporal decoder.
Dense teacher targets instead use the final block's features, projected to $d=256$ by the fixed matrix $R$ in Equation~\ref{eq:target}.
We construct $R$ by QR decomposition of a Gaussian matrix with seed $912$.
The target-field $\operatorname{LayerNorm}$ has no learnable affine parameters.
Each tubelet feature supplies both constituent frames before bilinear sampling.

Table~\ref{tab:supp-information-flow} distinguishes the visual target path from the surface representation path.
The interleaved path branches after block $8$ and reruns blocks $9$ through $11$ with the same backbone and LoRA parameters.
Thus, surface feedback changes the features read by subsequent Geometry Encoder layers without entering the visual features used as prediction targets.

\begin{table}[!ht]
\centering
\caption{Information flow in Touch-JEPA.
The visual and interleaved paths share backbone parameters within each encoder.
The EMA copy supplies targets and the evaluated representation.}
\label{tab:supp-information-flow}
\begingroup
\small
\setlength{\tabcolsep}{4pt}
\renewcommand{\arraystretch}{1.1}
\begin{tabular}{@{}p{0.18\linewidth}p{0.17\linewidth}p{0.17\linewidth}p{0.39\linewidth}@{}}
\toprule
Path & RGB input & Surface feedback & Output and use \\
\midrule
Student visual & Masked & None & Initial field $F_t$ \\
Student surface & Masked & First three layers & States $z$ for prediction and tactile supervision \\
EMA visual & Unmasked & None & Dense and patch targets \\
EMA surface & Unmasked & First three layers & Frozen states for downstream readouts \\
\bottomrule
\end{tabular}
\endgroup
\end{table}

\begin{figure*}[!tb]
\centering
{\setlength{\tabcolsep}{1pt}\renewcommand{\arraystretch}{0.5}%
\begin{tabular}{cccc}
  & {\scriptsize PVDB} & {\scriptsize HandSense} & {\scriptsize OpenTouch} \\[2pt]
  \glab{RGB} & \raisebox{-0.5\height}{\includegraphics[width=0.318\linewidth]{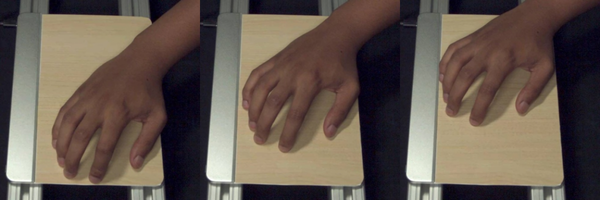}} & \raisebox{-0.5\height}{\includegraphics[width=0.318\linewidth]{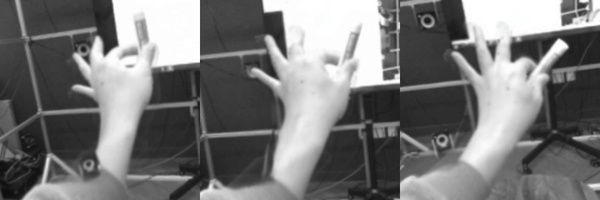}} & \raisebox{-0.5\height}{\includegraphics[width=0.318\linewidth]{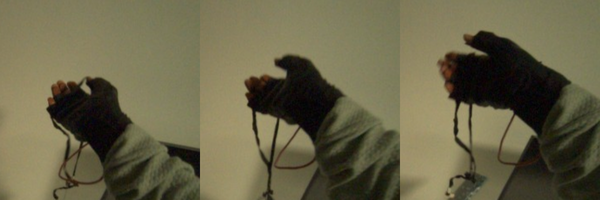}} \\
  \glab{GT} & \raisebox{-0.5\height}{\includegraphics[width=0.318\linewidth]{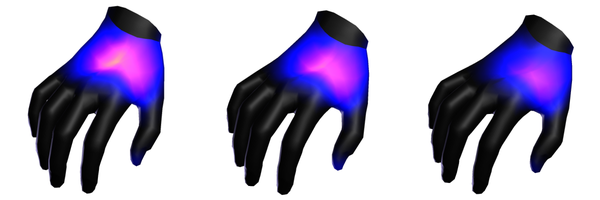}} & \raisebox{-0.5\height}{\includegraphics[width=0.318\linewidth]{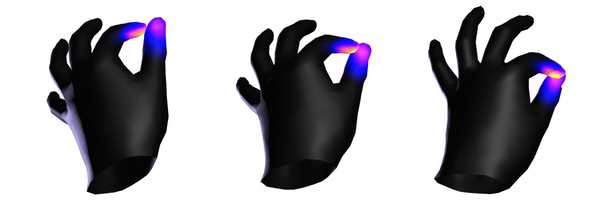}} & \raisebox{-0.5\height}{\includegraphics[width=0.318\linewidth]{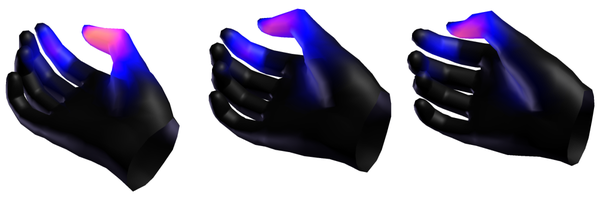}} \\
  \glab{\textbf{Ours (Joint)}} & \raisebox{-0.5\height}{\includegraphics[width=0.318\linewidth]{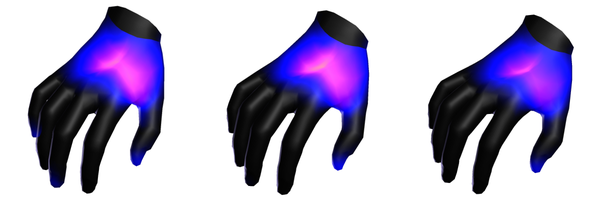}} & \raisebox{-0.5\height}{\includegraphics[width=0.318\linewidth]{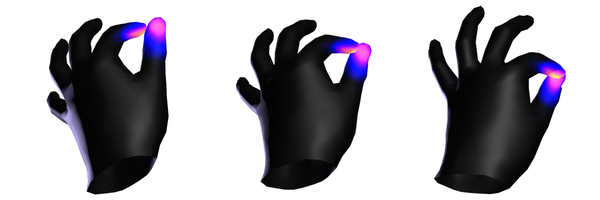}} & \raisebox{-0.5\height}{\includegraphics[width=0.318\linewidth]{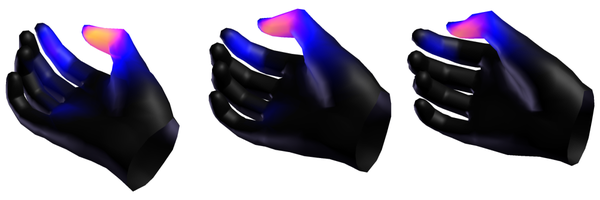}} \\
  \glab{Best comparator} & \raisebox{-0.5\height}{\includegraphics[width=0.318\linewidth]{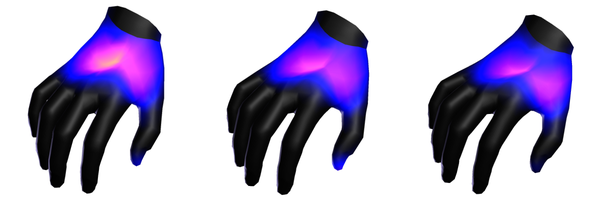}} & \raisebox{-0.5\height}{\includegraphics[width=0.318\linewidth]{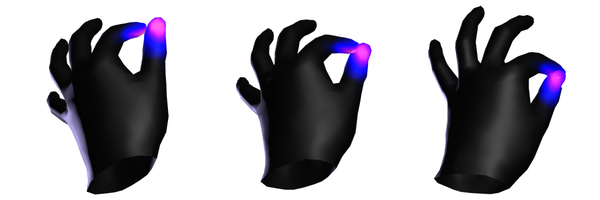}} & \raisebox{-0.5\height}{\includegraphics[width=0.318\linewidth]{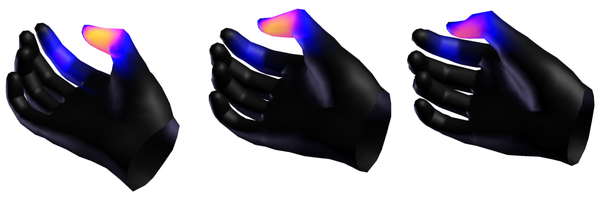}} \\
\end{tabular}}
\caption{
The examples of Figure~\ref{fig:main-vis} seen from the video camera.
Activation on the side facing away from the camera shows through the hand. The comparator is the supervised baseline with the highest contact AUROC in Table~\ref{tab:contact_comparison}.
}
\label{fig:3d-renderings}
\end{figure*}

\subsection{Global and Temporal Exchange}
\label{sec:supp-global-temporal}

Each of the four Geometry Encoder layers combines within-part attention over the $20$ template parts, cross-part attention, and a local feed-forward network with GELU activation.
The output projections of the cross-part and feed-forward branches are zero-initialized.

A fixed downsampling matrix $D_0\in\{0,1\}^{195\times778}$ selects $195$ template vertices, and a fixed sparse matrix $U_0\in\mathbb R^{778\times195}$ maps coarse updates back to the full surface.
For layer $\ell$, the persistent frame core $C_t^\ell\in\mathbb R^{216\times128}$ is updated by
\begin{equation}
\begin{aligned}
C_t^{\ell+1}
&=\operatorname{Core}_{\ell}\!\left(
C_t^\ell+\phi_{\mathrm{in}}^\ell
\bigl(\operatorname{Concat}(S_t,D_0H_t)\bigr)\right),\\
\Delta_t^\ell&=\phi_{\mathrm{out}}^\ell(C_t^{\ell+1}),\\
S_t&\leftarrow S_t+[\Delta_t^\ell]_{\mathrm{joint}},
\qquad H_t\leftarrow H_t+U_0[\Delta_t^\ell]_{\mathrm{surf}}.
\end{aligned}
\label{eq:supp-core}
\end{equation}
Concatenation joins token rows, with joints preceding surface tokens.
The core uses four-head self-attention and a feed-forward network, and the affine output map $\phi_{\mathrm{out}}^\ell$ is zero-initialized.

The core embeddings and the first three core attention modules are initialized from a FreiHAND-trained FastMETRO checkpoint~\citep{cho2022FastMETRO}.
The fourth attention module copies the third at initialization, and the remaining modules are initialized separately, identically in all training variants.

After each core update, a causal temporal block runs along each anatomical track, with weights shared across vertices and joints within a layer.
This restriction does not make the video backbone causal.
After the first three layers, in-image vertex states are bilinearly splatted to the neighboring patches and averaged over each two-frame tubelet.
A bias-free projection and zero-initialized channel gate add the result before the next interleaved ViT block.
Separate final $\operatorname{LayerNorm}$ operations yield vertex and joint states.

\begin{figure*}[!tb]
\centering
{\setlength{\tabcolsep}{0.5pt}\renewcommand{\arraystretch}{0.4}%
\begin{tabular}{ccccccccccc}
  & {\tiny PVDB} & {\tiny PVDB} & {\tiny PVDB} & {\tiny HandSense} & {\tiny HandSense} & {\tiny HandSense} & {\tiny OpenTouch} & {\tiny OpenTouch} & {\tiny OpenTouch} & {\tiny OpenTouch} \\[1pt]
  \glab{RGB} & \raisebox{-0.5\height}{\includegraphics[width=0.088\linewidth]{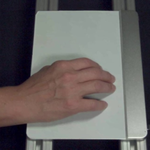}} & \raisebox{-0.5\height}{\includegraphics[width=0.088\linewidth]{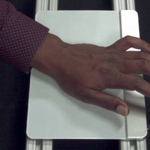}} & \raisebox{-0.5\height}{\includegraphics[width=0.088\linewidth]{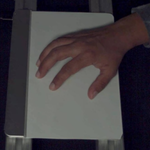}} & \raisebox{-0.5\height}{\includegraphics[width=0.088\linewidth]{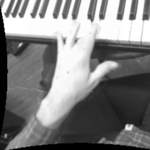}} & \raisebox{-0.5\height}{\includegraphics[width=0.088\linewidth]{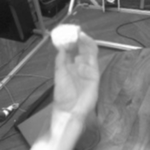}} & \raisebox{-0.5\height}{\includegraphics[width=0.088\linewidth]{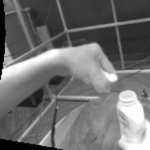}} & \raisebox{-0.5\height}{\includegraphics[width=0.088\linewidth]{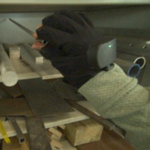}} & \raisebox{-0.5\height}{\includegraphics[width=0.088\linewidth]{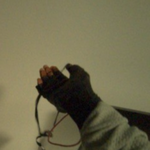}} & \raisebox{-0.5\height}{\includegraphics[width=0.088\linewidth]{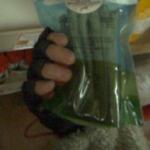}} & \raisebox{-0.5\height}{\includegraphics[width=0.088\linewidth]{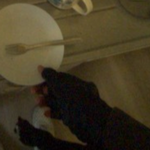}} \\
  \glab{GT} & \raisebox{-0.5\height}{\includegraphics[width=0.088\linewidth]{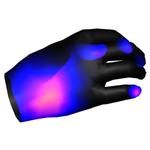}} & \raisebox{-0.5\height}{\includegraphics[width=0.088\linewidth]{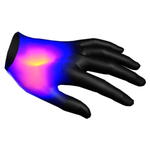}} & \raisebox{-0.5\height}{\includegraphics[width=0.088\linewidth]{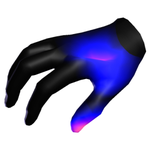}} & \raisebox{-0.5\height}{\includegraphics[width=0.088\linewidth]{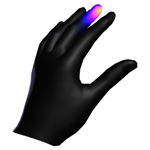}} & \raisebox{-0.5\height}{\includegraphics[width=0.088\linewidth]{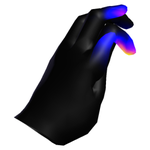}} & \raisebox{-0.5\height}{\includegraphics[width=0.088\linewidth]{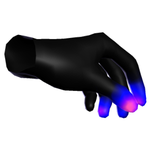}} & \raisebox{-0.5\height}{\includegraphics[width=0.088\linewidth]{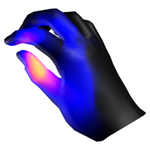}} & \raisebox{-0.5\height}{\includegraphics[width=0.088\linewidth]{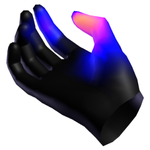}} & \raisebox{-0.5\height}{\includegraphics[width=0.088\linewidth]{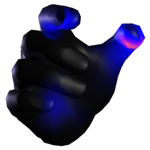}} & \raisebox{-0.5\height}{\includegraphics[width=0.088\linewidth]{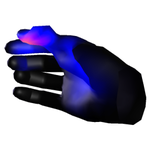}} \\
  \glab{\textbf{Ours (Joint)}} & \raisebox{-0.5\height}{\includegraphics[width=0.088\linewidth]{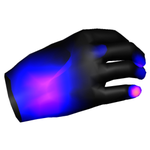}} & \raisebox{-0.5\height}{\includegraphics[width=0.088\linewidth]{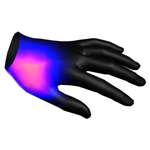}} & \raisebox{-0.5\height}{\includegraphics[width=0.088\linewidth]{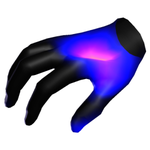}} & \raisebox{-0.5\height}{\includegraphics[width=0.088\linewidth]{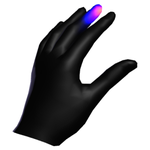}} & \raisebox{-0.5\height}{\includegraphics[width=0.088\linewidth]{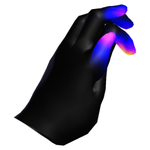}} & \raisebox{-0.5\height}{\includegraphics[width=0.088\linewidth]{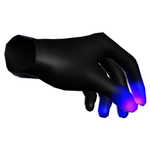}} & \raisebox{-0.5\height}{\includegraphics[width=0.088\linewidth]{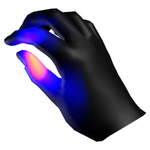}} & \raisebox{-0.5\height}{\includegraphics[width=0.088\linewidth]{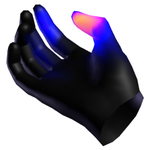}} & \raisebox{-0.5\height}{\includegraphics[width=0.088\linewidth]{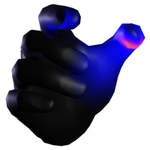}} & \raisebox{-0.5\height}{\includegraphics[width=0.088\linewidth]{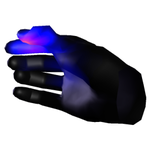}} \\[3pt]
  \glab{GT} & \raisebox{-0.5\height}{\includegraphics[width=0.088\linewidth]{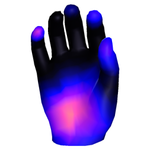}} & \raisebox{-0.5\height}{\includegraphics[width=0.088\linewidth]{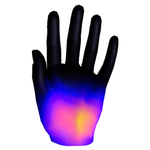}} & \raisebox{-0.5\height}{\includegraphics[width=0.088\linewidth]{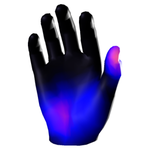}} & \raisebox{-0.5\height}{\includegraphics[width=0.088\linewidth]{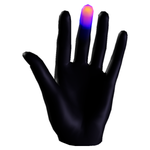}} & \raisebox{-0.5\height}{\includegraphics[width=0.088\linewidth]{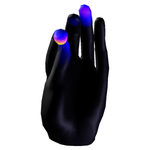}} & \raisebox{-0.5\height}{\includegraphics[width=0.088\linewidth]{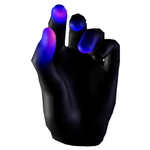}} & \raisebox{-0.5\height}{\includegraphics[width=0.088\linewidth]{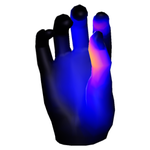}} & \raisebox{-0.5\height}{\includegraphics[width=0.088\linewidth]{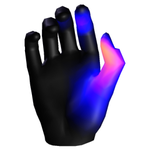}} & \raisebox{-0.5\height}{\includegraphics[width=0.088\linewidth]{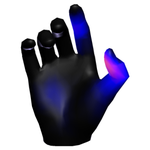}} & \raisebox{-0.5\height}{\includegraphics[width=0.088\linewidth]{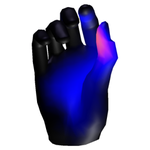}} \\
  \glab{\textbf{Ours (Joint)}} & \raisebox{-0.5\height}{\includegraphics[width=0.088\linewidth]{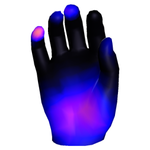}} & \raisebox{-0.5\height}{\includegraphics[width=0.088\linewidth]{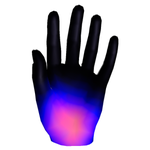}} & \raisebox{-0.5\height}{\includegraphics[width=0.088\linewidth]{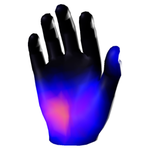}} & \raisebox{-0.5\height}{\includegraphics[width=0.088\linewidth]{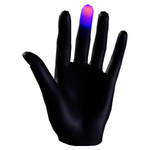}} & \raisebox{-0.5\height}{\includegraphics[width=0.088\linewidth]{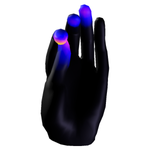}} & \raisebox{-0.5\height}{\includegraphics[width=0.088\linewidth]{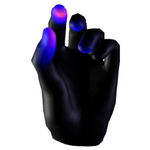}} & \raisebox{-0.5\height}{\includegraphics[width=0.088\linewidth]{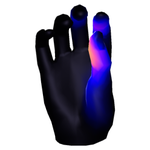}} & \raisebox{-0.5\height}{\includegraphics[width=0.088\linewidth]{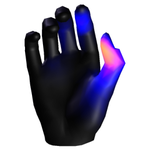}} & \raisebox{-0.5\height}{\includegraphics[width=0.088\linewidth]{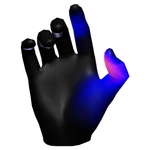}} & \raisebox{-0.5\height}{\includegraphics[width=0.088\linewidth]{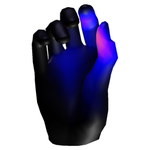}} \\
\end{tabular}}
\caption{
The examples of Figures~\ref{fig:additional-pvdb-handsense} to~\ref{fig:more-opentouch} from two viewpoints.
For each clip, the ground truth and Ours (Joint) at the frame with the largest measured total, seen from the video camera (upper rows) and the palm side (lower rows).
}
\label{fig:gallery}
\end{figure*}

\subsection{Mask Construction and Image Access}
\label{sec:supp-masking}

The interaction radius is $48\times384/224\simeq82.3$ pixels.
To select $t_c$, we find the first and last frames containing supported vertices.
Candidate bounds reserve three frames on either side and are clamped to $[6,10]$.
An integer is drawn uniformly from the resulting half-open interval when it is nonempty.
Otherwise, we use the midpoint split of the observed interval, clamped to $[6,10]$.

For anatomical masking, a clip selects a thumb and another finger with probability $0.30$, the palm with probability $0.175$, one finger with probability $0.35$, or no region with probability $0.175$.
The single-finger case adds a different second finger with probability $0.5$.
A selection is redrawn up to four attempts if a chosen region is visible in less than $5\%$ of its vertex-frame entries.
For hand bounding-box diagonal $b_t$ in pixels, the anatomical pixel mask uses discs of radius
$r_t^{\mathrm{anat}}=\operatorname{clip}(0.05b_t+3,5,17)$
around selected in-image vertices in every frame.
Both masks zero RGB values after input normalization.

Anatomically hidden vertices lose image access during initialization, per-layer lifting, and local image-window attention.
Associated joints also lose direct samples.
The local window radius is approximately $109.7$ pixels.
A vertex outside the image raster, which has no patch within this radius, attends to all patches.
Hidden vertices receive no window-attention update.
Geometry and communication through other parts, the core, and temporal tracks remain available.
This closes direct image reads while preserving the inference paths used to recover withheld evidence.

\section{Training Objectives and Joint Optimization}
\label{sec:supp-objectives}

\subsection{Dense Prediction and Patch Reconstruction}
\label{sec:supp-dense}

Let $h_v$ indicate membership in an anatomically hidden region,
$I_t=\mathbf1[t\geq t_c]$, and $\chi_{t,j}$ indicate a valid joint projection.
Before change-based reweighting, the dense weights are
\begin{equation}
\begin{aligned}
w_{t,v}^{0}
&=s_{t,v}\bigl(h_v+(1-h_v)[I_t+\lambda_c(k)(1-I_t)]\bigr),\\
w_{t,j}^{0}&=\chi_{t,j}[I_t+\lambda_c(k)(1-I_t)],
\qquad
\lambda_c(k)=0.2\min\!\left(1,\frac{k}{0.1K}\right),
\end{aligned}
\label{eq:supp-dense-weights}
\end{equation}
where $k$ is the training step and $K=12{,}000$ is the total number of updates.
For unhidden target vertices supported both currently and at $t_c-1$, we multiply $w_{t,v}^{0}$ by their mean absolute teacher-feature change divided by the mean change over eligible entries, clipped to $[0.25,4]$.
Other weights are unchanged.
Within each data lane, weighted vertex and joint errors share one normalization denominator.
The dense objective averages these normalized losses over lanes with nonzero support.

The residual predictor uses $\operatorname{LayerNorm}$ followed by a $256$-to-$512$-to-$256$ MLP with GELU and zero-initialized final affine layer.
The patch decoder uses a learned query plus an affine embedding of normalized patch position and frame index.
One four-head cross-attention block reads the same frame's $778$ vertex and $21$ joint states.
Output normalization and affine maps produce $d_p=128$ features.
The patch target is the normalized EMA counterpart of $F_t$.
Only interaction-masked patches in target frames contribute to Equation~\ref{eq:patch}, with the same within-lane normalization followed by averaging across lanes.

\subsection{Covariance Regularization}

WeakSIGReg samples at most $1024$ supported vertex states per device and uses an independently redrawn orthonormal sketch $Q\in\mathbb R^{256\times64}$ at each step.
For sampled states $Z\in\mathbb R^{n\times256}$, define centered sketches
$\widetilde Z=ZQ-\mathbf1\operatorname{mean}_{\mathrm{rows}}(ZQ)$.
The regularizer is
\begin{equation}
\mathcal L_{\mathrm{wSIG}}
=\left\|\frac{\widetilde Z^\top\widetilde Z}{n-1}-I_{64}\right\|_F.
\label{eq:supp-wsig}
\end{equation}
The Frobenius norm is not squared, and fewer than eight supported rows contribute zero.

\subsection{Tactile Supervision and Gradient Coupling}

Sensor annotations are lifted to vertices through fixed correspondences.
Each covered vertex receives the value and validity of its dominant sensor channel.
PressureVision DB already supplies vertex-level targets.
The tactile head uses a $256$-unit hidden layer, GELU, dropout $0.1$, and two zero-initialized outputs.
For training contact prevalence $r$, positive BCE entries receive weight
$\operatorname{clip}((1-r)/r,1,10)$.
Magnitude weights $\lambda_m$ are $1.0$, $1.0$, and $0.3$ for HandSense, PressureVision DB, and OpenTouch.
The magnitude normalization uses the encoder-training statistics in Table~\ref{tab:supp-normalization}.

The BCE and contact-conditioned log-magnitude terms update the encoder and head.
For the all-valid gated L1 term, evaluating the head on $\operatorname{sg}(z)$ preserves its parameter gradients while blocking encoder gradients.
Each component is normalized over its eligible entries within a clip and the sum is averaged over labeled clips.
The objective uses no tactile pseudo-labels or geometry-derived contact terms.

The tactile weight follows
$\lambda_{\mathrm{tac}}(k)=\min(1,0.1+0.9k/(0.1K))$.
Let $g_s$ and $g_t$ denote the corresponding unweighted SSL and tactile gradients at the shared vertex states, and write $\lambda=\lambda_{\mathrm{tac}}(k)$.
We apply
\begin{equation}
\begin{aligned}
g_t^+&=g_t-
\frac{\min(0,\langle g_t,g_s\rangle)}{\max(\|g_s\|_2^2,\epsilon)}g_s,\\
\alpha&=\min\!\left(1,
\frac{0.4\max(\|g_s\|_2,r_k)}{\max(\lambda\|g_t^+\|_2,\epsilon)}\right),
\qquad g_z=g_s+\lambda\alpha g_t^+.
\end{aligned}
\label{eq:supp-gradient-coupling}
\end{equation}
Here $\epsilon=10^{-20}$ and $r_k$ is the running mean SSL gradient norm collected at steps $100$ through $599$, held fixed afterwards and zero before collection.
The projection removes the opposing component at $z$~\citep{yu2020gradient}.
The cap uses the larger of the current and reference norms, while leaving $g_s$ unchanged.
The head receives ordinary gradients of the weighted tactile loss.

EMA updates the trainable visual and Geometry Encoder parameters after optimization.
The SSL predictor, patch decoder, and tactile head are outside this EMA copy.
The tactile head reads the student's masked-view vertex states from the same forward pass as the JEPA term, so the label gradient reaches the LoRA adapters, the visual-field projections, and the Geometry Encoder through both the contact and the log-magnitude terms.
Through the shared adapters, tactile gradients can also affect later visual targets, without generating tactile pseudo-labels.
For Supervised, the dense predictor is trained on detached states, so the encoder receives only tactile supervision.
Its tactile weight is one from the start and no gradient cap is applied.
All downstream readouts use unmasked frozen EMA states and discard the training predictor and tactile head.

\section{Readouts and Evaluation Protocol}
\label{sec:supp-readout}

\paragraph{Frozen representations and sensor correspondence.}
Evaluation fixes the EMA encoder and removes input masks.
The SSL predictor and the tactile head used in joint
training are discarded. A deterministic hand crop with
scale $1.5$ produces vertex features of shape
$16\times778\times256$, and Appendix~\ref{sec:supp-robustness}
repeats the evaluation without the crop.
Fixed row-normalized operators pool these features to
the sensor grid before the readout.
PressureVision DB uses the identity over $778$ vertices,
OpenTouch maps to $169$ taxels, and HandSense maps to
five fingertip channels. The same operators and training
clips are used for each Touch-JEPA training setting.

\paragraph{Readout fitting.}
For each dataset, we sample $1{,}500$ training clips with
seed $912$ and at most $400$ valid entries per clip.
A clip-level $20\%$ holdout is reserved for model selection.
Each feature dimension is standardized using the fitting
subset mean and standard deviation, with $10^{-6}$ added
to the denominator. The readout contains a $256$-unit
linear layer, GELU, dropout $0.1$, and a two-output linear
layer initialized to zero.
The contact probability $\hat c$ is obtained by applying
sigmoid to the contact logit.
The final normalized magnitude prediction is
$\hat y=\hat c\hat m$, where $\hat m$ is the nonnegative
magnitude output.
We minimize class-balanced binary cross-entropy plus
unit-weight L1 error of the gated magnitude.
The positive-class weight is the negative-to-positive
ratio in the fitting entries.
AdamW uses learning rate $10^{-3}$, weight decay $10^{-4}$,
and batch size $16{,}384$ for $40$ epochs (head seed $0$). We keep the epoch with the
lowest total holdout loss, without refitting on it.

\paragraph{Magnitude normalization.}
Encoder supervision and readout evaluation use separate
fixed training subsets of $120$ and $60$ clips per dataset.
Each scalar is the population standard deviation of all
valid measurements in its subset, pooled across clips,
frames, and channels.
Table~\ref{tab:supp-normalization} distinguishes these scales.
The normalized target $y$ in Equation~\ref{eq:tactile_loss}
is the raw measurement divided by $\nu_{\mathrm{enc}}$.
Readout targets are divided by $\nu_{\mathrm{eval}}$.
Before evaluation, the final gated prediction is converted
back to sensor units as
$\hat y^{\mathrm{raw}}=\nu_{\mathrm{eval}}\hat y$.
The reported errors of every method are normalized by
the same $\nu_{\mathrm{eval}}$.
OpenTouch contact uses a positive magnitude exceeding
the training-subset tenth percentile plus $0.25$ times
its standard deviation. This gives thresholds of
$108.54$ counts during encoder supervision and $107.58$
counts for readout fitting and evaluation.

\begin{table}[!htbp]
\centering
\caption{Fixed magnitude scales and evaluation support.
Scales use training data only.
Ablations and objective comparisons use all valid entries,
and baseline comparisons use common ones.}
\label{tab:supp-normalization}
\begingroup
\small
\setlength{\tabcolsep}{5pt}
\begin{tabular}{@{}lrrrr@{}}
\toprule
Dataset & $\nu_{\mathrm{enc}}$ & $\nu_{\mathrm{eval}}$
& All valid & Common \\
\midrule
OpenTouch & $434.146$ & $430.309$ & $6{,}811{,}376$ & $3{,}328{,}284$ \\
PressureVision DB & $5.354$ & $5.160$ & $3{,}230{,}307$ & $3{,}142{,}493$ \\
HandSense & $191.711$ & $181.521$ & $18{,}816$ & $18{,}811$ \\
\bottomrule
\end{tabular}
\endgroup
\end{table}

\paragraph{Comparison populations.}
Within each comparison, the evaluation mask intersects
annotation validity with the availability of every
method's predictions. Image-based pressure outputs
restrict the main comparison to their projection support.
The Touch-JEPA objective comparisons and ablations use
all annotation-valid entries, including entries outside
that support.

\paragraph{Contact metrics.}
AUROC and average precision use each method's contact score,
with Touch-JEPA using the contact logit $o$, and pool valid
test entries with $4{,}000$-bin histograms.
F1 pools true-positive, predicted-positive, and
ground-truth-positive counts.
On OpenTouch and HandSense, each method's score is thresholded
at the quantile of its own test scores that matches the
dataset's training contact rate, so no test label is used.
On PressureVision DB, contact is predicted when the final
pressure prediction, $\hat y^{\mathrm{raw}}$ for Touch-JEPA, is
at least $1$\,kPa, the labeling rule.
Temporal AUROC averages over clip-channel sequences with
at least four valid frames and both contact classes.

\paragraph{Magnitude metrics.}
For Touch-JEPA, all magnitude metrics use the final
native-unit prediction
$\hat y^{\mathrm{raw}}=\nu_{\mathrm{eval}}\hat c\hat m$,
where $\hat c=\sigma(o)$ is the contact probability
and $\hat m$ is the nonnegative normalized magnitude
from the same readout head.
Compared methods are evaluated using their own final
pressure predictions.
For annotation-valid comparison entries $\mathcal E$,
let $e_i=\hat y_i^{\mathrm{raw}}-y_i^{\mathrm{raw}}$
denote the native-unit prediction error. We compute
\begin{equation}
\mathrm{nMAE}=\frac{\sum_{i\in\mathcal E}|e_i|}
{\nu_{\mathrm{eval}}|\mathcal E|},
\qquad
\mathrm{nRMSE}=\frac{1}{\nu_{\mathrm{eval}}}
\sqrt{\frac{\sum_{i\in\mathcal E}e_i^2}{|\mathcal E|}}.
\label{eq:supp-magnitude-metrics}
\end{equation}
Temporal Spearman correlation is computed separately
for each sensor within a clip using only valid
ground-truth contact frames.
Eligible sequences contain at least four such frames
with nonconstant target magnitude.
Eligibility does not depend on predicted contact, and
constant predictions contribute zero.
The static prior is constant in every sequence, so it is
reported as N/A rather than zero.
For Touch-JEPA, conditioning on ground-truth contact
does not fix predicted contact confidence, so this metric evaluates the combined output, not the magnitude branch alone.

\paragraph{Baselines.}
The main comparison matches a budget of $6{,}144{,}000$ frame
presentations.
PressureFormer~\citep{zhao2025egopressure} is reimplemented with frozen HaMeR~\citep{pavlakos2024reconstructing} image
features and a UV pressure decoder, using supervised
PVDB vertices for its projection loss.
All outputs are mapped to the same sensor grids.
HOPE~\citep{jeon2026hope} and PressureVision++~\citep{grady2024pressurevision++}
additionally train on DexYCB contact labels.
On PressureVision DB, the hand meshes are fused from HaWoR fits
of the synchronized views, and the mesh-based baselines (HOPE,
PressureFormer) use the same geometry.
The static prior uses fixed training-set contact rates
and mean magnitudes per sensor entry, without video.

\paragraph{Uncertainty.}
Unless stated otherwise, paired bootstrap intervals resample test clips $200$
times with seed $912$, applying the same resample to
all compared methods. Difference intervals use the
$2.5$th and $97.5$th percentiles.
They capture test-sample variation for one training seed and fixed readouts.

\section{Additional Results}
\label{sec:supp-results}

\subsection{Training Budgets}
\label{sec:supp-budgets}
Figure~\ref{fig:budgets} compares Supervised and
Ours (Joint) after $192$k, $288$k, and $384$k training clips, scored
with the main-table readout on all valid test entries of each dataset
(the main comparison uses the entries shared with the baselines).
At every budget, Ours (Joint) has the higher contact AUROC and F1 on all
three test sets. Across the $54$ cells of three budgets, three test
sets, and six metrics, it is significantly better than Supervised in
$46$ cells. Supervised has the higher HandSense tAUC at every budget,
significantly so at $384$k.

\begin{figure}[!htbp]
\centering
\definecolor{armLAB}{HTML}{3775BA}
\definecolor{armJOINT}{HTML}{E2601B}
{\footnotesize
\textcolor{armLAB}{\rule[0.4ex]{0.8em}{1pt}}\,Supervised\quad
\textcolor{armJOINT}{\rule[0.4ex]{0.8em}{1pt}}\,Ours (Joint)\par}
\vspace{2pt}
\includegraphics[width=\linewidth]{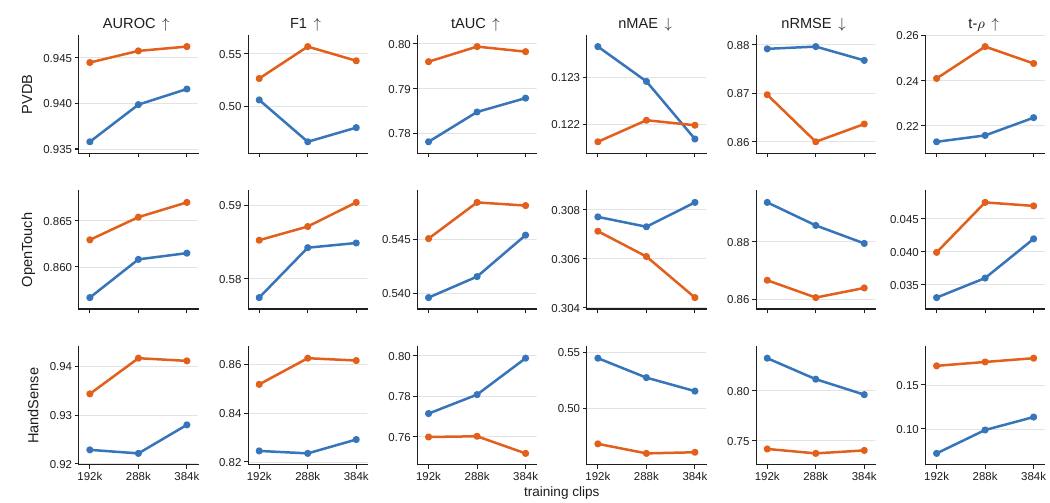}
\caption{Training budgets. Results after $192$k, $288$k, and $384$k training clips.}
\label{fig:budgets}
\end{figure}

\subsection{Evaluation without the Hand Crop}
\label{sec:supp-robustness}
Our encoder reads a hand crop computed from the estimated hand mesh.
Training samples both the observer view and the hand crop, so the
observer view is not out of distribution. With readouts refitted and
evaluated on each dataset's observer view without the crop, Ours (SSL)
and Ours (Joint) reach a
contact AUROC of $0.929$ and $0.951$ on PVDB, $0.839$ and $0.864$ on
OpenTouch, and $0.901$ and $0.928$ on HandSense, above the best
supervised baseline ($0.920$, $0.785$, and $0.844$).

\subsection{Sensor-Shared Intensity Axis}
\label{sec:supp-axis}
The three sensors report intensity in incompatible units
(nominal kPa on PVDB, uncalibrated counts on the two gloves).
We therefore ask whether the intensity read out on one sensor
orders the measurements of another. With the $192$k-clip checkpoints, a readout
fitted on one sensor is applied unchanged to the others and scored by rank
correlation (Figure~\ref{fig:sensor_axis}).
The readout is a two-layer head fitted on the training clips of the
source sensor, with the intensity loss applied to ground-truth contact
entries only, and is transferred together with the source
standardization.
For Ours (Joint), the mean cross-sensor rank correlation reaches
$0.73$ of the mean within-sensor one ($0.374$ against $0.509$),
against $0.36$ for Ours (SSL) ($0.171$ against $0.477$)
and $0.25$ for Supervised ($0.117$ against $0.470$).
Supervised transfers contact across sensors (mean cross-sensor
contact AUROC $0.854$, against $0.879$ for Ours (Joint) and $0.554$
for Ours (SSL)) but intensity much less.
The shared intensity axis is strongest under joint training,
consistent with both objectives shaping the same hand-surface latents.

\begin{figure}[!htbp]
\centering
\definecolor{armSSL}{HTML}{F4A259}
\definecolor{armLAB}{HTML}{3775BA}

\definecolor{armJOINT}{HTML}{E2601B}
{\footnotesize
\textcolor{armJOINT}{\rule[0.4ex]{0.8em}{3pt}}\,Ours (Joint)\quad
\textcolor{armSSL}{\rule[0.4ex]{0.8em}{3pt}}\,Ours (SSL)\quad
\textcolor{armLAB}{\rule[0.4ex]{0.8em}{3pt}}\,Supervised\par}
\vspace{2pt}
\includegraphics[width=\linewidth]{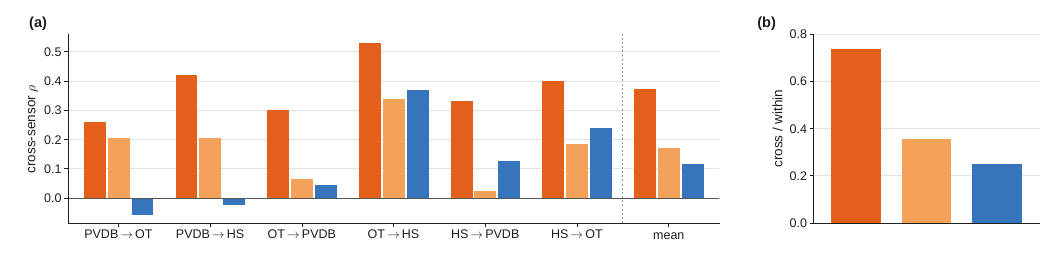}
\caption{A sensor-shared intensity axis. A readout fitted on
one sensor is applied unchanged to another. (a) Spearman correlation
over the contact entries of the target sensor, per transfer direction.
(b) Mean cross-sensor correlation relative to the within-sensor one.}
\label{fig:sensor_axis}
\end{figure}

\section{Failure Analysis}
\label{sec:supp-failure}

\paragraph{Analysis protocol.}
We examine challenging conditions using the test predictions
from the main comparisons.
Qualitative examples are selected to illustrate failures
and do not estimate their prevalence in the test sets.

\paragraph{Motion and limited surface visibility.}
The OpenTouch examples in Figure~\ref{fig:failure-opentouch}
show incomplete recovery of tactile responses under
image motion and limited surface visibility.
In the kitchen sequence, Ours (Joint) underestimates
the broad palm response present in the ground truth,
with predictions concentrated near the fingertips.
In the restaurant sequence, the predicted palm response
is similarly weak, while the thumb-tip response is overestimated
relative to the ground truth.
These examples illustrate errors in both the spatial
distribution and magnitude of predicted tactile responses.

\paragraph{Sparse contact events.}
Figure~\ref{fig:failure-handsense} illustrates weak or missing predictions for localized fingertip contacts.
In the piano sequence, the brief index-fingertip
response in the ground truth is substantially weaker
in Ours (Joint).
In the bottle-opening sequence, both Ours variants fail
to reproduce the index-fingertip response, while
PressureFormer captures part of it.
These examples show that brief or localized contact
can remain underrepresented in the predicted tactile
maps despite joint training.

\begin{figure}[!tb]
\centering
\begin{minipage}[t]{0.498\linewidth}
\vspace{0pt}
\centering
{\footnotesize (a) Image motion, kitchen\par}
\vspace{3pt}
{\setlength{\tabcolsep}{0.5pt}\renewcommand{\arraystretch}{0.4}%
\begin{tabular}{@{}cc@{}}
\glab{RGB} & \raisebox{-0.5\height}{\includegraphics[width=0.955\linewidth]{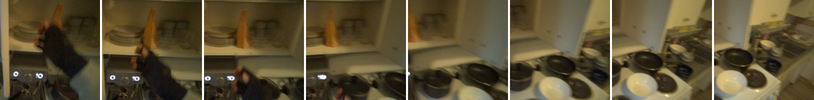}} \\
\glab{GT} & \raisebox{-0.5\height}{\includegraphics[width=0.955\linewidth]{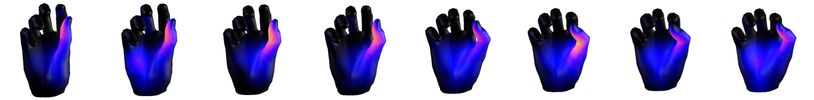}} \\
\glab{\textbf{Ours}\\\textbf{(Joint)}} & \raisebox{-0.5\height}{\includegraphics[width=0.955\linewidth]{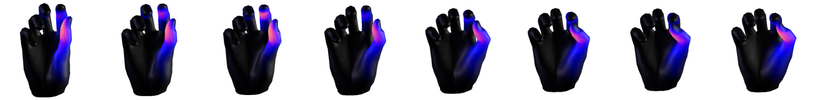}} \\
\glab{\textbf{Ours}\\\textbf{(SSL)}} & \raisebox{-0.5\height}{\includegraphics[width=0.955\linewidth]{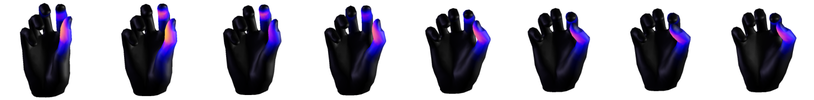}} \\
\glab{HOPE} & \raisebox{-0.5\height}{\includegraphics[width=0.955\linewidth]{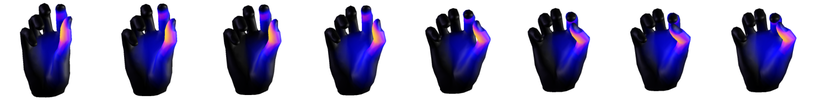}} \\
\glab{Pressure\\Former} & \raisebox{-0.5\height}{\includegraphics[width=0.955\linewidth]{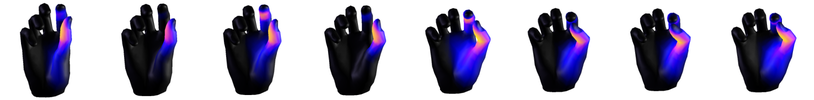}} \\
\glab{Pressure\\Vision++} & \raisebox{-0.5\height}{\includegraphics[width=0.955\linewidth]{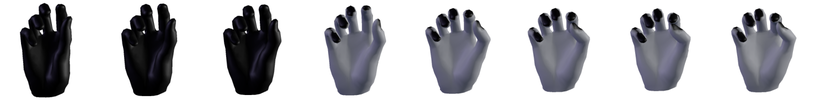}} \\
\glab{Pressure\\Vision} & \raisebox{-0.5\height}{\includegraphics[width=0.955\linewidth]{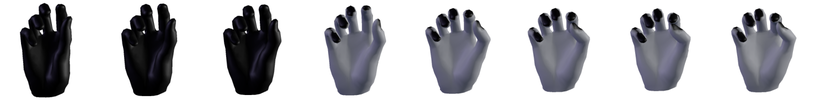}} \\
\end{tabular}}
\end{minipage}\hfill%
\begin{minipage}[t]{0.498\linewidth}
\vspace{0pt}
\centering
{\footnotesize (b) Limited visibility, restaurant\par}
\vspace{3pt}
{\setlength{\tabcolsep}{0.5pt}\renewcommand{\arraystretch}{0.4}%
\begin{tabular}{@{}cc@{}}
\phantom{\glab{RGB}} & \raisebox{-0.5\height}{\includegraphics[width=0.955\linewidth]{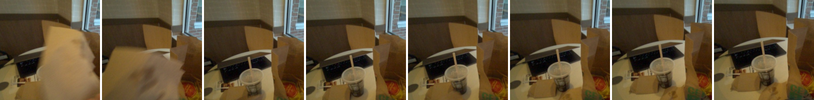}} \\
\phantom{\glab{GT}} & \raisebox{-0.5\height}{\includegraphics[width=0.955\linewidth]{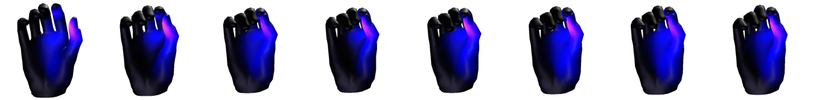}} \\
\phantom{\glab{\textbf{Ours}\\\textbf{(Joint)}}} & \raisebox{-0.5\height}{\includegraphics[width=0.955\linewidth]{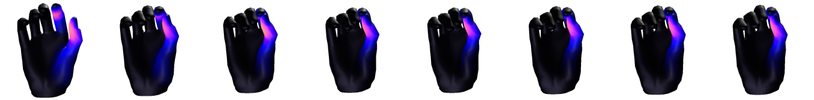}} \\
\phantom{\glab{\textbf{Ours}\\\textbf{(SSL)}}} & \raisebox{-0.5\height}{\includegraphics[width=0.955\linewidth]{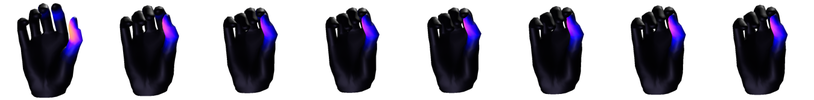}} \\
\phantom{\glab{HOPE}} & \raisebox{-0.5\height}{\includegraphics[width=0.955\linewidth]{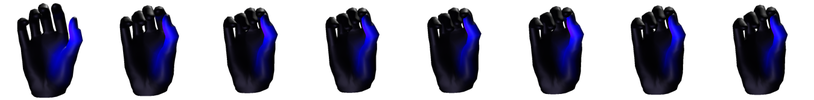}} \\
\phantom{\glab{Pressure\\Former}} & \raisebox{-0.5\height}{\includegraphics[width=0.955\linewidth]{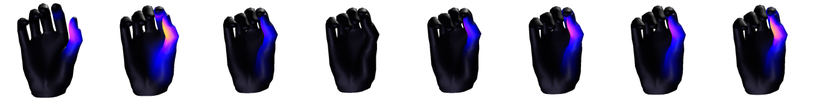}} \\
\phantom{\glab{Pressure\\Vision++}} & \raisebox{-0.5\height}{\includegraphics[width=0.955\linewidth]{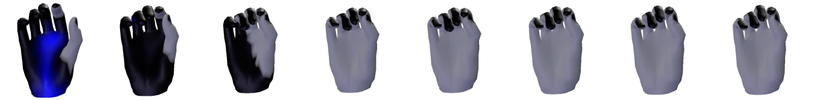}} \\
\phantom{\glab{Pressure\\Vision}} & \raisebox{-0.5\height}{\includegraphics[width=0.955\linewidth]{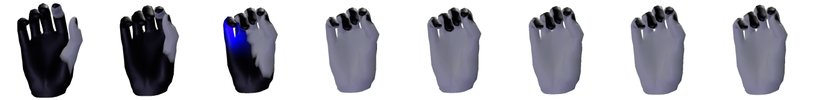}} \\
\end{tabular}}
\end{minipage}
\caption{
Failure cases on OpenTouch.
Eight frames per sequence progress from left to right. Brighter colors indicate greater sensor-native magnitude, and gray marks
missing predictions.
}
\label{fig:failure-opentouch}
\end{figure}

\section{Robotic Transfer and Manipulation}
\label{sec:supp-robot}

The robot studies examine two uses of the learned representation.
Zero-shot transfer applies a human-trained tactile predictor to real
Allegro-hand video. Action-conditioned planning instead trains a
dynamics predictor on simulated robot trajectories while keeping the
visual encoder frozen. The latter uses observed actions and robot state
but no tactile targets for learning or planning.

\subsection{Zero-Shot Tactile Estimation on Robotic Hands}
\label{sec:supp-allegro}

\paragraph{Data and correspondence.}
We evaluate the Allegro portion of the paired human and robot recordings
in HRDexDB~\citep{lim2026hrdexdb}. The collection contains $89$ robot
episodes.
Available synchronized camera views provide $2{,}754$ valid contexts
covering $10{,}562$ distinct robot frames.

Measured robot joint angles drive forward kinematics. A fixed retargeting
maps the index, middle, ring, and thumb fingertips to the corresponding
MANO fingertips. We fit the MANO pose with $500$ Adam steps at learning
rate $0.05$ and a pose-prior weight of $0.01$. This geometric proxy
provides image sampling locations without fitting to tactile data.
The virtual little finger has no robot target and is excluded from
scoring. In this real-video pipeline it remains in the encoder input
under the pose prior. Hand crops use an expansion factor of $1.5$ and
are resized to $384\times384$ for Touch-JEPA.

\paragraph{Human-trained predictions.}
For Touch-JEPA, the fixed HandSense correspondence operator pools vertex
features into fingertip features before standardization and readout.
We retain its four rows with robot counterparts. The saved heads are
fitted on the HandSense training split with the same recipe used for human
evaluation in Appendix~\ref{sec:supp-readout}. No robot examples fit or
adapt these heads. HOPE pools its predicted vertex field, while
PressureVision and PressureVision++ pool projected class logits before
decoding magnitude. PressureFormer reads the same view through its own
hand crop and pools its vertex logits over the fingertip regions before
decoding. For each robot frame, finite predictions from all
covering contexts and camera views are averaged. All methods are scored
on the same $42{,}207$ valid frame and fingertip entries.

\begin{figure}[!tb]
\centering
\begin{minipage}[t]{0.498\linewidth}
\vspace{0pt}
\centering
{\footnotesize (a) Piano, brief index-fingertip contact\par}
\vspace{3pt}
{\setlength{\tabcolsep}{0.5pt}\renewcommand{\arraystretch}{0.4}%
\begin{tabular}{@{}cc@{}}
\glab{RGB} & \raisebox{-0.5\height}{\includegraphics[width=0.955\linewidth]{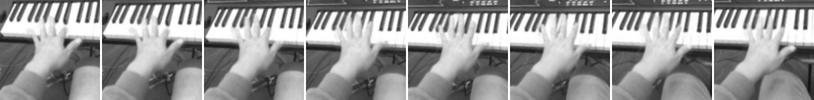}} \\
\glab{GT} & \raisebox{-0.5\height}{\includegraphics[width=0.955\linewidth]{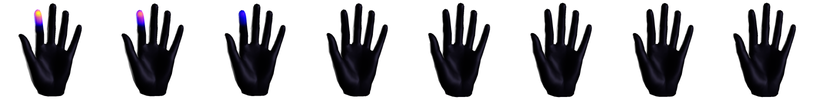}} \\
\glab{\textbf{Ours}\\\textbf{(Joint)}} & \raisebox{-0.5\height}{\includegraphics[width=0.955\linewidth]{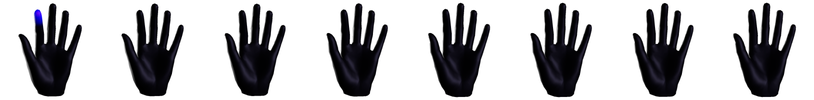}} \\
\glab{\textbf{Ours}\\\textbf{(SSL)}} & \raisebox{-0.5\height}{\includegraphics[width=0.955\linewidth]{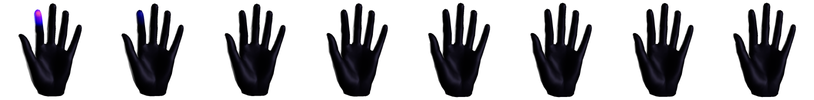}} \\
\glab{HOPE} & \raisebox{-0.5\height}{\includegraphics[width=0.955\linewidth]{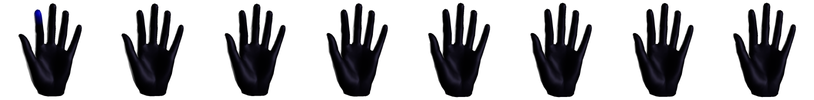}} \\
\glab{Pressure\\Former} & \raisebox{-0.5\height}{\includegraphics[width=0.955\linewidth]{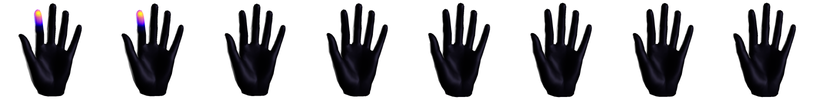}} \\
\glab{Pressure\\Vision++} & \raisebox{-0.5\height}{\includegraphics[width=0.955\linewidth]{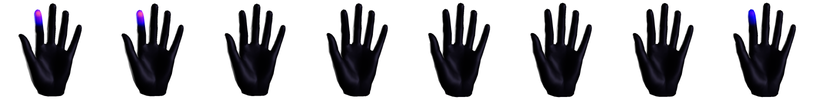}} \\
\glab{Pressure\\Vision} & \raisebox{-0.5\height}{\includegraphics[width=0.955\linewidth]{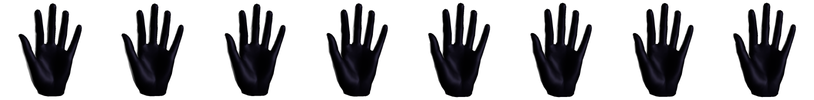}} \\
\end{tabular}}
\end{minipage}\hfill%
\begin{minipage}[t]{0.498\linewidth}
\vspace{0pt}
\centering
{\footnotesize (b) Bottle opening, index-fingertip contact\par}
\vspace{3pt}
{\setlength{\tabcolsep}{0.5pt}\renewcommand{\arraystretch}{0.4}%
\begin{tabular}{@{}cc@{}}
\phantom{\glab{RGB}} & \raisebox{-0.5\height}{\includegraphics[width=0.955\linewidth]{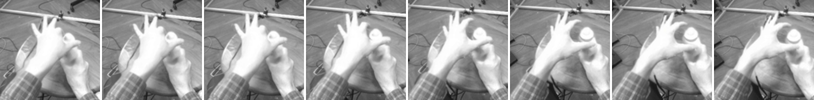}} \\
\phantom{\glab{GT}} & \raisebox{-0.5\height}{\includegraphics[width=0.955\linewidth]{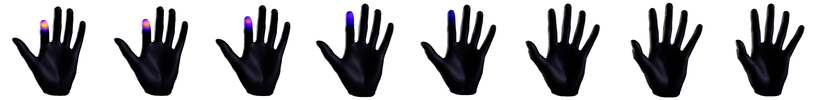}} \\
\phantom{\glab{\textbf{Ours}\\\textbf{(Joint)}}} & \raisebox{-0.5\height}{\includegraphics[width=0.955\linewidth]{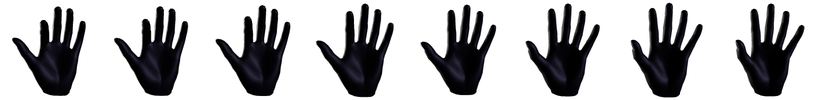}} \\
\phantom{\glab{\textbf{Ours}\\\textbf{(SSL)}}} & \raisebox{-0.5\height}{\includegraphics[width=0.955\linewidth]{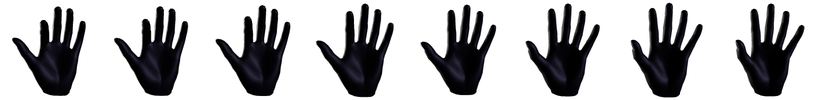}} \\
\phantom{\glab{HOPE}} & \raisebox{-0.5\height}{\includegraphics[width=0.955\linewidth]{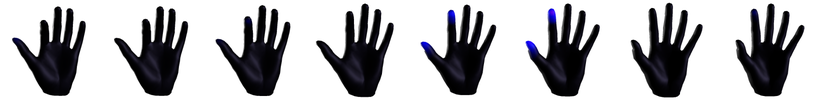}} \\
\phantom{\glab{Pressure\\Former}} & \raisebox{-0.5\height}{\includegraphics[width=0.955\linewidth]{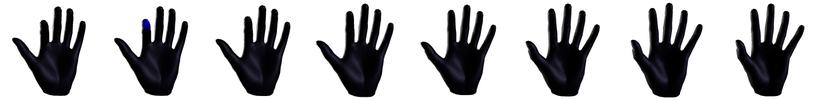}} \\
\phantom{\glab{Pressure\\Vision++}} & \raisebox{-0.5\height}{\includegraphics[width=0.955\linewidth]{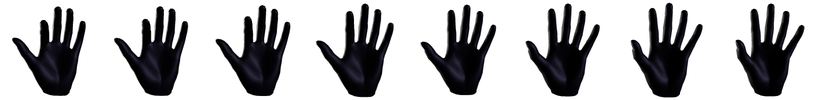}} \\
\phantom{\glab{Pressure\\Vision}} & \raisebox{-0.5\height}{\includegraphics[width=0.955\linewidth]{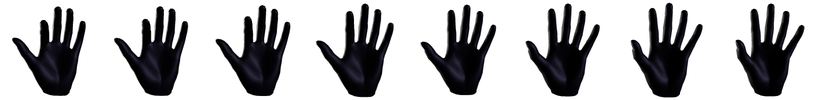}} \\
\end{tabular}}
\end{minipage}
\caption{
Failure cases on HandSense.
Display conventions follow Figure~\ref{fig:failure-opentouch}.
}
\label{fig:failure-handsense}
\end{figure}

\paragraph{Targets and statistics.}
A robot tactile sample is valid when its nearest timestamp is within
$50$\,ms of the video frame. Contact labels use a channel-specific
threshold equal to the tenth percentile plus $0.25$ standard deviations
of tactile readings from the designated robot training recordings.
These measurements define evaluation labels and do not update the
predictors. The readings are noisy and sparse~\citep{lim2026hrdexdb},
which bounds the attainable scores, and all methods are scored against the
same readings.
Pooled AUROC and average precision (AP) use contact scores over the common entries.
F1 thresholds each method's scores at the quantile that matches the
human training contact rate of $0.512$, so no robot label is used.

Magnitude correlation is computed within each episode and fingertip
using at least eight valid samples with nonconstant targets and
predictions. We average finite Spearman correlations over fingertips
and then over episodes. Unlike the human benchmark, a sequence with constant
predictions is skipped for that method. Paired
bootstrap comparisons resample episodes $1{,}000$ times with seed
$140914$ and report the $2.5$ and $97.5$ percentiles of each difference.
For joint minus SSL, the intervals are $[-0.087,-0.014]$ for AUROC,
$[-0.119,-0.040]$ for AP, $[-0.075,-0.008]$ for F1, and
$[-0.136,-0.017]$ for magnitude correlation. These intervals support
the SSL encoder choice in the planning experiment.

\paragraph{Additional evaluation on the Inspire hand.}
We additionally evaluate zero-shot tactile estimation
on sequences of the five-finger Inspire hand from
HRDexDB~\citep{lim2026hrdexdb}.
Table~\ref{tab:r2r_inspire} reports results on its five
fingertip channels.
Ours (SSL) achieves the highest scores across all four
metrics among the compared methods.

\begin{table}[!htb]
\centering
\caption{
Zero-shot transfer to the Inspire hand.
Evaluation covers five fingertip channels.
}
\label{tab:r2r_inspire}
\begingroup
\setlength{\tabcolsep}{8pt}
\renewcommand{\arraystretch}{1.15}
\small
\begin{tabular}{@{}lcccc@{}}
\toprule
Method
& AUROC $\uparrow$
& AP $\uparrow$
& F1 $\uparrow$
& $\rho$ $\uparrow$ \\
\midrule
PressureVision
& 0.494
& 0.145
& 0.218
& $-0.023$ \\
PressureVision++
& 0.479
& 0.137
& 0.213
& 0.002 \\
PressureFormer
& \rktwo{0.575}
& \rkthree{0.173}
& \rktwo{0.270}
& \rkthree{0.003} \\
HOPE
& \rkthree{0.568}
& \rktwo{0.178}
& \rkthree{0.257}
& \rktwo{0.014} \\
\midrule
\textbf{Ours (SSL)}
& \rkone{0.608}
& \rkone{0.195}
& \rkone{0.284}
& \rkone{0.059} \\
\textbf{Ours (Joint)}
& 0.509
& 0.150
& 0.222
& $-0.015$ \\
\bottomrule
\end{tabular}

\endgroup
\end{table}

\begin{figure}[!tb]
\centering
\includegraphics[width=\linewidth]{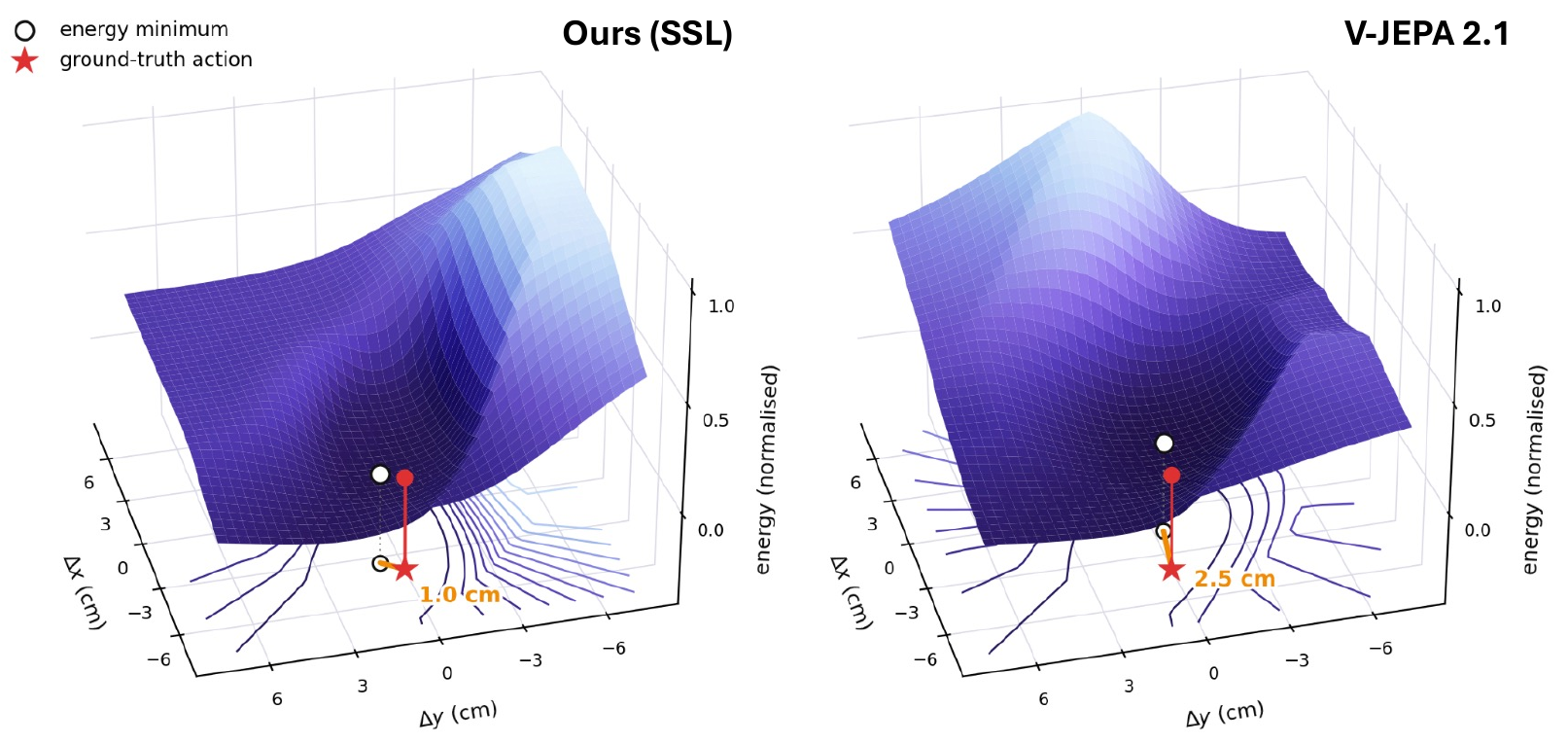}
\caption{
One-step scene prediction energy during object transport.
Horizontal translation commands are varied on a
$41\times41$ grid, with remaining command dimensions
fixed before controller projection.
Energy is the scene-feature L1 distance to the next
reference observation, normalized independently
to $[0,1]$ within each panel.
White circles mark sampled minima and red stars
denote the ground-truth action.
Annotated distances are measured in the displayed
command plane.
}
\label{fig:supp-energy}
\end{figure}

\subsection{Observations and Action-Conditioned Learning}
\label{sec:ac_details}

\paragraph{Simulated observations.}
We use a fixed-base KUKA iiwa arm and an Allegro hand in Isaac Gym.
Physics runs at $120$\,Hz with two substeps. A calibrated RGB camera
records $384\times384$ images at $4$\,Hz with a $55$-degree field of view.
Each encoder input contains the trailing $16$ actual frames, spanning
$3.75$ seconds and ending at the current observation. Although the
backbone is noncausal within this window, it receives no frames after
the decision time. The input is the full camera image with ImageNet
normalization.

The MANO proxy is fitted to the four fingertip pads using forward
kinematics and the pose fitting procedure above. In simulation, the
unsupported little finger's $114$ vertices and four joints are marked
invalid, leaving $664$ supported vertices. At decision $\tau$, the
observation is $\mathbf o_\tau=(B_\tau,U_\tau,\mathbf p_\tau)$.
Scene tokens $B_\tau\in\mathbb R^{36\times768}$ are obtained by
$4\times4$ average pooling of the final backbone layer's $24\times24$
patch grid at its last tubelet. Surface tokens
$U_\tau\in\mathbb R^{32\times256}$ average current-frame vertex states
within $32$ fixed groups of supported vertex indices. The state
$\mathbf p_\tau\in\mathbb R^{35}$ concatenates $23$ joint angles and
the flattened $3\times4$ palm pose. Each feature group uses channel-wise
training means and standard deviations, with a standard-deviation
floor of $0.05$.

\paragraph{Predictor and training data.}
The action-conditioned predictor has $24$ Transformer blocks of width
$1024$, $16$ attention heads, and feed-forward width $4096$.
Separate projections embed $36$ scene tokens, $32$ surface tokens,
one state token, and one action token. Learned slot and step embeddings
identify these $70$ tokens per decision. Block-causal attention permits
all tokens from the current and preceding decisions within an
eight-step context.

Training uses $696$ trajectories from $408$ initial scenes, with $232$
trajectories each for grasp-lift, place, and push-align. Push-align is
used only for training. Trajectories and reference demonstrations use
scene seeds distinct from evaluation. An eight-dimensional issued
action contains three palm-translation commands, four finger-flexion
commands, and thumb opposition. Action inputs record the issued
commands after controller projection.

For group $r\in\{B,U,\mathbf p\}$, let
$\mathcal L_r^{\mathrm{tf}}$ be mean absolute prediction error over
eight transitions conditioned on observed history. Let
$\mathcal L_r^{\mathrm{ar}}$ be the corresponding mean over a two-step
autoregressive rollout. The predictor objective is
\begin{equation}
\mathcal L_{\mathrm{AC}}
=\sum_{r\in\{B,U,\mathbf p\}}\lambda_r
\bigl(\mathcal L_r^{\mathrm{tf}}+\mathcal L_r^{\mathrm{ar}}\bigr),
\qquad (\lambda_B,\lambda_U,\lambda_{\mathbf p})=(1,1,0.1).
\label{eq:ac_objective}
\end{equation}
The autoregressive rollout feeds back predicted scene features,
surface features, and robot state, with gradients through both steps.
Future measured state is not supplied. Encoder features are frozen
targets, and object state, contact, and magnitude are excluded from this loss.
AdamW uses $1{,}024$ updates of $256$ windows, weight decay $0.04$,
gradient clipping at norm $1$, and seed $917$. The learning rate warms
up for $49$ updates to $4.25\times10^{-4}$, remains constant through
update $975$, and decays linearly to zero. Evaluation uses the final
predictor checkpoint.

\subsection{Planning and Control}

Planning minimizes the scene and surface distance in
Equation~\ref{eq:ac_energy} with a one-step horizon.
Reference observations at demonstration endpoints
$2$, $4$, $14$, and $18$ provide successive subgoals
for $2$, $2$, $10$, and $4$ decisions.
Each encoder computes its own reference features
using its training normalization.
Predicted robot state is rolled forward by the predictor
but does not enter the goal distance.

At each planning decision, the cross-entropy method
samples $400$ candidates for $10$ refinement iterations
and retains $10$ elites.
The search variables are three palm translations and
one grip value, which is copied to the five hand commands.
Initial search standard deviation is $0.7$,
with a floor of $0.03$.
Translation is limited to an L1 norm of $7.5$\,cm
per decision and a fixed workspace.
The controller executes the projected final mean through
inverse kinematics and a smooth joint ramp.

\paragraph{One-step energy landscape.}
Following the qualitative energy analysis in
V-JEPA~2~\citep{assran2025v},
Figure~\ref{fig:supp-energy} visualizes the scene-feature
prediction energy for a reference carry transition
in the place task.
The goal is the next observation in the demonstration.
In this example, the sampled minimum lies $1.0$\,cm
from the ground-truth action for Touch-JEPA-AC,
compared with $2.5$\,cm for V-JEPA~2.1
in the horizontal command plane.

\begin{figure}[!tb]
\centering

\begin{minipage}{0.95\linewidth}
\centering
\begingroup
\setlength{\tabcolsep}{1pt}
\renewcommand{\arraystretch}{0}

\begin{tabular}{@{}
    >{\centering\arraybackslash}m{12pt}
    *{6}{>{\centering\arraybackslash}
    m{\dimexpr(\linewidth-12pt-12\tabcolsep)/6\relax}}
@{}}
& {\scriptsize Start}
& {\scriptsize Closed}
& {\scriptsize Lift}
& {\scriptsize Carry}
& {\scriptsize Lower / release}
& {\scriptsize End} \\[3pt]

\rotatebox{90}{\tiny\textbf{Ours (SSL)}}
& \includegraphics[width=\linewidth]
  {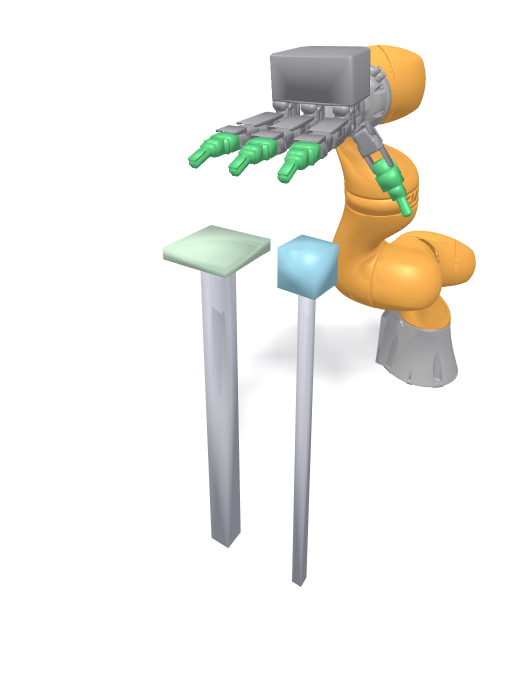}
& \includegraphics[width=\linewidth]
  {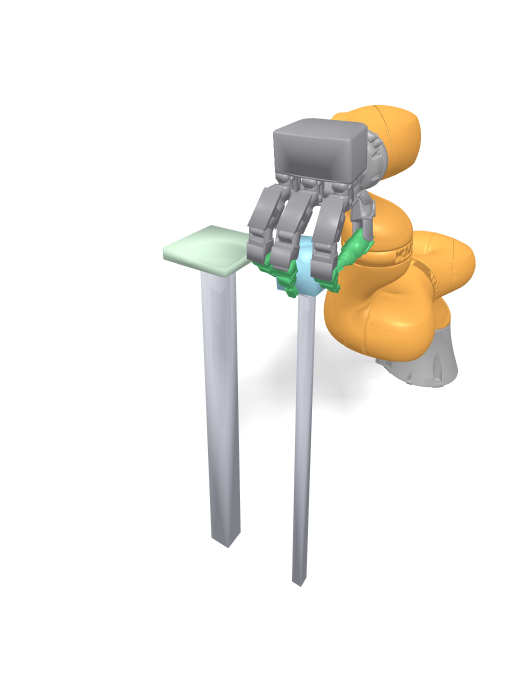}
& \includegraphics[width=\linewidth]
  {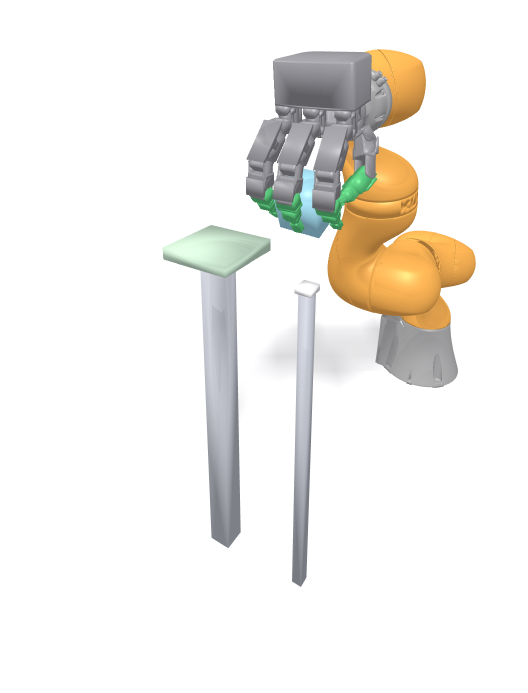}
& \includegraphics[width=\linewidth]
  {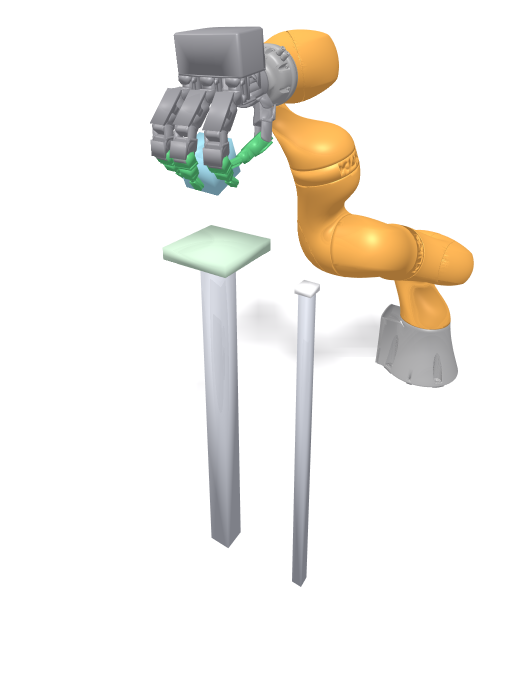}
& \includegraphics[width=\linewidth]
  {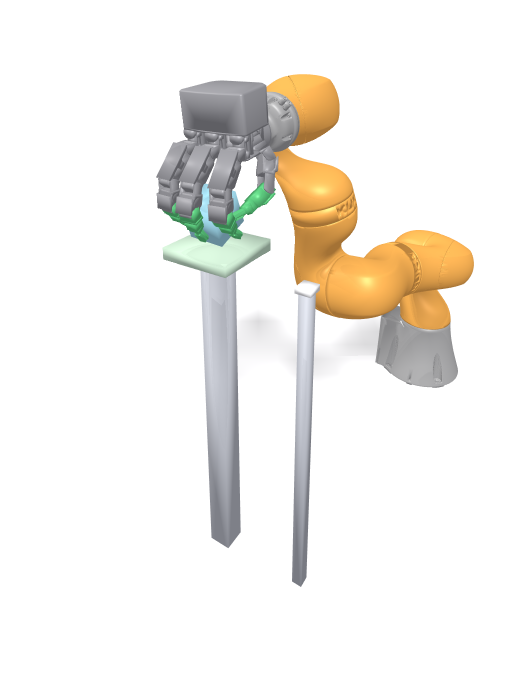}
& \includegraphics[width=\linewidth]
  {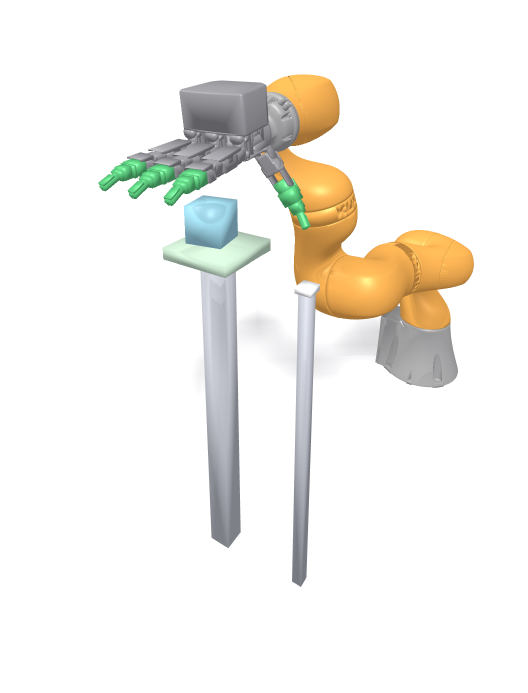} \\[2pt]

\rotatebox{90}{\tiny V-JEPA 2.1}
& \includegraphics[width=\linewidth]
  {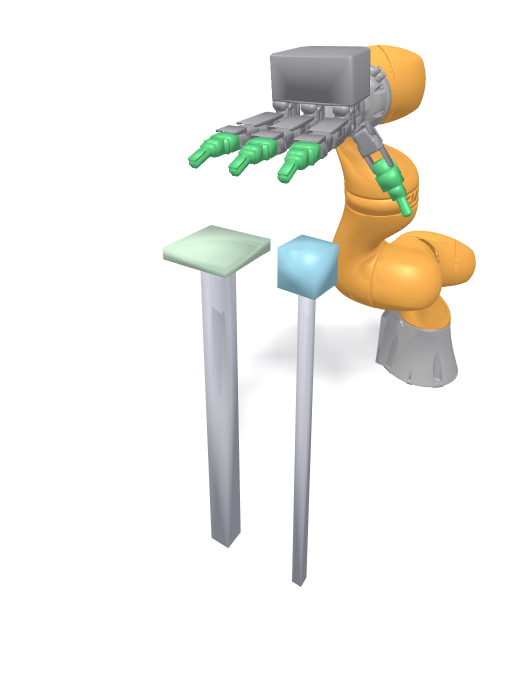}
& \includegraphics[width=\linewidth]
  {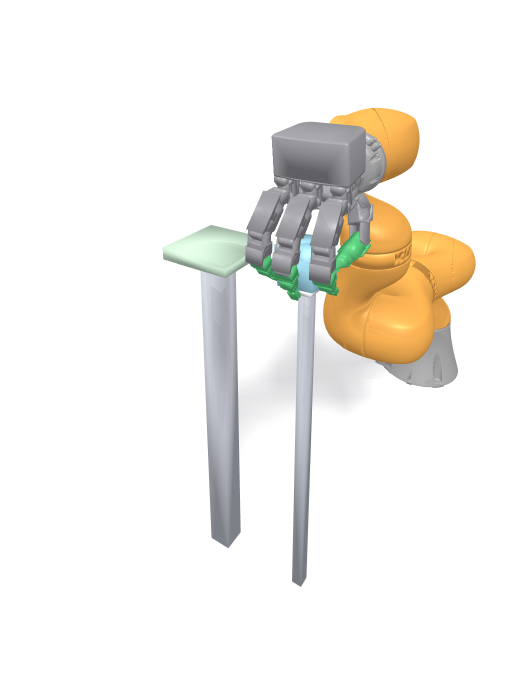}
& \includegraphics[width=\linewidth]
  {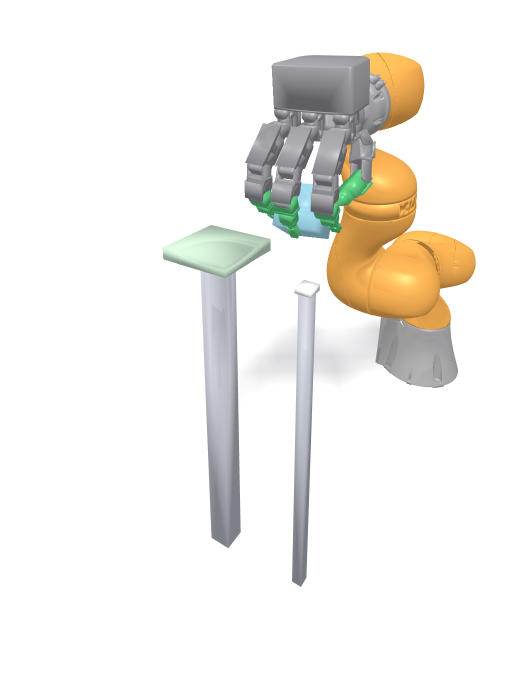}
& \includegraphics[width=\linewidth]
  {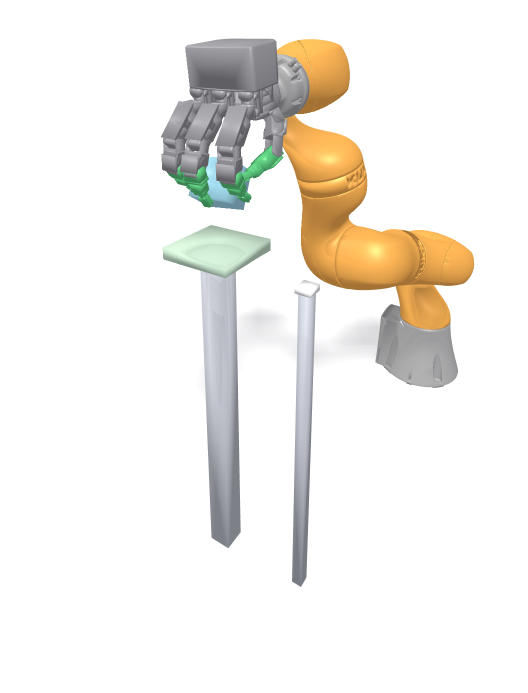}
& \includegraphics[width=\linewidth]
  {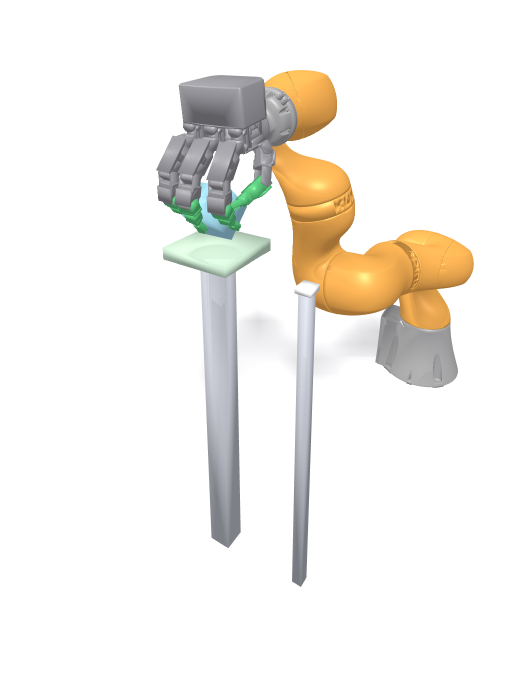}
& \includegraphics[width=\linewidth]
  {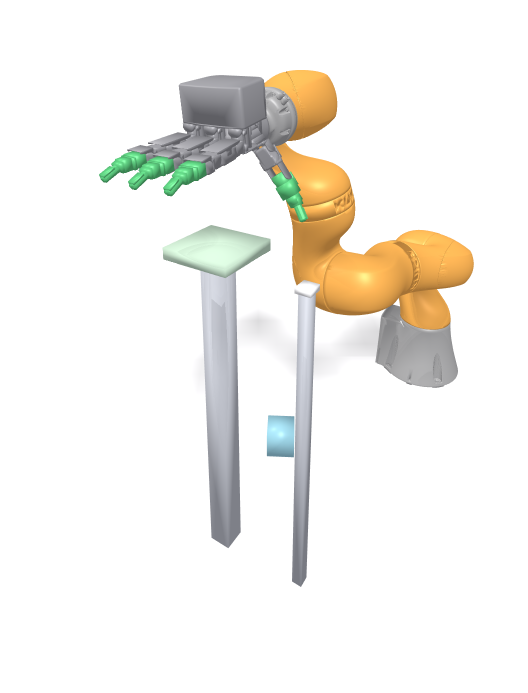} \\
\end{tabular}

\par\vspace{6pt}

\begingroup
\setlength{\tabcolsep}{6pt}
\renewcommand{\arraystretch}{1}

\begin{tabular}{@{}ccc@{}}
{\scriptsize Initial state}
& {\scriptsize Final reference goal}
& {\scriptsize Evaluation measurements} \\[3pt]

\includegraphics[height=0.27\linewidth]
    {appendix_rollouts_panels/place_ours_1_start.pdf}
&
\includegraphics[height=0.27\linewidth]
    {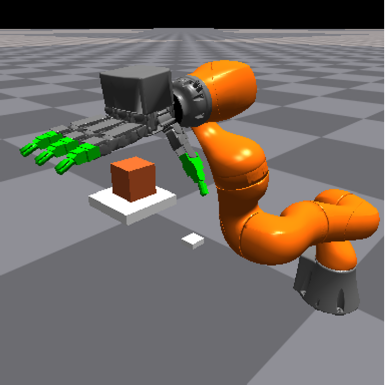}
&
\includegraphics[height=0.27\linewidth]
    {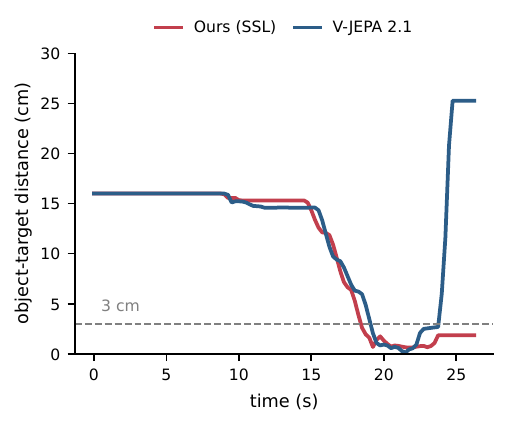}
\\
\end{tabular}
\endgroup

\endgroup
\end{minipage}

\caption{
Place rollout comparison.
Time progresses from left to right.
Curves show planar object-target distance.
Touch-JEPA-AC releases the object with a maximum
planar target error of $1.8$\,cm over the final second,
while V-JEPA~2.1 drops it away from the target.
}
\label{fig:supp-rollout-place}
\end{figure}

\subsection{Baselines and Evaluation Criteria}
\paragraph{Comparison settings.}
The V-JEPA~2.1 baseline uses the same released visual backbone
initialization, camera observations, fitted geometry, trajectories,
predictor training budget, subgoal schedule, CEM settings, and controller
as Touch-JEPA-AC. Its surface input samples concatenated features from
backbone blocks $5$, $8$, and $11$, followed by a fixed orthonormal
projection to $768$ channels and the same vertex grouping.
Thus, its surface tokens have width $768$, compared with Touch-JEPA's
learned width $256$. Predictor input and output projections accommodate
this difference.

Octo is adapted as a direct policy with a frozen backbone and trainable
action head and proprioceptive tokenizer. It receives two RGB frames,
two $35$-dimensional state vectors, and a goal image. A diffusion head
predicts a four-action chunk, and only the first is executed. The five hand commands are averaged into one grip value.
Training uses the same $696$ trajectories and $1{,}024$ updates of
$256$ samples. AdamW uses learning rate $3\times10^{-5}$, a
$100$-update warmup, and weight decay $0.1$.

\paragraph{Success criteria.}
Each task uses $50$ matched scenes across methods. Grasp-lift succeeds
when the minimum object-height increase during the final second is at
least $10$\,cm and the maximum summed fingertip-pad force over the
episode is at most $8$\,N. Place succeeds when, throughout the final
second, horizontal target error is at most $3$\,cm, vertical error is
at most $1$\,cm, and the object has been released. Object pose and
contact measurements are used to score these conditions.
Table~\ref{tab:supp-robot-ci} reports the counts with Wilson intervals.

\begin{table}[!htbp]
\centering
\caption{Planning success with uncertainty.
Counts use the same $50$ scenes per task. Parentheses give Wilson
$95\%$ confidence intervals for success percentages.}
\label{tab:supp-robot-ci}
\small
\setlength{\tabcolsep}{6pt}
\begin{tabular}{@{}lcc@{}}
\toprule
Method & Grasp-lift & Place \\
\midrule
Touch-JEPA-AC
& \rkone{$43/50$} $(73.8,93.0)$
& \rkone{$42/50$} $(71.5,91.7)$ \\
V-JEPA~2.1
& \rktwo{$27/50$} $(40.4,67.0)$
& \rktwo{$23/50$} $(33.0,59.6)$ \\
Octo
& $0/50$ $(0.0,7.1)$
& $0/50$ $(0.0,7.1)$ \\
\bottomrule
\end{tabular}
\end{table}

For grasp-lift, both planners succeed in $25$ scenes, only
Touch-JEPA-AC succeeds in $18$, only V-JEPA~2.1 succeeds in two,
and both fail in five. The corresponding place counts are
$18$, $24$, five, and three. Let $b$ count successes unique to
Touch-JEPA-AC and $c$ count successes unique to V-JEPA~2.1.
The exact two-sided McNemar probability is
\begin{equation}
p=\min\left(1,\;2\sum_{j=0}^{\min(b,c)}
\binom{b+c}{j}2^{-(b+c)}\right).
\label{eq:supp-mcnemar}
\end{equation}
This yields unadjusted $p=4.02\times10^{-4}$ and
$p=5.46\times10^{-4}$, respectively, consistent with the rounded
values in the main text. Confidence intervals describe variation
across evaluation scenes for one predictor training seed.

\endgroup

\begin{figure}[!tb]
\centering

\begin{minipage}{0.95\linewidth}
\centering
\begingroup
\setlength{\tabcolsep}{1pt}
\renewcommand{\arraystretch}{0}

\begin{tabular}{@{}
    >{\centering\arraybackslash}m{12pt}
    *{6}{>{\centering\arraybackslash}
    m{\dimexpr(\linewidth-12pt-12\tabcolsep)/6\relax}}
@{}}
& {\scriptsize Start}
& {\scriptsize Pre-grasp}
& {\scriptsize Closed}
& {\scriptsize Lift}
& {\scriptsize Carry}
& {\scriptsize End} \\[3pt]

\rotatebox{90}{\tiny\textbf{Ours (SSL)}}
& \includegraphics[width=\linewidth]
  {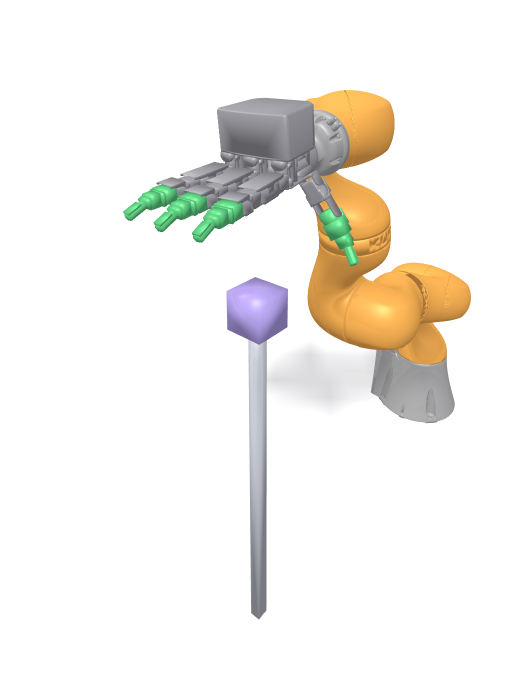}
& \includegraphics[width=\linewidth]
  {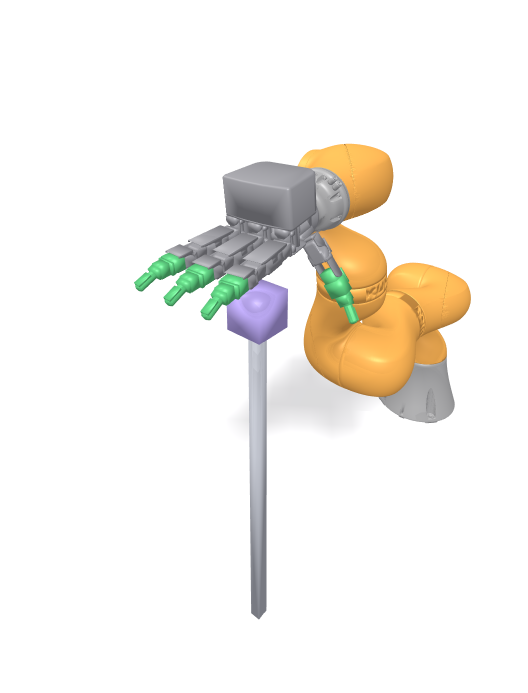}
& \includegraphics[width=\linewidth]
  {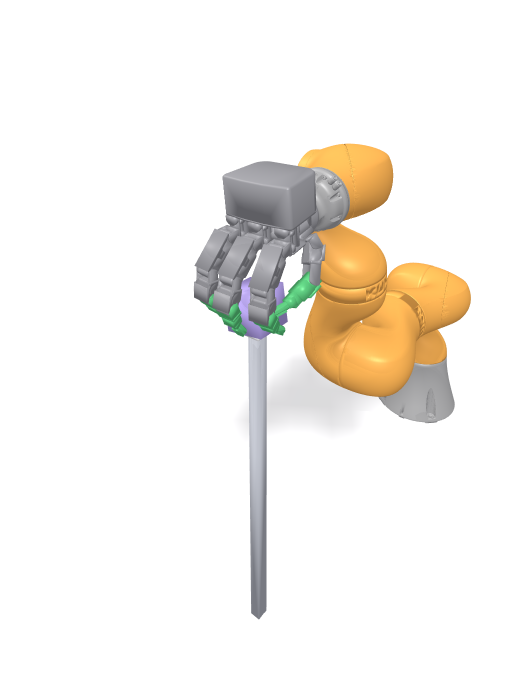}
& \includegraphics[width=\linewidth]
  {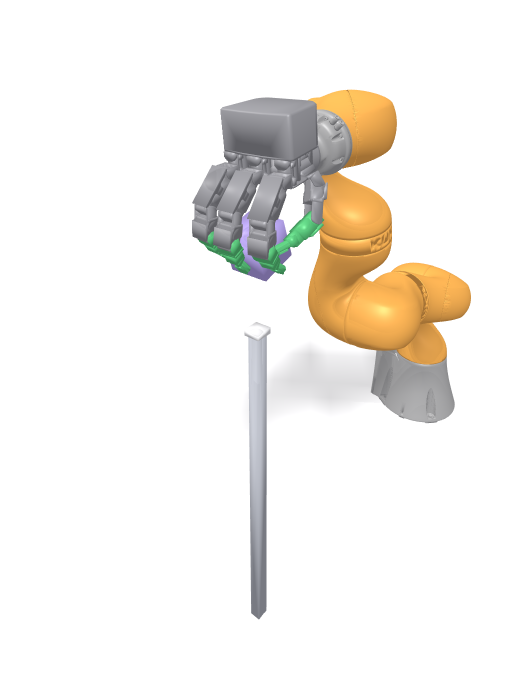}
& \includegraphics[width=\linewidth]
  {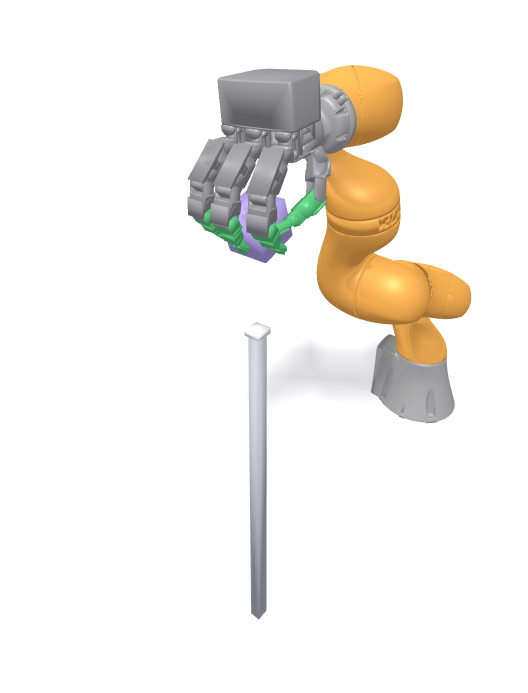}
& \includegraphics[width=\linewidth]
  {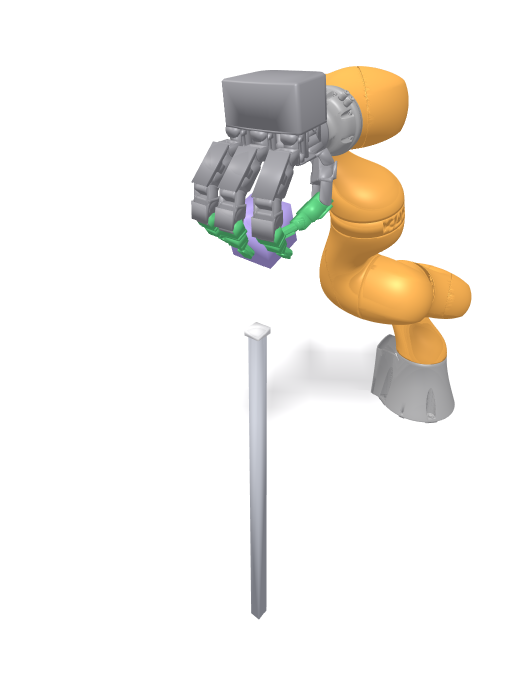} \\[2pt]

\rotatebox{90}{\tiny V-JEPA 2.1}
& \includegraphics[width=\linewidth]
  {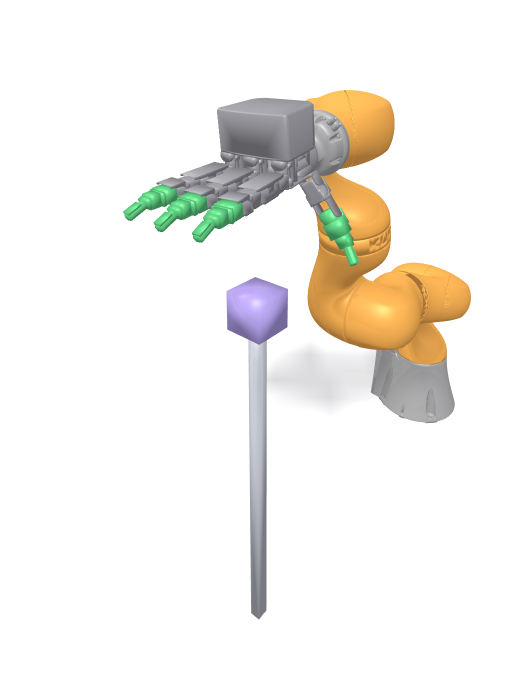}
& \includegraphics[width=\linewidth]
  {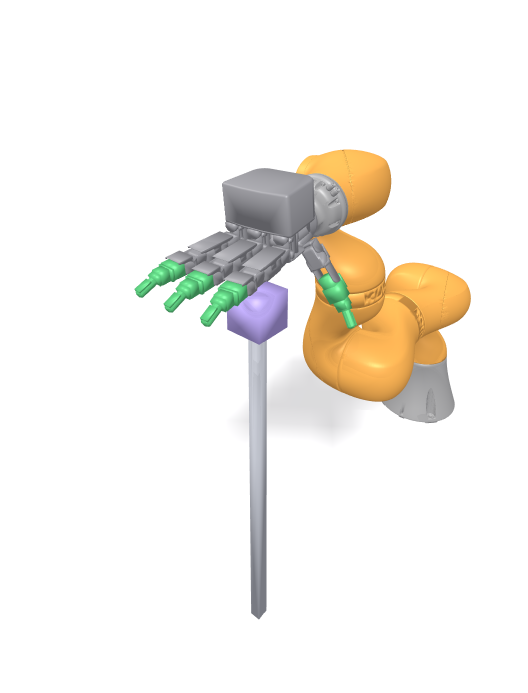}
& \includegraphics[width=\linewidth]
  {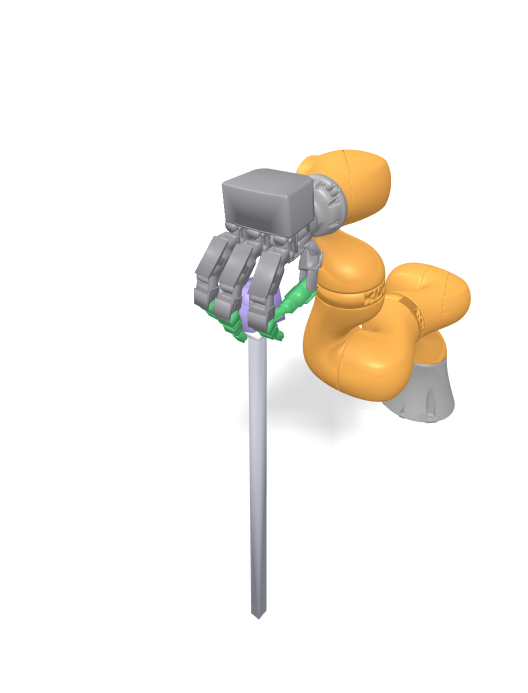}
& \includegraphics[width=\linewidth]
  {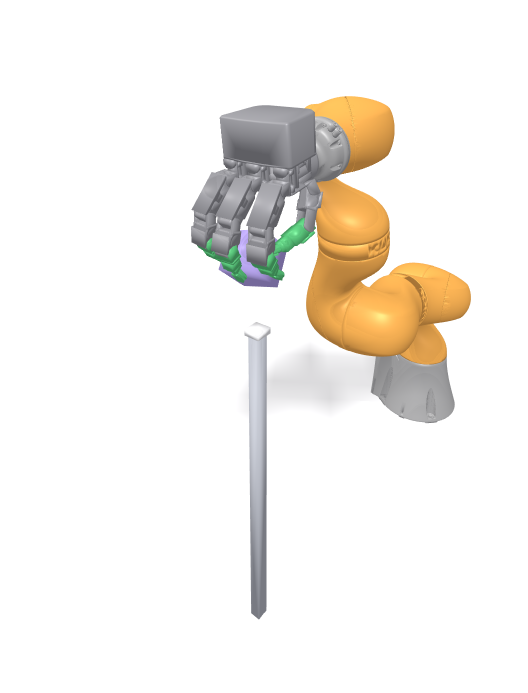}
& \includegraphics[width=\linewidth]
  {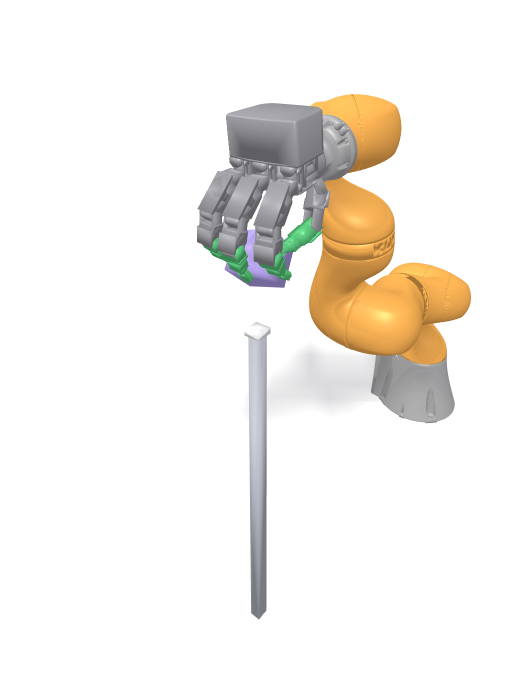}
& \includegraphics[width=\linewidth]
  {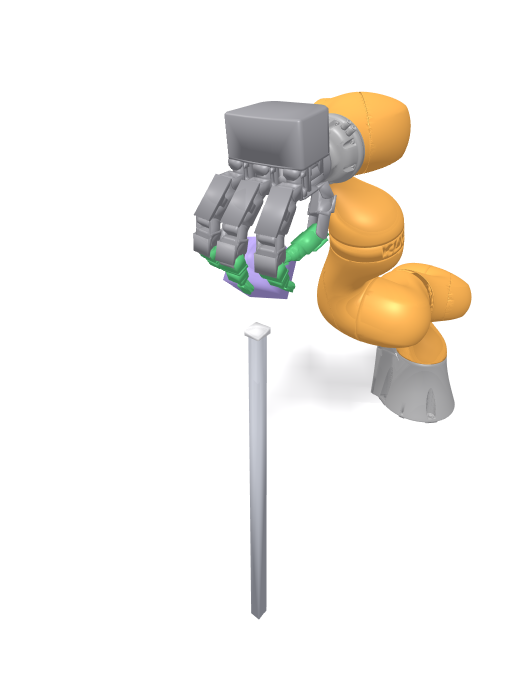} \\
\end{tabular}

\par\vspace{6pt}

\begingroup
\setlength{\tabcolsep}{6pt}
\renewcommand{\arraystretch}{1}

\begin{tabular}{@{}ccc@{}}
{\scriptsize Initial state}
& {\scriptsize Final reference goal}
& {\scriptsize Evaluation measurements} \\[3pt]

\includegraphics[height=0.27\linewidth]
    {appendix_rollouts_panels/grasp_ours_1_start.pdf}
&
\includegraphics[height=0.27\linewidth]
    {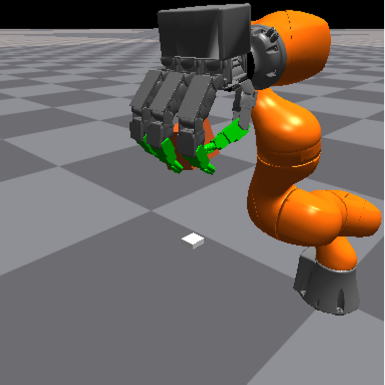}
&
\includegraphics[height=0.27\linewidth]
    {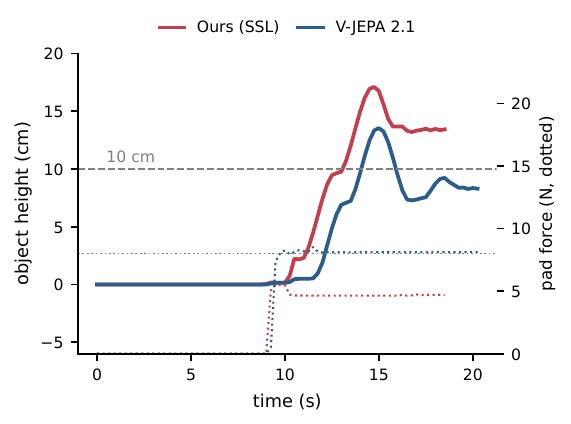}
\\
\end{tabular}
\endgroup

\endgroup
\end{minipage}

\caption{
Grasp-lift rollout comparison.
Time progresses from left to right.
Curves show object height above its initial position
and summed fingertip-pad force.
Both methods grasp and lift the object.
Touch-JEPA-AC maintains a minimum lift of $13$\,cm
over the final second with a peak force of $5.7$\,N,
whereas V-JEPA~2.1 maintains $8.2$\,cm and peaks at
$8.9$\,N, failing both the $10$\,cm height and $8$\,N
force criteria.
}
\label{fig:supp-rollout-grasp}
\end{figure}

\subsection{Qualitative Rollout Comparisons}
\label{sec:supp-rollouts}

Figure~\ref{fig:robot_rollouts} illustrates
Touch-JEPA-AC rollouts from the planner camera.
Figures~\ref{fig:robot_success} and~\ref{fig:supp-energy} to~\ref{fig:supp-rollout-grasp}
label Touch-JEPA-AC as Ours (SSL).
Figures~\ref{fig:supp-rollout-place}
and~\ref{fig:supp-rollout-grasp} additionally compare
Touch-JEPA-AC and V-JEPA~2.1 on matched scenes.
For each task, we show the lowest-index evaluation
scene in which Touch-JEPA-AC succeeds and V-JEPA~2.1 fails.
Rollout and initial-state panels are rendered from
recorded robot and object states, while goal panels
show the final reference camera images supplied
to the planner.
The accompanying curves report simulator measurements
used for evaluation.

\end{document}